\pdfoutput=1

\documentclass[11pt]{article}

\usepackage[preprint]{acl}

\usepackage{times}
\usepackage{latexsym}
\usepackage[T1]{fontenc}
\usepackage[utf8]{inputenc}
\usepackage{microtype}
\usepackage{inconsolata}
\usepackage{graphicx}

\usepackage{booktabs}
\usepackage{array}
\usepackage{amsmath}
\usepackage{amssymb}
\usepackage{xcolor}
\usepackage{colortbl}
\usepackage{threeparttable}
\newenvironment{tabnotes}{\begin{tablenotes}[flushleft]\scriptsize\itshape\setlength{\itemsep}{0pt}\setlength{\parskip}{0pt}}{\end{tablenotes}}
\usepackage{tikz}
\usepackage{soul}

\definecolor{rtShown}{HTML}{4C6FA5}\definecolor{rtShownE}{HTML}{314869}
\definecolor{rtSummarized}{HTML}{7290C1}\definecolor{rtSummarizedE}{HTML}{4A5E7D}
\definecolor{rtTold}{HTML}{94A3CE}\definecolor{rtToldE}{HTML}{606A86}
\definecolor{rtInstructed}{HTML}{B1A3CA}\definecolor{rtInstructedE}{HTML}{736A83}
\definecolor{rtSteered}{HTML}{C79BB9}\definecolor{rtSteeredE}{HTML}{816578}
\definecolor{rtTransplanted}{HTML}{D18EA4}\definecolor{rtTransplantedE}{HTML}{885C6B}
\newcommand{\mkcircle}[1]{\tikz[baseline=-0.52ex]\fill[#1] (0,0) circle (0.5ex);}
\newcommand{\mksquare}[1]{\tikz[baseline=-0.52ex]\fill[#1] (-0.45ex,-0.45ex) rectangle (0.45ex,0.45ex);}
\newcommand{\mktriup}[1]{\tikz[baseline=-0.52ex]\fill[#1] (-0.55ex,-0.47ex) -- (0.55ex,-0.47ex) -- (0,0.52ex) -- cycle;}
\newcommand{\mkdiamond}[1]{\tikz[baseline=-0.52ex]\fill[#1] (0,0.6ex) -- (0.5ex,0) -- (0,-0.6ex) -- (-0.5ex,0) -- cycle;}
\newcommand{\mktridown}[1]{\tikz[baseline=-0.52ex]\fill[#1] (-0.55ex,0.47ex) -- (0.55ex,0.47ex) -- (0,-0.52ex) -- cycle;}
\newcommand{\mkpentagon}[1]{\tikz[baseline=-0.52ex]\fill[#1] (90:0.56ex) -- (162:0.56ex) -- (234:0.56ex) -- (306:0.56ex) -- (18:0.56ex) -- cycle;}
\newcommand{\routename}[4]{\mbox{#4\hspace{0.18em}\textcolor{#2}{#3}}}
\newcommand{\Shown}{\routename{rtShown}{rtShownE}{Shown}{\mkcircle{rtShown}}}
\newcommand{\Summarized}{\routename{rtSummarized}{rtSummarizedE}{Summarized}{\mksquare{rtSummarized}}}
\newcommand{\Told}{\routename{rtTold}{rtToldE}{Told}{\mktriup{rtTold}}}
\newcommand{\Instructed}{\routename{rtInstructed}{rtInstructedE}{Instructed}{\mkdiamond{rtInstructed}}}
\newcommand{\Steered}{\routename{rtSteered}{rtSteeredE}{Steered}{\mktridown{rtSteered}}}
\newcommand{\Transplanted}{\routename{rtTransplanted}{rtTransplantedE}{Transplanted}{\mkpentagon{rtTransplanted}}}

\definecolor{promptbg}{HTML}{F1F1F1}
\newcommand{\prompt}[1]{{\sethlcolor{promptbg}\hyphenpenalty=10000 \exhyphenpenalty=10000 \hl{\texttt{#1}}}}
\usepackage[skins,raster]{tcolorbox}

\colorlet{promptrole}{black!50}
\newcommand{\ph}[1]{{\normalfont\itshape\ensuremath{\langle}#1\ensuremath{\rangle}}}
\newcommand{\role}[1]{{\normalfont\sffamily\scriptsize\bfseries\color{promptrole}\MakeUppercase{#1}}}
\newcommand{\vsp}{\textvisiblespace}

\colorlet{pbFrame}{rtShown!45!white}
\colorlet{pbBack}{rtShown!5!white}
\colorlet{pbTitleBack}{rtShown!13!white}
\colorlet{pbTitle}{rtShownE}
\colorlet{pbTag}{rtShown!17!white}
\colorlet{pbHair}{rtShown!32!white}
\colorlet{tbFrame}{rtTransplanted!50!white}
\colorlet{tbRule}{rtTransplanted}
\colorlet{tbBack}{rtTransplanted!6!white}
\colorlet{tbTitle}{rtTransplantedE}
\colorlet{tbTag}{rtTransplanted!22!white}
\colorlet{tbFoot}{rtTransplanted!14!white}

\newlength{\pbhang}
\newlength{\pblabelwd}
\newlength{\pblabelsep}
\newlength{\pbvargap}

\makeatletter
\newif\ifpb@firstvariant
\def\pb@labelfont{\normalfont\sffamily\bfseries\fontsize{7}{8}\selectfont}
\tcbset{
  pb@base/.style={
    enhanced, arc=2.5pt, boxrule=0.5pt, pb@inner,
    boxsep=0pt, left=6pt, right=6pt, top=3.5pt, bottom=3.5pt, middle=3pt,
    lefttitle=6pt, righttitle=6pt, toptitle=2pt, bottomtitle=2pt,
    fonttitle=\small\bfseries, fontlower=\small,
    lower separated, beforeafter skip=6pt plus 1pt,
  },
  pb@prompt/.style={pb@base,
    colframe=pbFrame, colback=pbBack, colbacktitle=pbTitleBack, coltitle=pbTitle, titlerule=0.4pt,
    colbacklower=white, segmentation style={solid, draw=pbHair, line width=0.4pt},
  },
  pb@trace/.style={pb@base,colframe=tbFrame, colback=tbBack, leftrule=2.6pt, top=3pt, bottom=3.5pt, middle=2.5pt,
    overlay={\begin{tcbclipframe}\fill[tbRule] (frame.south west) rectangle ([xshift=2.6pt]frame.north west);
      \end{tcbclipframe}},colbacklower=tbFoot, segmentation style={solid, draw=tbFrame, line width=0.4pt},
    space to upper,},
  pb@inner/.code={\ifinner\tcbset{box align=top}\fi},
  note/.code={\def\pb@note{#1}},
  labels/.code={\settowidth{\pblabelwd}{\pb@labelfont #1}\addtolength{\pblabelwd}{\pblabelsep}},
}

\newcommand{\pb@tag}[3]{\tikz[baseline=(pb@t.base)]\node[inner xsep=2.4pt, inner ysep=1.15pt, rounded corners=1.4pt, fill=#1, text=#2,
    font=\normalfont\sffamily\bfseries\fontsize{6}{7}\selectfont, text height=4.4pt, text depth=0pt]
    (pb@t) {\MakeUppercase{#3}};}
\newcommand{\pb@role}[1]{\pb@tag{pbTag}{pbTitle}{#1}\hskip4pt{\color{pbHair}\leaders\hrule height 2.45pt depth -2.05pt\hfill}}
\newcommand{\pb@variant}[1]{\par\ifpb@firstvariant\pb@firstvariantfalse\else\vskip\pbvargap\fi
  \leavevmode\llap{\hbox to\pblabelwd{\hss\pb@labelfont\color{pbTitle}#1\hskip\pblabelsep}}}

\def\prompt@eol{\par\prompt@open}
\def\prompt@open{\@ifnextchar\end{}{\@ifnextchar\variant{}{\leavevmode}}}
\begingroup
\catcode`\^^M=\active\catcode`\'=\active \gdef\prompt@lines{\catcode`\^^M=\active\let^^M\prompt@eol\catcode`\'=\active\let'\textquotesingle}\endgroup
\def\pb@body{\ttfamily\footnotesize\raggedright
  \parindent\z@ \parskip\z@ \leftskip\pblabelwd
  \hyphenpenalty\@M \exhyphenpenalty\@M
  \ifdefined\microtypesetup\microtypesetup{protrusion=false}\fi \everypar{\hangindent\pbhang\hangafter\@ne}\let\role\pb@role \let\variant\pb@variant \pb@firstvarianttrue
  \prompt@lines\pb@first}
\def\pb@first#1{\prompt@open}

\newenvironment{promptbox}[2][]{\let\pb@note\@empty \pblabelwd\z@
  \begin{tcolorbox}[pb@prompt,#1,title={#2}]\pb@body}
 {\par\ifx\pb@note\@empty\else\tcblower\pb@note\fi\end{tcolorbox}}

\newenvironment{tracebox}[3][]{\def\pb@answer{#3}\pblabelwd\z@
  \begin{tcolorbox}[pb@trace,#1]{\raggedright\small\bfseries\color{tbTitle}#2\par}\vskip1.5pt
  \pb@body}
 {\par\tcblower\pb@tag{tbTag}{tbTitle}{answer}\hskip5pt{\ttfamily\footnotesize\pb@answer}\end{tcolorbox}}

\newenvironment{promptrow}[1][]{\begin{tcbraster}[raster columns=2, raster equal height=rows, raster valign=top, raster column skip=8pt,
    raster row skip=6pt, raster before skip=6pt plus 1pt, raster after skip=6pt plus 1pt, #1]}
 {\end{tcbraster}}
\makeatother

\definecolor{linkblue}{HTML}{000099}
\hypersetup{colorlinks=true, citecolor=linkblue, linkcolor=linkblue, urlcolor=linkblue}

\title{Same Answer, Different Future: Dissociating Belief Content and Evidence Weight in Language Models}

\author{
  \textbf{Weihan Li\textsuperscript{1}},
  \textbf{Junhao Wu\textsuperscript{2}},
  \textbf{Xinlei Chen\textsuperscript{3}},
  \textbf{Tianshi Zheng\textsuperscript{4}}
  \\
  \normalsize
  \textsuperscript{1}The University of Tokyo,
  \textsuperscript{2}RWTH Aachen University,
  \textsuperscript{3}Harbin Institute of Technology, Shenzhen
  \\
  \normalsize
  \textsuperscript{4}The Hong Kong University of Science and Technology
  \\
  \normalsize
  \texttt{liweihan1107@g.ecc.u-tokyo.ac.jp}
}

\begin{document}
\maketitle

\begin{abstract}
Memory summaries, prompts and steering vectors set a language model's beliefs and are judged by the prediction they produce. A prediction does not say how firmly it is held. We separate a belief's \emph{content}, what it predicts now, from its \emph{weight}, the number of new observations it is worth: for an ideal observer, the amount of evidence behind it. In Bernoulli inference we set beliefs in nine open-weight models of 8 to 397B parameters and follow each belief with the same new observations. We argue that these models hold an answer without its evidence and recover its weight from the text they re-read. Weight therefore follows the form of the evidence and the question asked more than its amount: matched on their current estimate, a record of observations and a summary of its counts usually move apart under the same new observations, in either direction. Scale does not close the gap. Across the nine models 64 observations are worth 20 to 53 as a record and 9 to 151 as a summary, no closer to 64 at 397B than at 8B. Where no text stands behind a belief there is little to re-read: a steering vector can install a record's answer, and sixteen observations undo it where a neutral passage leaves it in place. Where reasoning first writes the counts into the context the weight returns, and the same 64 observations are worth 58 to 67 in either form. A matched answer is not a matched belief, and methods that set beliefs should be judged by the futures they produce.
\end{abstract}
\section{Introduction}
\label{sec:intro}

\begin{figure*}[t]
  \centering
  \includegraphics[width=\textwidth]{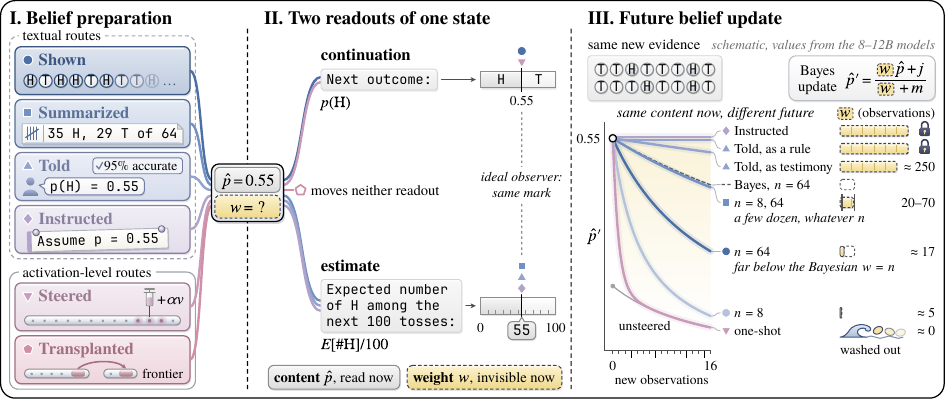}
  \caption{\textbf{A schematic of the protocol: a belief has a content that current readouts show and a weight that only an update reveals.} I. Six routes aim at the same prediction, 0.55 for heads. II. The prepared state is read as a continuation and as an estimate, and the route decides which readout shows the content. III. The same sixteen new observations follow every state, and how far each state moves gives the weight of its belief in observations (Bayes: $w = n$). Values are illustrative, taken from the 8 to 12B models.}
  \label{fig:framework}
\end{figure*}

Language models adapt their predictions to information given at inference time, and the same prediction can be reached by very different routes: observing outcomes, reading a summary of them, being told a rate, or having activations steered. Several of these routes are used to set beliefs on purpose: a memory summary stands in for the record it compresses, a system prompt states a rate, and a steering vector writes a belief into the residual stream. In a long interaction the model must then decide when such a belief should change and when it should hold \citep{xu2026when}. Studies of these routes compare them through the predictions they produce, against each other \citep{bigelow2025belief} or against a Bayesian posterior predictive \citep{falck2024is, gupta2025enough}. A prediction, however, does not say how the belief behind it will respond to new evidence.

A current prediction underdetermines a belief in two ways. First, it says nothing about how much evidence stands behind it. Under a uniform prior, one head in two tosses and twenty heads in forty both predict heads with probability 0.5, yet the next toss moves the first posterior far more than the second. Predictive notions of state therefore identify a state by its responses to future inputs \citep{nerode1958linear, shalizi2001computational, littman2002predictive}, a criterion that \citet{ma2026hidden} brought to language models by forking \emph{futures}, alternative continuations of the input, from a formed state. Second, a prediction is always read in a particular way. A model can continue a sequence of outcomes or state an estimate, and the two need not agree \citep{hu2023prompting, jang2026instructiontuned}. Two states hold the same belief only if they agree in every readout, now and under the same new evidence.

We separate two properties of a belief within each readout: its \emph{content}, the prediction the readout returns now, and its \emph{weight}, the number of new observations that prediction is worth. For an ideal observer the weight is the amount of evidence behind the belief, whatever form the evidence took and however the belief is read. Human judgment draws the same line between the strength of evidence and its weight, and confidence follows the first more than the second \citep{griffin1992weighing}. We measure both in Bernoulli inference, which has an exact Bayesian reference and lets the same evidence be written as an outcome sequence or as its counts. In nine open-weight models of 8 to 397B parameters we prepare beliefs through four textual routes and two activation-level routes (Figure~\ref{fig:framework}, I), read each as a next-outcome continuation and as an explicit estimate (II), and append the same new observations to every state (III). Six models of 8 to 32B run locally, where both readouts are read from token probabilities and activations can be written, and three of 122 to 397B are read from sampled estimates.

That equal predictions can rest on unequal evidence is familiar from Bayesian inference. Our cases are the reverse: the evidence is equal and the futures differ. A record of 64 observations and a summary of its counts are one belief for an ideal observer. Matched on their current estimate, the two move apart under the same sixteen observations in five of the six local models, by factors of 1.6 to 2.5 and in either direction (Section~\ref{sec:weight}). What a belief is worth depends less on the amount of evidence than on its form and on the question that reads it. In the 8 to 12B models a summary is worth a few dozen written-out observations whether it states 8 or 64, and in the 27 to 32B models a stated 64 is worth up to five times its record. Two questions with the same Bayesian answer give one told rate very different weights: under one no evidence moves it, and under the other it is worth 54 observations. None of this fades with size: on the same states the largest models are no closer to the evidence than the smallest (Table~\ref{tab:further}).

We trace this to what a model carries forward: \textbf{these models hold an answer without the evidence behind it}, and recover its weight each time from the text they re-read (Section~\ref{sec:discussion}). The form of the evidence then also decides where its content shows: in the 8 to 12B models a record moves the continuation and hardly the estimate, while its own counts, a told rate or an instruction move the estimate (Section~\ref{sec:content}). A steering vector installs the content of a record without its weight: sixteen observations undo it in the two 8B models, where a neutral passage of the same length leaves it in place, and it persists in a hybrid-attention model (Section~\ref{sec:writes}). When a reasoning step first writes the counts into the context, 64 observations are worth 58 to 67 in either form, and the answer follows the written count even when it is wrong (Section~\ref{sec:weight}). For methods that set beliefs on purpose, the current answer is therefore half the test: the belief must also be followed under new evidence.
\section{Content and weight of a belief}
\label{sec:state}

\providecommand{\dsgicon}[1]{\tikz[x=1mm,y=1mm,line cap=round,line join=round,baseline=1.3mm]{\useasboundingbox (0,0) rectangle (8,4);#1}}
\providecommand{\dsgpic}[1]{\node[inner sep=0pt,anchor=south west] at (0,0) {\includegraphics[width=8mm,height=4mm]{figures/icons/#1.pdf}};}
\providecommand{\dsgnl}{\,\tikz[baseline=0pt,x=1ex,y=1ex,line cap=round,line join=round]\draw[black,line width=0.42pt] (1.35,1.3)--(1.35,0.45)--(0.1,0.45) (0.5,0.85)--(0.1,0.45)--(0.5,0.05);\,}
\providecommand{\dsgnote}[1]{#1}
\providecommand{\dsgrole}[1]{{\colorlet{promptrole}{black}\def\scriptsize{\fontsize{5.2}{6}\selectfont}\role{#1}}}
\providecommand{\dsgrolecap}[1]{{\colorlet{promptrole}{black}\role{#1}}}
\providecommand{\dsgrule}{\arrayrulecolor{black!16}\specialrule{0.3pt}{0.15ex}{0.15ex}\arrayrulecolor{black}}

\begin{table*}[t]
  \centering
  \scriptsize
  \setlength{\tabcolsep}{4pt}
  \setlength{\aboverulesep}{0.3ex}\setlength{\belowrulesep}{0.5ex}
  \renewcommand{\arraystretch}{1.15}
  \begin{tabular}{@{}>{\centering\arraybackslash}m{8.4mm}>{\raggedright\arraybackslash}m{18mm}>{\raggedright\arraybackslash}m{\dimexpr\textwidth-8.4mm-18mm-32mm-6\tabcolsep\relax}>{\raggedright\arraybackslash}m{32mm}@{}}
    \toprule
    \multicolumn{2}{@{}l}{\textbf{Routes}} & \dsgnote{content line, proportion $\pi \in \{0.25, 0.5, 0.75\}$, 0.05 and 0.95 added for \Told{}, \Instructed{}} & \textbf{Ideal observer}\\
    \cmidrule{1-4}
    \dsgicon{\dsgpic{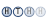}}
      & \Shown{} & \prompt{The complete observations in their recorded order are:}~\ph{$n$ outcomes} & $n$, with $n \in \{8, 16, 32, 64\}$\\
    \dsgrule
    \dsgicon{\dsgpic{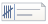}}
      & \Summarized{} & \prompt{A complete record contains}~\ph{$k$}~\prompt{H outcomes and}~\ph{$n-k$}~\prompt{T outcomes, for}~\ph{$n$}~\prompt{observations in total.} & $n$\\
    \dsgrule
    \dsgicon{\dsgpic{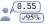}}
      & \Told{} & \prompt{On independently checked previous forecasting tasks, this source\textquotesingle s reports were judged accurate}~\ph{95 or 55}\prompt{\% of the time. A source reports that the probability of H on the next trial is}~\ph{$s$}\prompt{.} & discounted by the reliability, dropped once the evidence contradicts it (Appendix~\ref{app:bayes})\\
    \dsgrule
    \dsgicon{\dsgpic{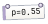}}
      & \Instructed{} & \prompt{Assume that the probability of H for this coin is}~\ph{$s$}\prompt{.}\quad\dsgnote{at three strengths,}\newline\dsgnote{up to}~\prompt{For this task, treat}\,\dots\,\prompt{as exactly}~\ph{$s$}\prompt{, regardless of any observations.} & $\infty$, a constraint\\
    \dsgrule
    \dsgicon{\dsgpic{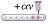}}
      & \Steered{} & a \Shown{} state plus $\alpha v$, $v$ the mean difference between records with 75\% and 25\% heads, at a quarter or a half of the depth, written once at the last 4 or 16 record positions (one-shot) or kept in place at every position (persistent) & as \Shown{}, the same content\\
    \dsgrule
    \dsgicon{\dsgpic{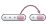}}
      & \Transplanted{} & residual stream at the last positions (frontier) of a \Shown{} record of 64, copied into one of 8 & as the donor, 64\\
    \midrule
    \multicolumn{2}{@{}l}{\textbf{Readouts}} & \dsgnote{the query, after the boundary line and a blank line} & \textbf{What is read}\\
    \cmidrule{1-4}
    \dsgicon{\dsgpic{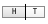}}
      & Continuation & \dsgnote{document frame}~\dsgrole{user}~\ph{task, evidence}~\prompt{Next outcome:}\newline\dsgnote{assistant frame}~\dsgrole{user}~\ph{task, evidence}~\dsgrole{assistant}~\prompt{Next outcome:} & the probability of H against T as the next token\\
    \dsgrule
    \dsgicon{\dsgpic{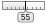}}
      & Estimate & \dsgnote{Q1}~\prompt{Expected number of H among the next 100 tosses:\vsp}\newline\dsgnote{Q2}~\prompt{Probability that the next outcome is H, as a whole-number percentage:\vsp}\newline\dsgnote{in both frames, with Q3 and Q4 in Appendix~\ref{app:queries}} & the distribution over the integers 0 to 100, its expectation\\
    \dsgrule
    \dsgicon{\dsgpic{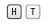}}
      & Decision check & \prompt{Next outcome? Answer with H or T.}~\dsgrole{assistant}~\ph{first token} & the first token, H against T\\
    \midrule
    \multicolumn{2}{@{}l}{\textbf{Futures}} & \dsgnote{new evidence, after the content line and before the boundary line} & \textbf{Parameters}\\
    \cmidrule{1-4}
    \dsgicon{\dsgpic{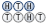}}
      & New segment & \dsgnl\prompt{New observations of this coin, in order:}~\ph{$m$ new outcomes}\quad\dsgnote{proportion $\kappa$ of H} & $m \in \{8, 16, 32\}$,\newline $\kappa \in \{0.25, 0.75\}$ (0.5 at 16)\\
    \dsgrule
    \dsgicon{\dsgpic{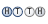}}
      & Extended line & \ph{Shown record line}~\ph{16 new outcomes}\quad\dsgnote{on the record's own line, after one space} & $m = 16$, $\kappa \in \{0.25, 0.75\}$\\
    \dsgrule
    \dsgicon{\dsgpic{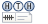}}
      & Count segment & \dsgnl\prompt{New observations of this coin:}~\ph{$k$}~\prompt{H and}~\ph{$m-k$}~\prompt{T in}~\ph{$m$}~\prompt{further tosses.} & $m$ as above, $\kappa \in \{0.25, 0.75\}$\\
    \dsgrule
    \dsgicon{\dsgpic{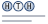}}
      & Neutral passage & \dsgnl\prompt{The town library is a quiet old brick building near the river. It has tall windows, long wooden tables, and a small} & 26 tokens in place of the segment, a placebo for \Steered{}\\
    \bottomrule
  \end{tabular}
  \caption{\textbf{The design: six routes to a belief, two readouts with a one-token decision check, and four futures.} Each row gives what the model reads (literal text on \protect\prompt{grey}, generated parts in \protect\ph{angle brackets}, \protect\dsgnl{} a line break) and, for a route, the weight in observations that a Bayesian observer with a uniform prior gives its content. \protect\dsgrolecap{user} and \protect\dsgrolecap{assistant} mark the chat turns that define the two frames (Section~\ref{sec:setup}). Every element is in Appendix~\ref{app:prompts}.}
  \label{tab:design}
\end{table*}

\paragraph{Task.}
A coin lands heads with an unknown probability. An observer with a uniform prior who has seen $k$ heads in $n$ tosses predicts heads on the next toss with probability
\begin{equation}
p = \frac{k + 1}{n + 2},
\label{eq:bayes}
\end{equation}
so the counts $(k, n)$ are a sufficient statistic: a record of the outcomes and a summary of its counts are the same evidence and give the same belief. The prediction leaves $n$ open, and $n$ decides how the belief is revised. After $m$ further tosses with $j$ heads the prediction is $(k + j + 1)/(n + m + 2)$, in which the old evidence stands against the new in the proportion $n$ to $m$ (Appendix~\ref{app:bayes}).

\paragraph{Two readouts.}
A model can express a prediction in two ways. Its \emph{continuation} is the probability it gives to heads as the next outcome of the sequence. Its \emph{estimate} is its answer to a question about the coin: by default the expected number of heads in the next hundred tosses, divided by one hundred (the \emph{expected-count question}, Q1), and in some tests the probability that the next outcome is heads (the \emph{next-toss question}, Q2), the quantity the continuation gives (Appendices~\ref{app:estimators} and \ref{app:queries}). An ideal observer derives both readouts from one posterior and gives the same number, so any disagreement between them belongs to the model.

\paragraph{Content and weight.}
The \emph{content} of a belief is the prediction $\hat p$ a readout returns now. Its \emph{weight} $w$ is the number of new observations that prediction is worth, which we measure from its futures. Take states that convey a proportion $\pi$ of heads, append the same $m$ new observations with a proportion $\kappa$ of heads, and regress the readout on both,
\begin{equation}
p_{\mathrm{post}} = c + a\,\pi + S\,\kappa .
\label{eq:two-slope}
\end{equation}
The slope $a$ is what remains of the old belief and the \emph{sensitivity} $S$ is what the new observations contribute. Their ratio, scaled by the number of new observations,
\begin{equation}
w = m\,\frac{a}{S},
\label{eq:exchange-rate}
\end{equation}
is the exchange rate between the old belief and new observations. A Bayesian observer whose prior is worth $w_0$ observations has $a = n/(w_0+n+m)$ and $S = m/(w_0+n+m)$, so $w = n$ (Appendix~\ref{app:bayes}): a record is worth its length, a summary the count it states, and an instruction infinitely many observations. Because $w$ is a ratio, a gain common to both slopes cancels, and a readout that compresses probabilities returns the same $w$. The \emph{retention} $R = a/a_0$, the share of the slope $a_0$ before the new observations that remains after them, compares routes whose content is read with different gain. A weight is measured within one readout and under the question that elicits it, and Section~\ref{sec:weight} shows that both change it.
\section{Experimental setup}
\label{sec:setup}

\begin{figure*}[t]
  \centering
  \includegraphics[width=\textwidth]{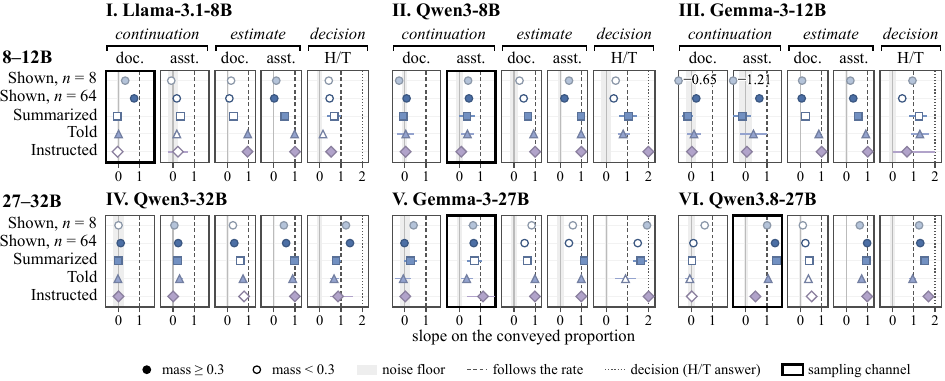}
  \caption{\textbf{A statement moves the estimate and a record the continuation, and in the larger models a record moves the estimate as well.}
    Slope of each readout on the proportion conveyed by the route (1: follows it), with 95\% family-bootstrap intervals. Columns: the continuation and the estimate in the document (doc.) and assistant (asst.) frames, and the one-token decision, where a slope of 2 is an answer that flips between T and H \citep{jang2026instructiontuned}. Heavy frame: the model's sampling channel. Gemma-3-12B's continuations at $n = 8$ run against the record (values printed).}
  \label{fig:channels}
\end{figure*}

Table~\ref{tab:design} lists every element of the design with the text the model read (full prompts: Appendix~\ref{app:prompts}).

\paragraph{Routes.}
Six routes prepare a belief, and each neighboring pair changes one thing. A \Shown{} record and its \Summarized{} counts are the same evidence as a sequence or as its sufficient statistic. A summary and a \Told{} rate set a count against testimony, and a told rate and an \Instructed{} probability set testimony against a constraint. The two activation-level routes change the medium. A \Steered{} state is a shown record plus a mean-difference direction, written once at the record's last positions (\emph{one-shot}) or kept in place at every position (\emph{persistent}). A \Transplanted{} state carries the last positions of a long record, its frontier, into a short one (Appendix~\ref{app:transplant}).

\paragraph{Frames and channels.}
A chat model can meet the query as a further line of the \dsgrolecap{user} turn (the \emph{document frame}) or as the opening of its own reply in the \dsgrolecap{assistant} turn (the \emph{assistant frame}), as in
{\abovedisplayskip=4pt\belowdisplayskip=4pt
\[
  {\footnotesize\begin{tabular}{@{}l@{\hspace{14pt}}l@{}}
    \textit{document frame} & \textit{assistant frame}\\[1pt]
    \dsgrolecap{user}\;\ph{task, evidence} & \dsgrolecap{user}\;\ph{task, evidence}\\
    \phantom{\dsgrolecap{user}}\;\prompt{Next outcome:} & \dsgrolecap{assistant}\;\prompt{Next outcome:}
  \end{tabular}}
\]}and each readout is taken in both. A readout is interpreted only where its admissible answers hold at least 0.3 of the probability mass. The weight experiments read one continuation and one estimate per model, chosen by a rule fixed before them. The \emph{sampling channel} is the continuation from which the model would sample the next outcome, in the frame where it follows a record of 64 observations with a slope above 0.3 and moves with new ones: the document frame in Llama-3.1-8B and the assistant frame in Qwen3-8B, Gemma-3-27B and Qwen3.8-27B. Gemma-3-12B and Qwen3-32B have none, and Gemma-3-27B's is close to deterministic, so its weights there are not interpreted. The \emph{report channel} is the estimate the model states in the assistant frame (Appendix~\ref{app:estimators}).

\paragraph{Futures.}
New observations arrive as a \emph{new segment} on a line of its own, as an \emph{extended line} that continues the record, or as a \emph{count segment} that states them as a count, and a neutral passage serves as a placebo. By default the new segment holds sixteen observations with 25\% or 75\% heads.

\paragraph{Models and statistics.}
Nine open-weight chat models follow one protocol, each read as deeply as it can be accessed. Six run locally, where both readouts are read from token probabilities and activations can be written: Llama-3.1-8B \citep{grattafiori2024llama3}, Qwen3-8B \citep{yang2025qwen3} and Gemma-3-12B \citep{gemmateam2025gemma3}, the \emph{smaller models}, and Qwen3-32B, Gemma-3-27B and Qwen3.8-27B \citep{qwenteam2026qwen38}, the \emph{larger models} (Appendix~\ref{app:models}). Qwen3.8-27B is a hybrid: three of every four layers replace attention over earlier positions with a recurrent state. The \emph{largest models}, Qwen3.5-122B, Qwen3-235B and Qwen3.5-397B, are mixture-of-experts models with 10 to 22B active parameters, read through sampled estimates (Appendix~\ref{app:further}). Seven predictions for the larger models were derived from the smaller ones and fixed before the larger models were run (Appendix~\ref{app:predictions}). Intervals are 95\% bootstrap intervals over families, the independently generated records with their wordings (15 to 180 per cell), and equivalence is tested against each model's wording noise floor (Appendix~\ref{app:estimators}).
\section{Content: the format of the evidence decides the readout}
\label{sec:content}

\begin{figure*}[t]
  \centering
  \includegraphics[width=\textwidth]{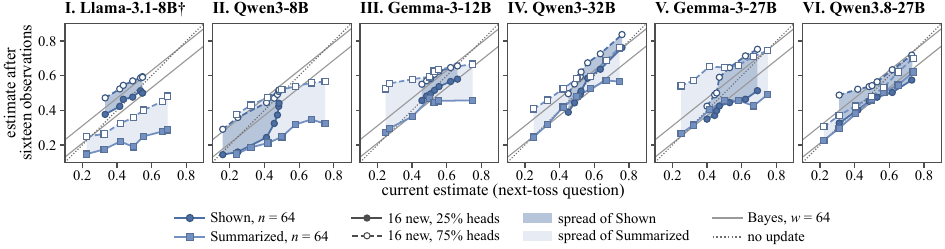}
  \caption{\textbf{Same answer, different future.}
    Estimate after sixteen new observations against the current estimate, under the next-toss question, for a record and a summary of 64 observations. Solid: after sixteen observations with 25\% heads. Hollow: with 75\% heads. The shaded gap between the two is how far a route moves. Grey: a Bayesian observer whose estimate is worth 64 observations. $^\dagger$: some answers after a summary are counts of the new segment, which lowers their level and leaves the gap unchanged (Appendix~\ref{app:matched}, with all routes and readouts).}
  \label{fig:matched-main}
\end{figure*}

\begin{figure*}[t]
  \centering
  \includegraphics[width=\textwidth]{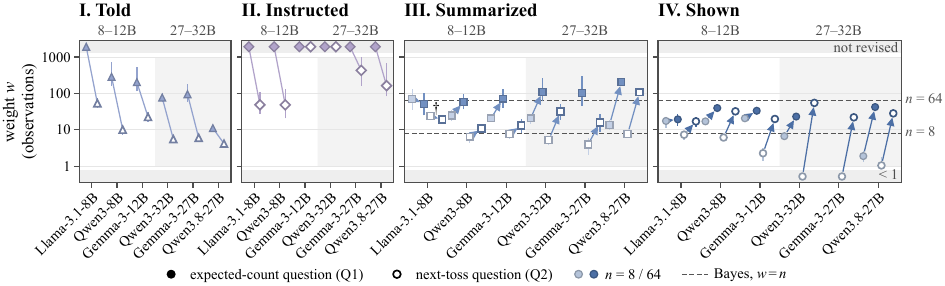}
  \caption{\textbf{What a belief is worth in the report channel, by route, model and question.} Weight $w$ in observations under sixteen new observations (filled: expected-count question, hollow: next-toss question). Arrows run from $n = 8$ to $n = 64$ against the Bayesian values (dashed). Top band: not revised (the interval of $S$ includes zero, or $w$ exceeds 1,000). Bottom band: worth less than one observation. $^\dagger$: some answers are counts of the new segment (Appendix~\ref{app:matched}). Values in Appendix~\ref{app:queries}, the largest models in Tables~\ref{tab:further} and \ref{tab:further-weights}.}
  \label{fig:weights}
\end{figure*}

\paragraph{A statement moves the estimate, a record the continuation.}
In the smaller models the estimate follows what is stated and hardly what is shown (Figure~\ref{fig:channels}). A told rate, an instruction or the counts of a record move it with slopes of 0.5 to 1.0 (0.31 for a summary of 64 in Gemma-3-12B), and a record of 64 observations with 0.02, 0.20 and 0.32. The continuation does the reverse where it can be read. Llama-3.1-8B's follows a record of 64 with a slope of 0.74 and a told rate with 0.01, and Qwen3-8B's follows the record with 0.45, an instruction with 0.09 and a told rate or a summary in part (0.38 and 0.35). A record and its counts, one belief for an ideal observer, are thus held in different readouts. Asking for the next toss narrows this gap without closing it: under the next-toss question the smaller models' estimates follow a record of 64 at 0.43 to 0.66 and its counts at 0.93 to 1.02 (Appendix~\ref{app:queries}). The one-token decision check answers from the statement as well (Appendix~\ref{app:controls}). The separation is one reason why in-context outcomes outweigh a stated bias in next-token probabilities \citep{gupta2025enough}: the statement barely reaches that readout.

\paragraph{The record reaches the estimate through computation, aloud in the smaller models and silent in the larger.}
Counting closes the gap in Qwen3-8B: after it counts the heads aloud, its estimate follows the record at 0.96, and after a thinking block at 0.84. A statement arrives in the form the estimate needs, and the model repeats it. A record must first be compressed, and counting and reasoning are that compression, done out loud. Simulation is the reverse conversion: after Qwen3-8B writes out sixteen outcomes of the coin, its continuation carries instructed and summarized content at 0.93 and 1.03, against 0.10 and 0.27 before (Appendix~\ref{app:controls}). The larger models compress in silence, and better than they count: their estimates follow a record of 64 at 0.59 in Qwen3-32B and 0.97 in Qwen3.8-27B, although they count its heads correctly in only 5 and 7\% of states, and Qwen3.8-27B's continuation follows a told rate in turn. With scale, the two readouts come to read the same evidence.
\section{Weight: route and question decide what a belief is worth}
\label{sec:weight}

\paragraph{States with the same estimate have different futures.}
The most direct test needs no regression. We take states of a record and of a summary of 64 observations whose estimates under the next-toss question, where a record reaches the estimate in every model (Section~\ref{sec:content}), agree within 0.02, and append the same sixteen observations to both (Figure~\ref{fig:matched-main}). In four models the summary then moves 1.6 to 2.2 times as far as the record, and in Qwen3.8-27B the record moves 2.5 times as far as the summary. Only Qwen3-32B moves both alike, by the amount a Bayesian observer of 64 observations would (Appendix~\ref{app:matched}), and it does so under this question only: under the expected-count question it gives the summary five times the weight of the record (Table~\ref{tab:further}). The same answer has a different future in five of the six models. How far a state moves also depends on the gain of the readout, which cancels in the weights that follow.

\paragraph{Neither a record nor its counts carry the weight of the evidence.}
In the sampling channel a record is worth far less than its length (Appendix~\ref{app:rates}): 17 observations at $n = 64$ in Llama-3.1-8B, where the weight hardly grows beyond $n = 32$, 3 to 7 at every length in Qwen3-8B, and a third of its length in Qwen3.8-27B. In the report channel of the smaller models the weight of a summary hardly follows the count it states: from a stated 8 to a stated 64 it changes by a factor of 0.7 to 3.5 under the expected-count question and 0.8 to 1.7 under the next-toss question, where the count grows eightfold (Figure~\ref{fig:weights}). In the larger models the stated count begins to matter, and the form of the evidence still decides how much. The weight of a summary grows 5 times in Qwen3-32B and 15 times in Qwen3.8-27B, past the 64 it states and to four to five times the weight of a record of 64 (Table~\ref{tab:further}).

\paragraph{Evidence is weighted by the unit in which it arrives.}
At $n = 64$ the same sixteen outcomes, continued on the record's own line instead of in a new segment, are all but ignored: the continuation responds to them with a slope of zero in Llama-3.1-8B, 0.08 in Qwen3-8B and 0.18 in Qwen3.8-27B (Appendix~\ref{app:rates}). The form of new evidence matters as well: stated as a count, sixteen further observations move the estimate about twice as far as the same observations written out, in Qwen3-32B and Qwen3.8-27B as in the smaller models, and against them the weight of a summary in the smaller models grows 2.5 to 4.0 times with its eightfold count (Appendix~\ref{app:controls}). These models weigh evidence as the text lays it out, and the same outcomes are worth more or less depending on where they stand.

\paragraph{A statement is weighted by its kind, and the question that reads it sets the scale.}
Under the expected-count question no model revises an instruction, which acts as a constraint. A told rate is a rule for Llama-3.1-8B, which leaves it in place after sixteen new observations, and testimony for the other five models, which revise it with weights from 11 to 288 observations. An observer with a prior on the source's reliability would discount the statement from the start and drop it once the evidence contradicts it, and only Qwen3-32B does both (Appendix~\ref{app:rates}). The next-toss question turns every told rate into testimony. Llama-3.1-8B now revises it, the weights in the other models fall by factors of 3 to 29, and four models revise an instruction as well (Figure~\ref{fig:weights}, Appendix~\ref{app:queries}).

\begin{table}[t]
  \centering
  \footnotesize
  \setlength{\tabcolsep}{2.6pt}
  \providecommand{\wtrackunit}{0.15cm}\renewcommand{\wtrackunit}{0.115cm}
  \begin{threeparttable}
  \providecommand{\cint}[1]{{\fontsize{7}{8}\selectfont #1}}
\providecommand{\wtrackunit}{0.15cm}\providecommand{\wtrack}[5]{\tikz[baseline=-0.55ex, x=\wtrackunit, y=1ex]{\fill[black!7] (-3,-0.62) rectangle (3,0.62);
    \draw[black!55, line width=0.5pt] (0,-0.62) -- (0,0.62);
    \draw[#4, line width=0.7pt] (#2,0) -- (#3,0);
    \if#5c\fill[#4] (#1,0) circle (0.42ex);\fi
    \if#5s\fill[#4] ([xshift=-0.4ex,yshift=-0.4ex]#1,0) rectangle ++(0.8ex,0.8ex);\fi
    \if#5d\fill[#4] ([yshift=0.55ex]#1,0) -- ([xshift=0.45ex]#1,0) -- ([yshift=-0.55ex]#1,0) -- ([xshift=-0.45ex]#1,0) -- cycle;\fi
    \if#5r\draw[#4, line width=0.7pt, ->] (1.6,0) -- (3,0);\fi
    \if#5l\draw[#4, line width=0.7pt, ->] (-1.6,0) -- (-3,0);\fi}}
\providecommand{\notrack}{\tikz[baseline=-0.55ex, x=\wtrackunit, y=1ex]{\fill[black!7] (-3,-0.62) rectangle (3,0.62);}}
\begin{tabular}{@{}l cc cc@{}}
  \toprule
  & \multicolumn{2}{c}{Expected count} & \multicolumn{2}{c}{Next toss} \\
  \cmidrule(lr){2-3} \cmidrule(l){4-5}
  Model & record & summary & record & summary \\
  \midrule
  Llama-3.1-8B & \makebox[1.5em][r]{25}\,\wtrack{-1.36}{-1.97}{-0.90}{rtShown}{c}\rlap{$^\dagger$} & \makebox[1.5em][r]{65}\,\wtrack{0.03}{-0.44}{0.47}{rtSummarized}{s} & \makebox[1.5em][r]{21}\,\wtrack{-1.60}{-1.94}{-1.28}{rtShown}{c} & \makebox[1.5em][r]{22}\,\wtrack{-1.54}{-1.91}{-1.24}{rtSummarized}{s} \\
  Qwen3-8B & \makebox[1.5em][r]{40}\,\wtrack{-0.69}{-0.90}{-0.49}{rtShown}{c} & \makebox[1.5em][r]{56}\,\wtrack{-0.20}{-0.58}{0.18}{rtSummarized}{s} & \makebox[1.5em][r]{32}\,\wtrack{-1.02}{-1.25}{-0.83}{rtShown}{c} & \makebox[1.5em][r]{13}\,\wtrack{-2.29}{-2.52}{-2.07}{rtSummarized}{s} \\
  Gemma-3-12B & \makebox[1.5em][r]{46}\,\wtrack{-0.48}{-0.73}{-0.26}{rtShown}{c} & \makebox[1.5em][r]{52}\,\wtrack{-0.29}{-0.78}{0.21}{rtSummarized}{s} & \makebox[1.5em][r]{21}\,\wtrack{-1.58}{-1.77}{-1.40}{rtShown}{c} & \makebox[1.5em][r]{13}\,\wtrack{-2.26}{-2.67}{-1.88}{rtSummarized}{s} \\
  \addlinespace
  Qwen3-32B & \makebox[1.5em][r]{24}\,\wtrack{-1.44}{-1.60}{-1.30}{rtShown}{c} & \makebox[1.5em][r]{128}\,\wtrack{1.00}{0.49}{1.55}{rtSummarized}{s} & \makebox[1.5em][r]{53}\,\wtrack{-0.26}{-0.56}{0.00}{rtShown}{c} & \makebox[1.5em][r]{42}\,\wtrack{-0.60}{-1.05}{-0.13}{rtSummarized}{s} \\
  Gemma-3-27B & \makebox[1.5em][r]{n.r.}\,\notrack & \makebox[1.5em][r]{n.r.}\,\notrack & \makebox[1.5em][r]{24}\,\wtrack{-1.42}{-1.73}{-1.12}{rtShown}{c} & \makebox[1.5em][r]{13}\,\wtrack{-2.31}{-2.86}{-1.84}{rtSummarized}{s} \\
  Qwen3.8-27B & \makebox[1.5em][r]{38}\,\wtrack{-0.73}{-0.92}{-0.55}{rtShown}{c} & \makebox[1.5em][r]{151}\,\wtrack{1.24}{1.07}{1.42}{rtSummarized}{s} & \makebox[1.5em][r]{28}\,\wtrack{-1.19}{-1.34}{-1.05}{rtShown}{c} & \makebox[1.5em][r]{107}\,\wtrack{0.74}{0.54}{0.94}{rtSummarized}{s} \\
  \addlinespace
  Qwen3.5-122B & \makebox[1.5em][r]{28}\,\wtrack{-1.18}{-1.42}{-0.97}{rtShown}{c} & \makebox[1.5em][r]{37}\,\wtrack{-0.79}{-1.07}{-0.51}{rtSummarized}{s} & \makebox[1.5em][r]{23}\,\wtrack{-1.50}{-1.80}{-1.23}{rtShown}{c} & \makebox[1.5em][r]{21}\,\wtrack{-1.58}{-1.84}{-1.34}{rtSummarized}{s} \\
  Qwen3-235B & \makebox[1.5em][r]{34}\,\wtrack{-0.91}{-1.08}{-0.75}{rtShown}{c} & \makebox[1.5em][r]{100}\,\wtrack{0.65}{0.19}{1.12}{rtSummarized}{s} & \makebox[1.5em][r]{28}\,\wtrack{-1.17}{-1.43}{-0.91}{rtShown}{c} & \makebox[1.5em][r]{32}\,\wtrack{-0.99}{-1.30}{-0.65}{rtSummarized}{s} \\
  Qwen3.5-397B & \makebox[1.5em][r]{20}\,\wtrack{-1.69}{-1.93}{-1.45}{rtShown}{c} & \makebox[1.5em][r]{8.7}\,\wtrack{-2.89}{-3.00}{-2.51}{rtSummarized}{s} & \makebox[1.5em][r]{27}\,\wtrack{-1.27}{-1.55}{-1.02}{rtShown}{c} & \makebox[1.5em][r]{25}\,\wtrack{-1.35}{-1.73}{-1.00}{rtSummarized}{s} \\
  \bottomrule
\end{tabular}
  \begin{tabnotes}
    \item n.r.: not read (answer mass below 0.3). $^\dagger$: the estimate hardly follows this record (slope 0.02).
  \end{tabnotes}
  \end{threeparttable}
  \caption{\textbf{What 64 observations are worth in the nine models}, on the same states (Appendix~\ref{app:matched}). Track: $w$ relative to 64 on a log scale from $1/8$ to 8, with its 95\% interval (tick: Bayes, $w = 64$).}
  \label{tab:further}
\end{table}

\paragraph{Up to 397B parameters, the weight comes no closer to the evidence.}
Table~\ref{tab:further} puts the nine models on the states of Figure~\ref{fig:matched-main}, where the weights of the local models lie within a factor of 1.4 of those in Figure~\ref{fig:weights}. Across them 64 observations are worth 20 to 53 as a record and 9 to 151 as a summary, on average a factor of 2.1 from 64 in the smaller and the larger models and 2.5 in the largest. What changes with size is the form of the dependence. In the largest models a record and its summary weigh alike under the next-toss question, at 21 to 32 observations, and part under the expected-count question: the summary is worth 1.3 and 2.9 times its record in Qwen3.5-122B and Qwen3-235B and 0.44 times in Qwen3.5-397B, where 64 stated observations are worth 9 under one question and 25 under the other (Appendix~\ref{app:further}).

\paragraph{Written counts carry their weight.}
When Qwen3.8-27B, Qwen3-32B and Qwen3.5-122B reason before they answer, their traces write the counts of the old evidence, add the new ones and answer from the total, a miscount included (Appendix~\ref{app:reasoning}). A record and a summary of 64 observations are then worth 58 to 67. Of three proprietary models, read through their replies, GPT-6 Sol gives a summary 0.55 to 0.73 times the weight of its record, and Gemini 4 Argon and Claude Opus 5.5 weigh both near 64, the second with a reasoning step that cannot be switched off (Appendix~\ref{app:hosted}).
\section{Steering: the same write is undone in two models and kept in a third}
\label{sec:writes}

\begin{figure}[t]
  \centering
  \includegraphics[width=\columnwidth]{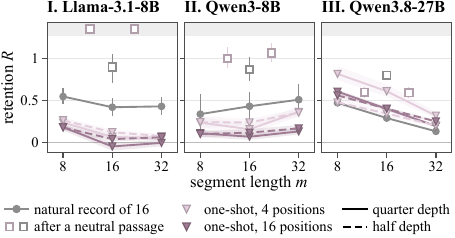}
  \caption{\textbf{The same one-shot write is undone in Llama-3.1-8B and Qwen3-8B and kept in Qwen3.8-27B.}
    Retention $R$ of the written content against the number of new observations, with the natural record of sixteen in grey. Hollow squares: retention after a neutral passage of the same length, for the two writes of four positions and the natural record. Llama-3.1-8B's two written values, 1.56 and 1.66, lie in the top band. Family-bootstrap 95\% intervals.}
  \label{fig:writes-main}
\end{figure}

\paragraph{In the two 8B models a steering vector installs the answer, and the first evidence undoes it.}
A one-shot write of the mean-difference direction between records with 75\% and 25\% heads, at the last positions of a \Shown{} state, installs in both the content difference between those records (Appendix~\ref{app:writes}). After sixteen new observations the write retains at most 0.12 of its content in Llama-3.1-8B and at most 0.24 in Qwen3-8B, where the natural records retain 0.42 and 0.43 (Figure~\ref{fig:writes-main}). The sensitivity to the new observations is unchanged, so only the written content is lost. Evidence does the undoing: after a neutral passage of the same length the written content is still in place. The write also stays in the readout of the route it imitates: between the two signs of the vector the estimate differs by at most 0.005 in Llama-3.1-8B and 0.05 in Qwen3-8B. Kept in place as a persistent write, the vector is an offset on the log-odds that evidence does not remove \citep{bigelow2025belief}.

\paragraph{In Qwen3.8-27B the write is kept like a natural belief.}
Written at sixteen positions, the same write installs the full content difference and retains 0.40 of it after sixteen observations, where the natural record retains 0.29 (0.34 and 0.61 at four positions). It also reaches the estimate, which moves by about half the difference between the natural records. A matched current output therefore does not say whether a steered belief will hold, and the model that keeps it is the one here whose layers mostly carry a recurrent state. The other three models have no continuation in which a write can be judged (Appendix~\ref{app:writes}).

\paragraph{Weight cannot be written in as a direction.}
The difference between records of 64 and 8 observations, added to the short record, leaves its weight at 3 to 5 observations against 17 for the long record in Llama-3.1-8B, and at 2 to 3 against 21 in Qwen3.8-27B. Copying the frontier of a long record into a short one moves the recipient toward the donor by at most 0.14 of the content difference and leaves its weight where it was: 1.8 to 2.1 observations in Qwen3.8-27B, against 2.0 under sham and 21 for the long record (Appendix~\ref{app:transplant}). Neither the direction nor the frontier carries a long record's weight.
\section{Related work}
\label{sec:related}

\paragraph{States defined by their futures.}
We keep the forked futures of \citet{ma2026hidden} and make the route by which a belief arrives the independent variable. Belief-state geometry \citep{shai2024transformers, balcells2026belief, sarfati2026shape} decodes the belief of a model continuing a stream of evidence from its activations, one of our two readouts, and \citet{li2026erased} separate what stays decodable from what decisions still use. Task and function vectors \citep{hendel2023incontext, todd2024function} carry a task's identity at the last position, where a record keeps none of its weight.

\paragraph{Routes to a belief.}
\citet{bigelow2025belief} matched beliefs installed by demonstrations and by activation steering through their current answer curves and fitted a closed form in which steering adds a constant to the log-odds while evidence accumulates sublinearly. Their steering acts throughout the forward pass that reads the evidence, as in our persistent write, and the one-shot write falls outside that form. \citet{slocum2025believe} find that a fact implanted by prompting, editing or fine-tuning is held with different depth, \citet{cohen2024evaluating} that an edited fact fails to propagate to the facts that depend on it, and \citet{davidson2025common} that demonstrations and instructions recruit different components. Whether later computation uses a written activation has been studied separately \citep{shih2026when, kang2026promptactivation, heyman2026steer, cheng2026decomposing, mishra2026steered}. How far a context moves an answer has been measured against prior knowledge \citep{du2024context}, and we measure this resistance in observations.

\paragraph{Readouts of a probabilistic belief.}
In-context learning has been explained as implicit Bayesian inference \citep{xie2022explanation}, and behavioral studies compare next-outcome predictions with the posterior predictive \citep{falck2024is, gupta2025enough, chlon2025llms}, which in the Bernoulli case carries no information about the evidence behind it. Stating a belief differs from sampling it: models answer questions about stated probabilities that their completions ignore \citep{mondal2026implicit}, cannot state the bias of a coin they were shown \citep{treutlein2024connecting}, and describe distributions they cannot sample from \citep{hu2023prompting, jang2026instructiontuned, meister2025benchmarking, gu2026illusion}. \citet{smolin2026beliefs} find that a single latent degree of belief predicts the outputs of strong models, and the smaller models here have at best one such degree per readout. How a model revises a stated belief depends on the confidence and reliability of its source \citep{kumaran2026competing, du2026its, atwell2026basil} and on how it is asked \citep{cekinmez2026query, lei2026elicitation}.
\section{Discussion and conclusion}
\label{sec:discussion}

An ideal observer keeps one count of its evidence and answers every question from it. In nine models of 8 to 397B parameters, the weight of a belief is set instead by the form of its evidence and by the question that reads it. With size the dependence takes another form, as a stated count goes from nearly ignored to over-weighted. The predictions fixed in advance split accordingly: those on what sets the weight held in five of six cases, those on its form in three of eight (Appendix~\ref{app:predictions}).

We propose that these models keep no count of the evidence behind a belief: each readout re-estimates from the tokens it attends to, and the weight of a belief is whatever that re-reading recovers. A one-shot write then has content without weight, because no evidence stands behind it when the readout looks back. A summary states its count as a number, and a direct answer takes the proportion from it and leaves the amount. The question changes the weight because it changes what is re-read. Hiding the first 48 outcomes of a record of 64 from the positions after it leaves Llama-3.1-8B 0.34 of its content and 0.29 of its weight, although the visible outcomes were computed with the hidden ones in view (Appendix~\ref{app:visibility}). A reasoning step tests the account from the other side: once the counts are written into the context, the answer follows the written total and the weight returns (Section~\ref{sec:weight}). ``The model believes $p$'' is therefore incomplete twice: it must say which readout is meant, and it leaves open what the belief is worth (Appendix~\ref{app:philosophy}).

Two consequences follow for practice. A steered belief should be tested by appending evidence, since its output does not show whether it will hold (Section~\ref{sec:writes}). A memory summary does not update like the record it replaces, and the error has no fixed sign, so no constant corrects it. Memory compaction should preserve the weight of a belief along with its prediction, and stating the count does so only if the model also computes with it. A matched answer is not a matched belief, and methods that set beliefs should be judged by the futures they produce.
\section*{Limitations}
We study in-context beliefs about the rate of a Bernoulli source, a task with an exact reference and sufficient statistic. Multinomial posteriors and richer priors are natural extensions. We study beliefs reported without intermediate text. Reasoning writes evidence into the context, so a reasoned answer reports a belief about a written count; we use it to test our account (Appendix~\ref{app:reasoning}).

Gemma-3-12B has no sampling channel in our prompt format, where its continuation moves against new evidence (Appendix~\ref{app:controls}), so it enters the weight experiments only through its estimate. Among the larger models, Qwen3-32B has no sampling channel and Gemma-3-27B's is close to deterministic, so the comparison across scale rests on the report channel and, for activation writes, on Qwen3.8-27B alone. The largest and three proprietary models are read as served, from sampled estimates alone (Appendices~\ref{app:further} and \ref{app:hosted}). For the proprietary models, their replies cannot reveal whether weight arises in the answer itself or in preceding computation.

As the only hybrid model whose activations we write, Qwen3.8-27B does not separate architecture from model generation, and no attention-only model of its size offers a continuation in which a write can be judged (Appendix~\ref{app:writes}). Our evidence on where the weight resides is behavioral: neither a copied frontier nor a mean-difference direction gives a short record the weight of a long one, with intervals in Qwen3.8-27B and point estimates for the copied frontier in Llama-3.1-8B (Appendices~\ref{app:transplant}, \ref{app:writes} and \ref{app:visibility}). A circuit-level account of how attention re-reads distributed evidence, and whether a model can predict its own updating weight \citep{lindsey2025introspective, binder2025looking, ferrara2026owmi}, remain open (Appendix~\ref{app:philosophy}). The routes we test set a belief in the context or activations. Other routes, such as external memory and retrieval, model editing and fine-tuning, are not tested here; the exchange-rate protocol applies once a future can be forked from the resulting state. One informative comparison is between two models fine-tuned to the same prediction, one on evidence-like examples and the other directly on the target probability. Whether the two revise that prediction at the same rate under contradicting evidence, and along the same curve, is a question for future work.
\section*{Ethics Statement}
This study involves no human subjects and no personal data. The task is produced programmatically, and no result concerns a real physical, biological or social system. Six models are open weights run locally. The three largest models, Qwen3.5-122B and Qwen3.5-397B \citep{qwenteam2026qwen35} and Qwen3-235B \citep{yang2025qwen3}, and three proprietary models, GPT-6 Sol \citep{openai2026gpt6sol}, Gemini 4 Argon \citep{google2026gemini4argon} and Claude Opus 5.5 \citep{anthropic2026opus55}, were accessed through provider APIs under the providers' terms of use. Every call requested a sampling temperature of 1.0.

\bibliography{references}

@misc{ma2026hidden,
  author        = {Ma, SiYuan and Luo, Yiqin and Zhangji and Xiao, Canran and Gao, Albert and Wang, Wei and Wu, Qiwei and Li, Xinran and Wei, Jinfeng and Zhang, Qixin},
  title         = {Hidden {APIs} in Language Models: Discovering Reusable Causal Interfaces from Forked Futures},
  year          = {2026},
  eprint        = {2607.27617},
  archivePrefix = {arXiv},
  primaryClass  = {cs.AI}
}

@misc{li2026erased,
  author        = {Li, Weihan and Zheng, Tianshi and Wu, Junhao and Chen, Xinlei},
  title         = {Erased, Rerouted, or Rescaled? Post-Training and the Causal Quotient of a Language Model's Belief State},
  year          = {2026},
  eprint        = {2610.05292},
  archivePrefix = {arXiv},
  primaryClass  = {cs.LG}
}

@inproceedings{bigelow2025belief,
  author        = {Bigelow, Eric and Wurgaft, Daniel and Wang, YingQiao and Goodman, Noah and Ullman, Tomer and Tanaka, Hidenori and Lubana, Ekdeep Singh},
  title         = {Belief Dynamics Reveal the Dual Nature of In-Context Learning and Activation Steering},
  booktitle     = {Proceedings of the 43rd International Conference on Machine Learning},
  series        = {Proceedings of Machine Learning Research},
  volume        = {306},
  pages         = {8248--8276},
  year          = {2026},
  eprint        = {2511.00617},
  archivePrefix = {arXiv},
  primaryClass  = {cs.LG}
}

@misc{sarfati2026shape,
  author        = {Sarfati, Rapha{\"e}l and Bigelow, Eric and Wurgaft, Daniel and Boppana, Siddharth and Merullo, Jack and Geiger, Atticus and Lewis, Owen and McGrath, Tom and Lubana, Ekdeep Singh},
  title         = {The Shape of Beliefs: Geometry, Dynamics, and Interventions along Representation Manifolds of Language Models' Posteriors},
  year          = {2026},
  eprint        = {2602.02315},
  archivePrefix = {arXiv},
  primaryClass  = {cs.CL}
}

@misc{balcells2026belief,
  author        = {Balcells, Daniel and Lee, Andrew Jun and Rastogi, Chirag and Riechers, Paul M. and Shai, Adam and Poncini, Xavier},
  title         = {Large Language Models Develop Belief State Geometry In-Context},
  year          = {2026},
  eprint        = {2609.17376},
  archivePrefix = {arXiv},
  primaryClass  = {cs.LG}
}

@inproceedings{shai2024transformers,
  author        = {Shai, Adam S. and Marzen, Sarah E. and Teixeira, Lucas and Oldenziel, Alexander Gietelink and Riechers, Paul M.},
  title         = {Transformers represent belief state geometry in their residual stream},
  booktitle     = {Advances in Neural Information Processing Systems},
  volume        = {37},
  pages         = {75012--75034},
  year          = {2024},
  eprint        = {2405.15943},
  archivePrefix = {arXiv},
  primaryClass  = {cs.LG}
}

@inproceedings{xie2022explanation,
  author        = {Xie, Sang Michael and Raghunathan, Aditi and Liang, Percy and Ma, Tengyu},
  title         = {An Explanation of In-context Learning as Implicit {Bayesian} Inference},
  booktitle     = {The Tenth International Conference on Learning Representations},
  year          = {2022},
  eprint        = {2111.02080},
  archivePrefix = {arXiv},
  primaryClass  = {cs.CL}
}

@article{griffin1992weighing,
  author        = {Griffin, Dale and Tversky, Amos},
  title         = {The Weighing of Evidence and the Determinants of Confidence},
  journal       = {Cognitive Psychology},
  volume        = {24},
  number        = {3},
  pages         = {411--435},
  year          = {1992},
  doi           = {10.1016/0010-0285(92)90013-R}
}

@inproceedings{falck2024is,
  author        = {Falck, Fabian and Wang, Ziyu and Holmes, Chris},
  title         = {Is In-Context Learning in Large Language Models {Bayesian}? {A} Martingale Perspective},
  booktitle     = {Proceedings of the 41st International Conference on Machine Learning},
  series        = {Proceedings of Machine Learning Research},
  volume        = {235},
  pages         = {12784--12805},
  year          = {2024},
  eprint        = {2406.00793},
  archivePrefix = {arXiv},
  primaryClass  = {stat.ML}
}

@misc{chlon2025llms,
  author        = {Chlon, Leon and Sheaib, Fatima and Khamis, Zein and Chlon, Maggie and El Zein, Mahdi and Awada, MarcAntonio M.},
  title         = {{LLMs} are {Bayesian}, In Expectation, Not in Realization},
  year          = {2025},
  eprint        = {2507.11768},
  archivePrefix = {arXiv},
  primaryClass  = {stat.ML}
}

@inproceedings{gupta2025enough,
  author        = {Gupta, Ritwik and Corona, Rodolfo and Ge, Jiaxin and Wang, Eric and Klein, Dan and Darrell, Trevor and Chan, David M.},
  title         = {Enough Coin Flips Can Make {LLM}s Act {B}ayesian},
  booktitle     = {Proceedings of the 63rd Annual Meeting of the Association for Computational Linguistics (Volume 1: Long Papers)},
  pages         = {7634--7655},
  year          = {2025},
  address       = {Vienna, Austria},
  publisher     = {Association for Computational Linguistics},
  doi           = {10.18653/v1/2025.acl-long.377},
  url           = {https://aclanthology.org/2025.acl-long.377/},
  eprint        = {2503.04722},
  archivePrefix = {arXiv},
  primaryClass  = {cs.CL}
}

@article{kumaran2026competing,
  author        = {Kumaran, Dharshan and Fleming, Stephen M. and Markeeva, Larisa and Heyward, Joe and Banino, Andrea and Mathur, Mrinal and Pascanu, Razvan and Osindero, Simon and De Martino, Benedetto and Veli{\v{c}}kovi{\'c}, Petar and Patraucean, Viorica},
  title         = {Competing Biases underlie Overconfidence and Underconfidence in {LLMs}},
  journal       = {Nature Machine Intelligence},
  volume        = {8},
  number        = {4},
  pages         = {614--627},
  year          = {2026},
  doi           = {10.1038/s42256-026-01217-9},
  eprint        = {2507.03120},
  archivePrefix = {arXiv},
  primaryClass  = {cs.LG}
}

@inproceedings{atwell2026basil,
  author        = {Atwell, Katherine and Heydari, Pedram and Sicilia, Anthony and Alikhani, Malihe},
  title         = {{BASIL}: {Bayesian} Assessment of Sycophancy in {LLMs}},
  booktitle     = {Proceedings of the 2026 ACM Conference on Fairness, Accountability, and Transparency},
  pages         = {6613--6642},
  year          = {2026},
  publisher     = {Association for Computing Machinery},
  doi           = {10.1145/3805689.3812404},
  eprint        = {2508.16846},
  archivePrefix = {arXiv},
  primaryClass  = {cs.AI}
}

@misc{smolin2026beliefs,
  author        = {Smolin, Alex and Wilder, Bryan},
  title         = {Beliefs and Behavior in Language Models},
  year          = {2026},
  eprint        = {2609.07943},
  archivePrefix = {arXiv},
  primaryClass  = {cs.AI}
}

@inproceedings{xu2026when,
  author        = {Xu, Haoming and Xu, Weihong and Li, Zongrui and Wang, Mengru and Yao, Yunzhi and Wu, Chiyu and Shang, Jin and Gong, Yu and Deng, Shumin},
  title         = {When Should Models Change Their Minds? {C}ontextual Belief Management in Large Language Models},
  booktitle     = {Proceedings of the 2026 Conference on Empirical Methods in Natural Language Processing},
  year          = {2026},
  eprint        = {2605.30219},
  archivePrefix = {arXiv},
  primaryClass  = {cs.AI},
  note          = {To appear}
}

@inproceedings{davidson2025common,
  author        = {Davidson, Guy and Gureckis, Todd M. and Lake, Brenden M. and Williams, Adina},
  title         = {Do different prompting methods yield a common task representation in language models?},
  booktitle     = {Advances in Neural Information Processing Systems},
  volume        = {38},
  pages         = {70782--70822},
  year          = {2025},
  eprint        = {2505.12075},
  archivePrefix = {arXiv},
  primaryClass  = {cs.CL}
}

@inproceedings{heyman2026steer,
  author        = {Heyman, Geert and Vandeputte, Frederik},
  title         = {Steer Like the {LLM}: Activation Steering that Mimics Prompting},
  booktitle     = {Proceedings of the 43rd International Conference on Machine Learning},
  series        = {Proceedings of Machine Learning Research},
  volume        = {306},
  pages         = {42968--42997},
  year          = {2026},
  eprint        = {2605.03907},
  archivePrefix = {arXiv},
  primaryClass  = {cs.CL}
}

@misc{cheng2026decomposing,
  author        = {Cheng, Fan L. and Kriegeskorte, Nikolaus},
  title         = {Decomposing how prompting steers behavior},
  year          = {2026},
  eprint        = {2606.03093},
  archivePrefix = {arXiv},
  primaryClass  = {cs.AI}
}

@inproceedings{todd2024function,
  author        = {Todd, Eric and Li, Millicent L. and Sharma, Arnab Sen and Mueller, Aaron and Wallace, Byron C. and Bau, David},
  title         = {Function Vectors in Large Language Models},
  booktitle     = {The Twelfth International Conference on Learning Representations},
  year          = {2024},
  eprint        = {2310.15213},
  archivePrefix = {arXiv},
  primaryClass  = {cs.CL}
}

@inproceedings{hendel2023incontext,
  author        = {Hendel, Roee and Geva, Mor and Globerson, Amir},
  title         = {In-Context Learning Creates Task Vectors},
  booktitle     = {Findings of the Association for Computational Linguistics: EMNLP 2023},
  pages         = {9318--9333},
  year          = {2023},
  address       = {Singapore},
  publisher     = {Association for Computational Linguistics},
  doi           = {10.18653/v1/2023.findings-emnlp.624},
  url           = {https://aclanthology.org/2023.findings-emnlp.624/},
  eprint        = {2310.15916},
  archivePrefix = {arXiv},
  primaryClass  = {cs.CL}
}

@article{lindsey2025introspective,
  author        = {Lindsey, Jack},
  title         = {Emergent Introspective Awareness in Large Language Models},
  journal       = {Transformer Circuits Thread},
  year          = {2025},
  url           = {https://transformer-circuits.pub/2025/introspection/index.html},
  eprint        = {2601.01828},
  archivePrefix = {arXiv},
  primaryClass  = {cs.CL}
}

@misc{ferrara2026owmi,
  author        = {Ferrara, Emilio},
  title         = {Open-Weight Masked Introspection: Measuring What Language Models Can Report About Their Own Computation},
  year          = {2026},
  eprint        = {2608.20569},
  archivePrefix = {arXiv},
  primaryClass  = {cs.AI}
}

@inproceedings{binder2025looking,
  author        = {Binder, Felix J and Chua, James and Korbak, Tomek and Sleight, Henry and Hughes, John and Long, Robert and Perez, Ethan and Turpin, Miles and Evans, Owain},
  title         = {Looking Inward: Language Models Can Learn About Themselves by Introspection},
  booktitle     = {The Thirteenth International Conference on Learning Representations},
  year          = {2025},
  eprint        = {2410.13787},
  archivePrefix = {arXiv},
  primaryClass  = {cs.CL}
}

@article{nerode1958linear,
  author        = {Nerode, Anil},
  title         = {Linear Automaton Transformations},
  journal       = {Proceedings of the American Mathematical Society},
  volume        = {9},
  number        = {4},
  pages         = {541--544},
  year          = {1958},
  doi           = {10.1090/S0002-9939-1958-0135681-9}
}

@article{shalizi2001computational,
  author        = {Shalizi, Cosma Rohilla and Crutchfield, James P.},
  title         = {Computational Mechanics: Pattern and Prediction, Structure and Simplicity},
  journal       = {Journal of Statistical Physics},
  volume        = {104},
  number        = {3--4},
  pages         = {817--879},
  year          = {2001},
  doi           = {10.1023/A:1010388907793}
}

@inproceedings{littman2002predictive,
  author        = {Littman, Michael L. and Sutton, Richard S. and Singh, Satinder},
  editor        = {Dietterich, Thomas G. and Becker, Suzanna and Ghahramani, Zoubin},
  title         = {Predictive Representations of State},
  booktitle     = {Advances in Neural Information Processing Systems 14},
  year          = {2002},
  publisher     = {MIT Press},
  url           = {https://proceedings.neurips.cc/paper_files/paper/2001/hash/1e4d36177d71bbb3558e43af9577d70e-Abstract.html}
}

@article{mondal2026implicit,
  author        = {Mondal, Manuel and Dolamic, Ljiljana and Bovet, G{\'e}r{\^o}me and Cudr{\'e}-Mauroux, Philippe and Audiffren, Julien},
  title         = {Implicit Probabilistic Reasoning Does Not Reflect Explicit Answers in Large Language Models},
  journal       = {Transactions on Machine Learning Research},
  year          = {2026},
  url           = {https://openreview.net/forum?id=HaaAY4ZXPa},
  eprint        = {2406.14986},
  archivePrefix = {arXiv},
  primaryClass  = {cs.AI}
}

@inproceedings{treutlein2024connecting,
  author        = {Treutlein, Johannes and Choi, Dami and Betley, Jan and Marks, Samuel and Anil, Cem and Grosse, Roger and Evans, Owain},
  title         = {Connecting the Dots: {LLMs} can Infer and Verbalize Latent Structure from Disparate Training Data},
  booktitle     = {Advances in Neural Information Processing Systems},
  volume        = {37},
  pages         = {140667--140730},
  year          = {2024},
  eprint        = {2406.14546},
  archivePrefix = {arXiv},
  primaryClass  = {cs.CL}
}

@inproceedings{jang2026instructiontuned,
  author        = {Jang, Chaemin and Kim, Jihee and Lee, Dongman},
  title         = {Instruction-Tuned Language Models Cannot Sample from Distributions They Can Describe},
  booktitle     = {Findings of the Association for Computational Linguistics: EMNLP 2026},
  pages         = {48757--48777},
  year          = {2026},
  address       = {Budapest, Hungary},
  publisher     = {Association for Computational Linguistics},
  doi           = {10.18653/2026.findings-emnlp.2278},
  url           = {https://aclanthology.org/2026.findings-emnlp.2278/},
  eprint        = {2607.25292},
  archivePrefix = {arXiv},
  primaryClass  = {cs.AI}
}

@inproceedings{meister2025benchmarking,
  author        = {Meister, Nicole and Guestrin, Carlos and Hashimoto, Tatsunori},
  title         = {Benchmarking Distributional Alignment of Large Language Models},
  booktitle     = {Proceedings of the 2025 Conference of the Nations of the Americas Chapter of the Association for Computational Linguistics: Human Language Technologies (Volume 1: Long Papers)},
  pages         = {24--49},
  year          = {2025},
  address       = {Albuquerque, New Mexico},
  publisher     = {Association for Computational Linguistics},
  doi           = {10.18653/v1/2025.naacl-long.2},
  url           = {https://aclanthology.org/2025.naacl-long.2/},
  eprint        = {2411.05403},
  archivePrefix = {arXiv},
  primaryClass  = {cs.CL}
}

@inproceedings{gu2026illusion,
  author        = {Gu, Xiangming and De, Soham and Titsias, Michalis and Markeeva, Larisa and Veli{\v{c}}kovi{\'c}, Petar and Pascanu, Razvan},
  title         = {The Illusion of Stochasticity in {LLMs}},
  booktitle     = {Third Conference on Language Modeling},
  year          = {2026},
  eprint        = {2604.06543},
  archivePrefix = {arXiv},
  primaryClass  = {cs.CL}
}

@misc{lei2026elicitation,
  author        = {Lei, Ge and Cooper, Samuel J.},
  title         = {Elicitation Matters: How Prompts and Query Protocols Shape {LLM} Surrogates under Sparse Observations},
  year          = {2026},
  eprint        = {2605.04764},
  archivePrefix = {arXiv},
  primaryClass  = {cs.CL}
}

@inproceedings{cekinmez2026query,
  author        = {Cekinmez, Jasin and Wu, Addison J. and Griffiths, Thomas L.},
  title         = {Query Timing Produces Opposite Positional Biases Between {LLMs} and Humans},
  booktitle     = {Findings of the Association for Computational Linguistics: EMNLP 2026},
  pages         = {28741--28758},
  year          = {2026},
  address       = {Budapest, Hungary},
  publisher     = {Association for Computational Linguistics},
  doi           = {10.18653/2026.findings-emnlp.1354},
  url           = {https://aclanthology.org/2026.findings-emnlp.1354/}
}

@misc{slocum2025believe,
  author        = {Slocum, Stewart and Minder, Julian and Dumas, Cl{\'e}ment and Sleight, Henry and Greenblatt, Ryan and Marks, Samuel and Wang, Rowan},
  title         = {Believe It or Not: How Deeply do {LLMs} Believe Implanted Facts?},
  year          = {2025},
  eprint        = {2510.17941},
  archivePrefix = {arXiv},
  primaryClass  = {cs.CL}
}

@misc{shih2026when,
  author        = {Shih, Benjamin and Winnicki, John and Darve, Eric},
  title         = {When Does Activation Steering Change What a Model Computes From?},
  year          = {2026},
  eprint        = {2606.29522},
  archivePrefix = {arXiv},
  primaryClass  = {cs.LG}
}

@inproceedings{mishra2026steered,
  author        = {Mishra, Aayush and Khashabi, Daniel and Liu, Anqi},
  title         = {Steered {LLM} Activations are Non-Surjective},
  booktitle     = {ICLR 2026 Workshop on Scientific Methods for Understanding Deep Learning},
  year          = {2026},
  eprint        = {2604.09839},
  archivePrefix = {arXiv},
  primaryClass  = {cs.AI}
}

@misc{kang2026promptactivation,
  author        = {Kang, Diancheng and Liu, Zheyuan and Ma, Ningshan and Huang, Yue and Tan, Zhaoxuan and Jiang, Meng},
  title         = {Prompt-Activation Duality: Improving Activation Steering via Attention-Level Interventions},
  year          = {2026},
  eprint        = {2605.10664},
  archivePrefix = {arXiv},
  primaryClass  = {cs.CL}
}

@inproceedings{du2026its,
  author        = {Du, Kevin and K{\"u}mpel, Clara and Wastl, Michelle and Warstadt, Alex},
  title         = {It{'}s Not What You Say, It{'}s How You Say It: Evaluating {LLM} Responses to Expressions of Belief},
  booktitle     = {Proceedings of the 64th Annual Meeting of the {A}ssociation for {C}omputational {L}inguistics (Volume 1: Long Papers)},
  pages         = {3137--3151},
  year          = {2026},
  address       = {San Diego, California, United States},
  publisher     = {Association for Computational Linguistics},
  doi           = {10.18653/v1/2026.acl-long.142},
  url           = {https://aclanthology.org/2026.acl-long.142/},
  eprint        = {2607.18232},
  archivePrefix = {arXiv},
  primaryClass  = {cs.CL}
}

@inproceedings{hu2023prompting,
  author        = {Hu, Jennifer and Levy, Roger},
  title         = {Prompting is not a substitute for probability measurements in large language models},
  booktitle     = {Proceedings of the 2023 Conference on Empirical Methods in Natural Language Processing},
  pages         = {5040--5060},
  year          = {2023},
  address       = {Singapore},
  publisher     = {Association for Computational Linguistics},
  doi           = {10.18653/v1/2023.emnlp-main.306},
  url           = {https://aclanthology.org/2023.emnlp-main.306/},
  eprint        = {2305.13264},
  archivePrefix = {arXiv},
  primaryClass  = {cs.CL}
}

@inproceedings{du2024context,
  author        = {Du, Kevin and Sn{\ae}bjarnarson, V{\'e}steinn and Stoehr, Niklas and White, Jennifer C. and Schein, Aaron and Cotterell, Ryan},
  title         = {Context versus Prior Knowledge in Language Models},
  booktitle     = {Proceedings of the 62nd Annual Meeting of the Association for Computational Linguistics (Volume 1: Long Papers)},
  pages         = {13211--13235},
  year          = {2024},
  address       = {Bangkok, Thailand},
  publisher     = {Association for Computational Linguistics},
  doi           = {10.18653/v1/2024.acl-long.714},
  url           = {https://aclanthology.org/2024.acl-long.714/},
  eprint        = {2404.04633},
  archivePrefix = {arXiv},
  primaryClass  = {cs.CL}
}

@misc{grattafiori2024llama3,
  author        = {Grattafiori, Aaron and others},
  title         = {The {Llama 3} Herd of Models},
  year          = {2024},
  eprint        = {2407.21783},
  archivePrefix = {arXiv},
  primaryClass  = {cs.AI}
}

@misc{yang2025qwen3,
  author        = {Yang, An and others},
  title         = {{Qwen3} Technical Report},
  year          = {2025},
  eprint        = {2505.09388},
  archivePrefix = {arXiv},
  primaryClass  = {cs.CL}
}

@misc{gemmateam2025gemma3,
  author        = {{Gemma Team}},
  title         = {{Gemma 3} Technical Report},
  year          = {2025},
  eprint        = {2503.19786},
  archivePrefix = {arXiv},
  primaryClass  = {cs.CL}
}

@article{cohen2024evaluating,
  author        = {Cohen, Roi and Biran, Eden and Yoran, Ori and Globerson, Amir and Geva, Mor},
  title         = {Evaluating the Ripple Effects of Knowledge Editing in Language Models},
  journal       = {Transactions of the Association for Computational Linguistics},
  volume        = {12},
  pages         = {283--298},
  year          = {2024},
  publisher     = {MIT Press},
  doi           = {10.1162/tacl_a_00644},
  url           = {https://aclanthology.org/2024.tacl-1.16/},
  eprint        = {2307.12976},
  archivePrefix = {arXiv},
  primaryClass  = {cs.CL}
}

@misc{qwenteam2026qwen38,
  author        = {{Qwen Team}},
  title         = {{Qwen3.8-Max}: A New Bar for Coding and Cowork},
  year          = {2026},
  month         = aug,
  howpublished  = {Blog post, the reference given by the model card of {Qwen3.8-27B}},
  url           = {https://qwen.ai/blog?id=qwen3.8}
}

@misc{qwenteam2026qwen35,
  author        = {{Qwen Team}},
  title         = {{Qwen3.5}: Towards Native Multimodal Agents},
  year          = {2026},
  month         = feb,
  howpublished  = {Blog post, the reference given by the model cards of the {Qwen3.5} models},
  url           = {https://qwen.ai/blog?id=qwen3.5}
}

@misc{openai2026gpt6sol,
  author        = {{OpenAI}},
  title         = {Introducing {GPT-6 Sol} and {Luna}},
  year          = {2026},
  month         = sep,
  howpublished  = {Release post},
  url           = {https://openai.com/index/introducing-gpt-6-sol-and-luna/}
}

@misc{google2026gemini4argon,
  author        = {{Google}},
  title         = {{Gemini 4 Argon}},
  year          = {2026},
  howpublished  = {Gemini API documentation},
  url           = {https://ai.google.dev/gemini-api/docs/models/gemini-4-argon}
}

@misc{anthropic2026opus55,
  author        = {{Anthropic}},
  title         = {Introducing {Claude Opus 5.5}},
  year          = {2026},
  month         = sep,
  howpublished  = {Release post},
  url           = {https://www.anthropic.com/news/claude-opus-5-5}
}

\appendix
\renewcommand{\thefigure}{A\arabic{figure}}
\setcounter{figure}{0}
\renewcommand{\thetable}{A\arabic{table}}
\setcounter{table}{0}

\renewcommand{\topfraction}{0.9}
\renewcommand{\dbltopfraction}{0.9}
\renewcommand{\textfraction}{0.07}
\renewcommand{\floatpagefraction}{0.75}
\renewcommand{\dblfloatpagefraction}{0.75}
\setcounter{topnumber}{3}
\setcounter{dbltopnumber}{2}
\setcounter{totalnumber}{4}

\section{Three notions of an internal state}
\label{app:philosophy}

We distinguish three operational notions of an internal state. A \emph{report} notion identifies a state by what the model says about it, which is our estimate. A \emph{functional} notion identifies it by what it does next and how it revises, which is our continuation together with the forked futures. An \emph{access} notion asks whether the model can predict its own revision better than an observer who sees only the text, which we leave open. Each notion induces a partition of prepared states, and the relations between the partitions are empirical. In the smaller models the report and functional partitions cross: a told rate separates states only under the first, and a shown record only under the second. In the larger models the two partitions draw closer: the estimates of Qwen3-32B and Qwen3.8-27B follow a shown record as well, and Qwen3.8-27B's continuation follows a told rate.

\section{Bayes derivations}
\label{app:bayes}

\paragraph{Records and summaries.}
With a $\mathrm{Beta}(\alpha_0, \beta_0)$ prior of strength $w_0 = \alpha_0 + \beta_0$, after $n$ observations with a proportion $\pi$ of heads and a further $m$ with a proportion $\kappa$, the posterior predictive probability of heads is
\begin{equation}
p_{\mathrm{post}} = \frac{\alpha_0 + n\pi + m\kappa}{w_0 + n + m}.
\end{equation}
Read as a function of $\pi$ and $\kappa$ it is linear, with $c = \alpha_0/(w_0+n+m)$, $a = n/(w_0+n+m)$ and $S = m/(w_0+n+m)$, so that $w = m\,a/S = n$ and the remainder $1 - a - S = w_0/(w_0+n+m)$ is the share of the readout that returns to the prior mean. The retention of the old content after the segment is $R = a/a_0 = (w_0+n)/(w_0+n+m)$. With no record, $S = m/(w_0+m)$, which gives $w_0 = m(1-S)/S$. For the uniform prior, $w_0 = 2$ and the predictive is $(k+1)/(n+2)$. A summary that states the counts $(k, n)$ is the sufficient statistic of the record and gives the same posterior. The same observer gives one hundred times the predictive as its estimate, so for a Bayesian observer the slopes of the estimate coincide with those of the continuation.

\paragraph{A told rate.}
A statement that the probability is $s$ has a Bayesian reference only together with a model of the source. Taking the statement at its word is the source model with reliability one: the observer holds $s$ whatever it then observes, $a = 1$ and $S = 0$. An observer who holds that the source is right with probability $r$, and otherwise uninformative, predicts $r s + (1-r)/2$ before any observation, so its content slope is $r$, and afterwards keeps a posterior over the two cases. After $m$ new observations with $k$ heads, with $L = s^k (1-s)^{m-k}$ the likelihood of the evidence under the stated rate and $B$ the beta function, the probability that the source is right is
\begin{equation}
\rho = \frac{r L}{r L + (1-r)\, B(k+1,\, m-k+1)},
\end{equation}
and the predictive is $\rho\, s + (1-\rho)\,(k+1)/(m+2)$. The reference keeps the statement and then drops it: $\rho$ stays near one while the evidence is compatible with $s$ and falls to near zero once it is not, faster for a smaller $r$. In the two-slope regression the drop appears as a lower retention of the statement at a sensitivity to the evidence that comes from the second term. Appendix~\ref{app:rates} reports which parts of this reference each model follows. An instruction to assume $s$ is the limit $r = 1$ under any evidence.

\paragraph{Paired states.}
Two states $A$ and $B$ that differ only in content, such as the two signs of a one-shot write, give the paired form $a = \big(p_{\mathrm{post}}(A) - p_{\mathrm{post}}(B)\big)/(\pi_A - \pi_B)$. When the content scale is taken as the difference between their current readouts, this slope is the retention $R$ of the installed content after the future, and $S$ is estimated from the same two states. Where $w$ differs across $m$, we report $w(m)$.

\section{Estimator details, bootstrap and integer readout}
\label{app:estimators}

Within a cell, the readout of every state after every future, the probability of H for a continuation and the expected count divided by one hundred for an estimate, is regressed on $(\pi, \kappa)$ by least squares, and $(a, S, w)$ follow from Eqs.~\ref{eq:two-slope} and \ref{eq:exchange-rate}. The unit of resampling is the family, the independently generated record with its wording and its order of futures. A cell of the channel map holds 180 shown records per length, 24 summaries per stated count, 80 told rates and 15 instructions, and a cell of the weight experiments holds 60 records per length, 24 summaries, 48 told rates and 15 instructions. An instruction has one state per level and strength, and its intervals resample the orders of the new segment. We draw 2,000 bootstrap resamples of families with replacement, recompute the three quantities and report percentile intervals.

\paragraph{Unrevised weights and equivalence.}
When the lower bound of $S$ reaches zero, $w$ diverges, and we report the upper bound of $S$ and the corresponding lower bound of $w$ instead of a point estimate. A weight is reported as not revised, and printed $\infty$ in the tables, where the interval of $S$ includes zero or the weight exceeds a thousand observations. Equivalence between two routes or two layouts is tested by two one-sided tests with a tolerance equal to the model's wording noise floor, the mean absolute difference between the four wordings of the same state. For a continuation the tolerance is that difference in probability units divided by the 0.5 range of the content proportion, the slope it amounts to (0.07 for Llama-3.1-8B's document-frame continuation, 0.16 and 0.10 for the assistant-frame continuations of Qwen3-8B and Qwen3.8-27B), and for an estimate it is the difference in expected count. The steering comparison of Section~\ref{sec:writes} uses the within-family standard deviation of the same wordings in logit units (0.15 for Llama-3.1-8B). Paired contrasts, such as sixteen against four written positions, a reliable against an unreliable source, or one layout against another, resample the same families on both sides.

\paragraph{Answer mass.}
The mass of a readout is the probability assigned to the admissible answers, H and T for a continuation and the integers from 0 to 100 for an estimate. A cell whose median mass is below 0.3 is drawn hollow, marked n.r.\ in the tables, and not interpreted.

\paragraph{Channels.}
The weight experiments read one continuation and one estimate per model, chosen by a rule fixed before them. A model's \emph{sampling channel} is the continuation whose slope on a record of 64 observations exceeds 0.3, whose median answer mass is at least 0.3, and whose response to a new segment of sixteen observations is positive, with the document frame tried first. Its \emph{report channel} is the estimate in the assistant frame. Llama-3.1-8B's sampling channel is its document-frame continuation. Qwen3-8B, whose document-frame continuation does not follow the record, Gemma-3-27B and Qwen3.8-27B have theirs in the assistant frame. Gemma-3-12B has none, since in both frames its continuation moves against a new segment (Appendix~\ref{app:controls}). Qwen3-32B has none either. Its document-frame continuation follows a record of 64 at 0.10 and its assistant-frame continuation at 0.24, the latter because it sits near certainty of heads (a probability of H between 0.87 and 0.99) while still responding to a new segment with a slope of 0.86. Gemma-3-12B and Qwen3-32B therefore enter the weight experiments through their report channel alone.

Gemma-3-27B's continuation is close to deterministic, with a median probability of H near zero for records of 16 or more, so its sampling-channel weights and its activation writes are reported without interpretation. Qwen3.8-27B's continuation is steeper than the record (a slope of 1.38 at $n = 64$) without collapsing to the mode. Two groups of readouts fail the mass criterion: every document-frame readout of Qwen3.8-27B, where the model ends the turn, and Gemma-3-27B's expected-count estimate after a shown record or a summary of 8, which opens with a blank line. The other readouts that fall below the criterion in some cells are marked where they occur.

\paragraph{Integer readout.}
Llama-3.1-8B tokenizes the integers from 0 to 100 as single tokens, and its estimate is their distribution at the first position after the query, with a check at each integer that the token completes the number (a leaf check against continuations such as a further digit). The other five models tokenize integers digit by digit, and the readout is a tree: each digit position is forked from the cached state, a branch is followed only while its first digit carries at least $10^{-4}$ of the mass, and a branch terminates when the next token is a terminator or a period followed by nothing or by a zero, which are read as the integer. Branches that continue into an operator, a decimal fraction or further text are excluded from the mass. The same rule is applied to every model, and the mass it leaves is reported with every cell.

\section{Control experiments}
\label{app:controls}

\begin{figure*}[t]
  \centering
  \includegraphics[width=\textwidth]{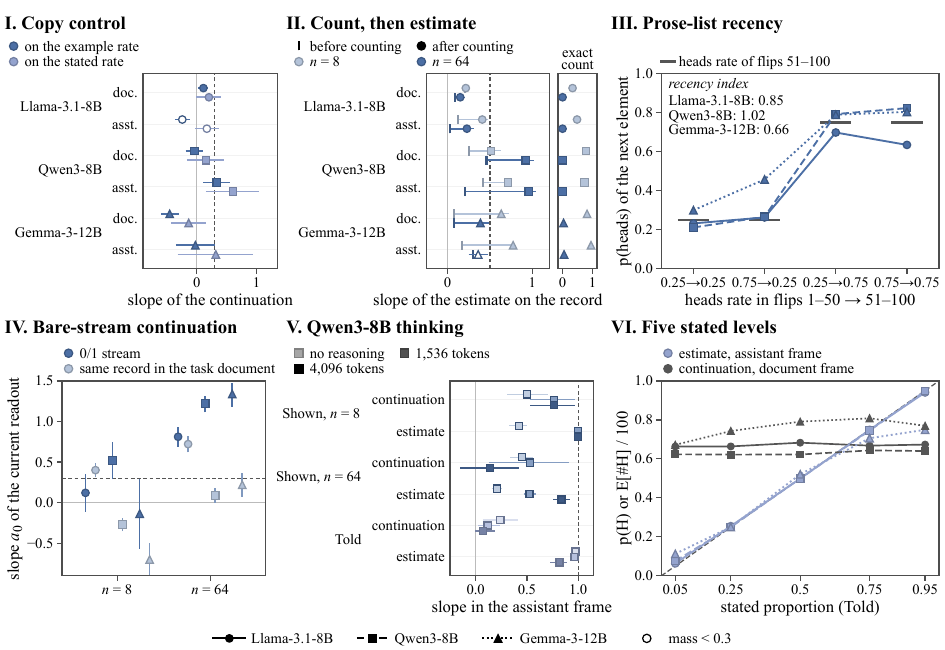}
  \caption{\textbf{Control experiments for the channel map.} I. Copy control. II. Count, then estimate (with the exact-count accuracy). III. Prose-list recency. IV. Bare-stream continuation. V. Qwen3-8B in thinking mode, 1,536 against 4,096 reasoning tokens, with the non-thinking readout as baseline. VI. The five stated levels. Llama-3.1-8B, Qwen3-8B and Gemma-3-12B. In II, Gemma-3-12B's assistant-frame point after counting at $n = 64$ is unreadable (mass 0.04).}
  \label{fig:controls}
\end{figure*}

Each control changes one part of the document around the channel map of Figure~\ref{fig:channels} or replaces it (Figure~\ref{fig:controls}, Appendix~\ref{app:prompts}). The prose list, the bare stream and extended reasoning are run on the three smaller models, and the other controls on all six.

\paragraph{Copy control.}
A statement is followed by four example outcomes whose proportion differs from the stated rate by one half, and the continuation is regressed jointly on the example and the stated proportions. The document-frame continuation follows the examples with slopes of 0.11 in Llama-3.1-8B and $-0.45$ in Gemma-3-12B, and Qwen3-8B's assistant-frame continuation follows the examples and the statement together (0.33 and 0.60). Four outcomes beside a statement thus move the continuation far less than a record does (0.74 for a record of 64 in Llama-3.1-8B), and in Qwen3-8B the statement enters as well.

\paragraph{Count, then estimate.}
On shown records the model is first asked for the number of H in the record, its answer is committed, and the expected-count question follows. We report the accuracy of the count, the slope of the estimate on the true proportion before and after counting, and the slope on the model's own count. Llama-3.1-8B follows its own count with slopes of 0.19 to 0.71 and the true proportion with at most 0.41. After counting, Qwen3-8B's estimate follows a record of 64 at 0.96. Gemma-3-12B at $n = 64$ writes an arithmetic expression where the estimate should be. At $n = 64$ Qwen3-32B, Gemma-3-27B and Qwen3.8-27B count correctly in 5, 13 and 7\% of the states, and the estimate after counting follows the record at 0.72 in Qwen3-32B (0.60 before, on the states of this control) and at 0.68 in Qwen3.8-27B (0.97 before).

\paragraph{New evidence stated as a count.}
The new segment states the numbers of H and T in a further $m$ tosses in place of the outcomes themselves. In the smaller models sixteen further observations written out move the estimate with a slope of at most 0.26, and stated as a count with a slope of up to 0.51, at every record length, while a count hardly moves Llama-3.1-8B's document-frame continuation. The larger models keep the order: the slope on the new evidence is 0.45 to 0.51 as a count against 0.22 to 0.27 written out in Qwen3-32B, and 0.80 to 0.89 against 0.23 to 0.47 in Qwen3.8-27B. Gemma-3-27B's estimate after a count segment fails the mass criterion. Against a count segment the weight of a summary in the smaller models grows 2.5 to 4.0 times between a stated 8 and a stated 64, in the first wording of the segment from 19 [8.0, 32] to 72 [43, 107] observations in Llama-3.1-8B, from 12 [7.8, 16] to 45 [29, 69] in Qwen3-8B and from 12 [6.4, 19] to 47 [27, 86] in Gemma-3-12B. Against written-out observations it grows 0.7 to 3.5 times (Table~\ref{tab:queries}).

\paragraph{Prose list.}
One hundred flips are written as \prompt{I flipped a coin and it landed on heads, then on tails, \ldots} with the first and last fifty at different rates, and the next element is read. The design follows \citet{gupta2025enough}, whose sequence of one hundred flips changes its rate at the midpoint. The recency index, the slope on the last fifty minus the slope on the first fifty, is 0.85, 1.02 and 0.66 in Llama-3.1-8B, Qwen3-8B and Gemma-3-12B. It measures the same recency as the discount factor that \citet{gupta2025enough} fit to the earlier counts before each new observation, which lies below one in every model they study.

\begin{figure}[t]
  \centering
  \includegraphics[width=\columnwidth]{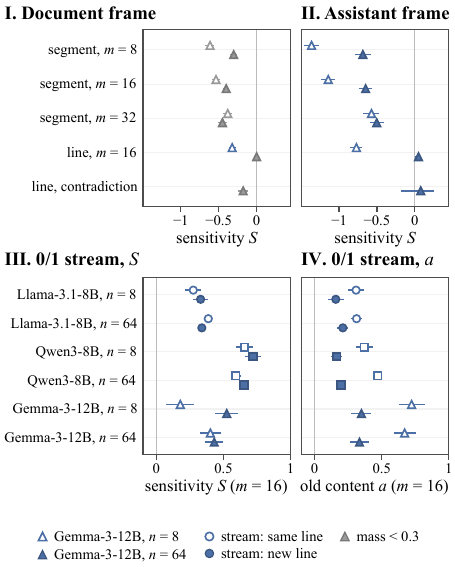}
  \caption{\textbf{Layout dependence in the bare stream and Gemma-3-12B's response to new evidence.} I, II. Gemma-3-12B: the slope $S$ on the new evidence under each layout (segment: a new segment; line: the extended line) in the document and assistant frames, at $n = 8$ (hollow) and 64 (filled). III, IV. Bare 0/1 stream: $S$ and $a$ for the three smaller models with the new evidence on the same line (hollow) or on a new line (filled).}
  \label{fig:exploratory}
\end{figure}

\paragraph{Bare stream.}
The same records are written as a 0/1 stream and read as the next symbol. A record of 64 outcomes is continued with slopes of 0.81, 1.22 and 1.34 in the three smaller models. Under both layouts of new evidence the slope on the new evidence is positive in each of them, with no primacy on the same line and a lower weight of the record after a line break (Figure~\ref{fig:exploratory}).

\paragraph{Extended reasoning.}
Qwen3-8B is read in thinking mode with 1,536 and 4,096 reasoning tokens, the two assistant-frame readouts prefilled after the reasoning block. After the block the estimate follows a record of 64 at 0.84 and a told rate somewhat less than without reasoning (0.78 against 0.99), and the continuation no longer follows the record.

\paragraph{Five stated levels and a one-token decision check.}
The stated routes are read at 0.05 and 0.95 as well as at the three central levels. A one-token decision check reads the assistant's unprefilled H or T answer. Under an instruction Qwen3-8B's probability of H moves from near zero to near one across the stated levels, Gemma-3-12B decides only under the strongest instruction, and Llama-3.1-8B's answer follows the stated level with a slope of about 0.5 without collapsing to the mode. The decision check thus answers from a statement, as the estimate does.

\paragraph{Gemma-3-12B's response to new evidence.}
Under the new-segment layout Gemma-3-12B's continuation moves against the new evidence at every segment length in both frames, and under the extended line and the strong contradiction at $n = 64$ it is near zero (Figure~\ref{fig:exploratory}), so Gemma-3-12B has no weight in the document format. In a bare stream its response is positive.

\paragraph{Segmented records.}
The same 64 observations are written as $g \in \{1, 2, 4, 8\}$ segments, each headed \prompt{Observations of this coin, session $i$ of $g$, in order:}, and read in the model's sampling channel before and after a new segment of 8, 16 or 32 observations, with the same families at every $g$. Under a new segment of sixteen, Llama-3.1-8B's record is worth 10.6 observations at $g = 1$ and 18.8, 17.9 and 17.3 at $g = 2$, 4 and 8, the last three equal to the unsegmented record of Section~\ref{sec:weight}. The one-segment heading alone lowers the weight. In Qwen3-8B the weight falls with $g$ (4.3, 1.7 and about zero from $g = 4$ on) as the slope on the new segment rises to about one, while the current continuation follows the record more closely at $g = 8$ than at $g = 1$ (0.98 against 0.44). In Qwen3.8-27B the weight also falls with $g$, from 28 observations at one segment to 9 at eight. Gemma-3-12B's current continuations move further against the record when it is split.

\paragraph{Simulate, then sample.}
Before the continuation is read, the model is asked to write sixteen outcomes of the coin (\prompt{Simulated outcomes for this coin, in order:}), sampled at temperature one and restricted to H and T, four samples per state, and the continuation is read after the written outcomes. We report the slope of the written proportion on the content of the state, the slope of the continuation on the content before and after the simulation, and, within states, the slope of the continuation on the written proportion. Llama-3.1-8B's written outcomes ignore a told rate (0.02) and follow a shown record (0.50), and its continuation follows what it wrote on told states (0.79 within states). Qwen3-8B's written outcomes follow instructed and summarized content (0.64 and 0.90) and a told rate in part (0.30). Its continuation after the simulation carries the instructed and summarized content (0.93 and 1.03, against 0.10 and 0.27 before) and the told rate little more than before (0.46 against 0.38). Gemma-3-12B's continuation runs against the outcomes it wrote ($-1.17$ within told states). Qwen3.8-27B's continuation follows what it wrote (1.05 within told states), and Qwen3-32B's assistant-frame continuation does so at 0.26. No simulation collapses to the mode.

After a shown record the written outcomes are mostly a copy: Qwen3-8B repeats the record's first sixteen outcomes in 217 of 240 samples and Gemma-3-12B in all 240 of its assistant frame (Llama-3.1-8B in 20), and the three larger models repeat the record's opening as well. The slope on shown content (0.87 in Qwen3-8B) thus measures copying, and the stated routes carry the test.

\section{Transplanted states}
\label{app:transplant}

\begin{figure}[t]
  \centering
  \includegraphics[width=\columnwidth]{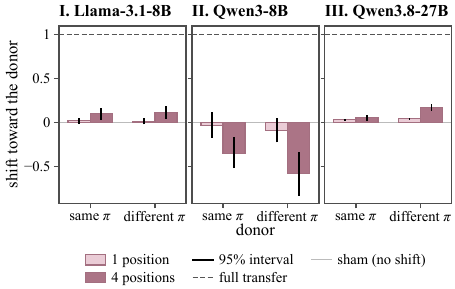}
  \caption{\textbf{A copied frontier moves the recipient little or not at all toward the donor.} Shift of the recipient toward the donor, relative to sham, for same-$\pi$ and different-$\pi$ donors with one and four copied positions. I. Llama-3.1-8B. II. Qwen3-8B. III. Qwen3.8-27B. The shift quoted in the text is the difference between the two donor conditions. Family-bootstrap 95\% intervals.}
  \label{fig:transplant}
\end{figure}

A \Transplanted{} state copies the residual stream at the last one or four positions of a \Shown{} state with 64 observations into the same positions of a \Shown{} state with 8, over a band of middle layers, and is judged against sham and same-proportion donors. In Llama-3.1-8B (20 pairs per proportion) the recipient moves toward the donor by $-0.02$ to $0.05$ of the natural content difference, with every interval including zero, and copying four positions erases about half of the recipient's own content (Figure~\ref{fig:transplant}). No detectable part of the long record travels with its frontier, and copying the frontier disturbs the recipient's own content. In Qwen3-8B the recipient does not move toward the donor either: the four shifts lie between $-0.33$ and 0.01, and the one interval that excludes zero lies below it. In Qwen3.8-27B a small part of the content travels with four frontier positions: the recipient moves toward the donor by 0.10 to 0.14 of the content difference, with intervals above zero, and by 0.02 or less when one position is copied.

\paragraph{The recipient's weight.}
The copied frontier gives the recipient none of the donor's weight. In Qwen3.8-27B the recipient of a donor with its own proportion is worth 2.1 [1.5, 2.6] observations after one copied position and 1.8 [1.2, 2.3] after four, against 2.0 [1.4, 2.6] under sham and 21.3 [19.8, 22.9] for a record of 64. The transplant closes 0.003 [$-0.012$, 0.018] and $-0.011$ [$-0.040$, 0.016] of that gap, and the recipient's slopes on its record and on the new segment are equivalent to sham within the wording noise floor. In Llama-3.1-8B the point estimates are 6.3 and 5.8 observations, against 6.5 [4.8, 8.3] under sham and 17.2 [15.0, 19.5] for the long record (point estimates: the readouts of this run were not retained). The content that four copied positions remove from its current continuation is back after the segment, where the slope on the record is 0.19 against 0.21 under sham. Qwen3-8B offers no test of weight, since its record of 64 is worth no more than its record of 8 (5.9 [4.7, 7.1] against 7.2 [4.9, 9.7]).

\section{Activation writes in detail}
\label{app:writes}

\paragraph{One-shot writes.}
The mean-difference direction between records with 75\% and 25\% heads is added once, with either sign, to the last four or sixteen positions of a \Shown{} record of sixteen observations, at a quarter or at a half of the depth, and the write is read in the model's sampling channel (Table~\ref{tab:writes}). In Llama-3.1-8B the write installs 92 to 109\% of the natural content difference and retains at most 0.12 of it after sixteen new observations, where the natural record retains 0.42 (Section~\ref{sec:writes}). In Qwen3-8B it installs 89 to 154\% and retains 0.07 to 0.24, where the natural record retains 0.43. Three of its four configurations retain less than half of what the natural record retains, and the fourth, four positions at half the depth, retains 0.24 [0.20, 0.28]. In Qwen3.8-27B the write installs 101 and 103\% of the natural content difference with sixteen positions and 31 and 50\% with four. After sixteen observations it retains 0.34 to 0.61 of its content where the natural record retains 0.29, and it stays above the natural record after eight and thirty-two.

\begin{figure}[t]
  \centering
  \includegraphics[width=\columnwidth]{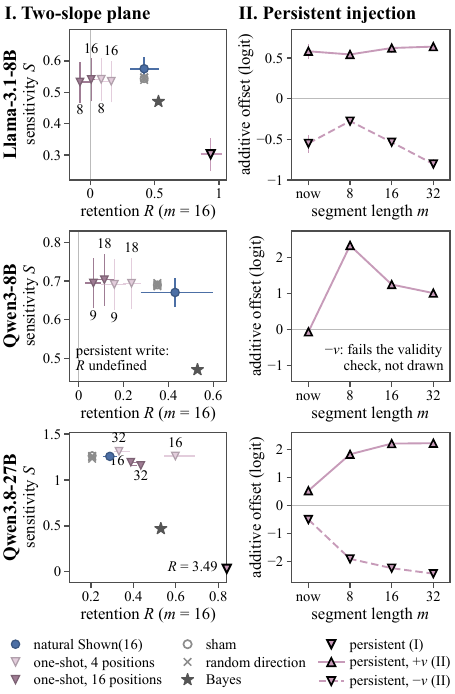}
  \caption{\textbf{Activation writes in Llama-3.1-8B, Qwen3-8B and Qwen3.8-27B.} One row per model. I. Sensitivity to a new segment of sixteen observations against the retention of the old content, for the natural record, the one-shot writes (numbers: the layer of the write) and the vector kept in place, which lies beyond the axis in Qwen3.8-27B and has no defined retention in Qwen3-8B. II. The vector kept in place at every position, as an offset on the log-odds against sham, by the number of new observations. In Qwen3-8B its negative sign fails the validity check of a write and is not drawn. Family-bootstrap 95\% intervals.}
  \label{fig:writes-detail}
\end{figure}

In all three models the sensitivity to the segment is unchanged (Figure~\ref{fig:writes-detail}, I). The write reaches Qwen3.8-27B's estimate: between the two signs of the vector the estimate differs by 0.17 to 0.18 with sixteen positions and by 0.07 to 0.08 with four, against 0.33 between the natural records, while Llama-3.1-8B's estimate differs by at most 0.005 and Qwen3-8B's by at most 0.05.

\paragraph{Persistent writes.}
Kept in place at every position, the vector is an offset on the log-odds that damps the response to evidence, constant in Llama-3.1-8B and growing with the number of new observations in Qwen3.8-27B (Figure~\ref{fig:writes-detail}, II). In Qwen3-8B the vector kept in place leaves the current continuation where it was and shifts it by 1.0 to 2.3 on the log-odds once new observations arrive, while the sensitivity to sixteen observations falls from 0.67 to 0.10. Its negative sign fails the validity check of a write (Appendix~\ref{app:prompts}), and with no content installed in the current continuation its retention is undefined.

\paragraph{The other three models.}
The other three models have no continuation in which a write can be judged. In Gemma-3-27B the natural record of sixteen has no content in the continuation, so the installed fraction and the natural retention are undefined. In Qwen3-32B the assistant-frame continuation gives heads with a probability of 0.91 to 0.98 whatever the record, and the write installs at most 0.16 of the small difference that remains. In Gemma-3-12B the assistant-frame continuation runs against the record, giving heads with a probability of 0.62 after a record with 25\% heads and 0.43 after one with 75\%, and it moves against new observations as well, so the retention of a write has no reading there.

\begin{table*}[t]
  \centering
  \footnotesize
  \caption{\textbf{One-shot writes against the natural record of sixteen, in the three models whose sampling channel carries the record.} Installed fraction of the natural content difference, retention $R$ after 8, 16 and 32 new observations, sensitivity $S$ to a segment of sixteen, retention after a neutral passage in place of the segment, and the difference of the estimate between the two signs of the vector (for the natural record, between records with 75\% and 25\% heads). K: positions written. Family-bootstrap 95\% intervals.}
  \label{tab:writes}
  \setlength{\tabcolsep}{3.5pt}
  \providecommand{\cint}[1]{{\fontsize{7}{8}\selectfont #1}}
\begin{threeparttable}
\begin{tabular}{@{}lccccc@{}}
  \toprule
  & \multicolumn{2}{c}{Quarter depth} & \multicolumn{2}{c}{Half depth} & Natural \\
  \cmidrule(lr){2-3} \cmidrule(lr){4-5}
  & K4 & K16 & K4 & K16 & record \\
  \midrule
  \multicolumn{6}{@{}l}{\textit{Llama-3.1-8B}, layers 8 and 16, document-frame continuation} \\
  \quad Installed fraction & 0.97 \cint{[0.82, 1.14]} & 0.92 \cint{[0.78, 1.10]} & 0.97 \cint{[0.80, 1.16]} & 1.09 \cint{[0.95, 1.29]} & 1 \\
  \quad $R$ after 8 & 0.24 \cint{[0.20, 0.27]} & 0.18 \cint{[0.14, 0.21]} & 0.26 \cint{[0.23, 0.30]} & 0.18 \cint{[0.15, 0.21]} & 0.55 \cint{[0.46, 0.65]} \\
  \quad $R$ after 16 & 0.07 \cint{[0.04, 0.09]} & $-$0.05 \cint{[$-$0.10, $-$0.01]} & 0.12 \cint{[0.09, 0.15]} & 0.04 \cint{[0.01, 0.07]} & 0.42 \cint{[0.31, 0.53]} \\
  \quad $R$ after 32 & 0.04 \cint{[0.01, 0.07]} & $-$0.01 \cint{[$-$0.05, 0.03]} & 0.07 \cint{[0.04, 0.10]} & 0.06 \cint{[0.03, 0.08]} & 0.43 \cint{[0.32, 0.54]} \\
  \quad $S$, segment of 16 & 0.54 \cint{[0.47, 0.61]} & 0.53 \cint{[0.47, 0.60]} & 0.54 \cint{[0.47, 0.60]} & 0.54 \cint{[0.47, 0.61]} & 0.58 \cint{[0.54, 0.61]} \\
  \quad $R$ after a neutral passage & 1.56 \cint{[1.46, 1.68]} & -- & 1.66 \cint{[1.54, 1.81]} & -- & 0.90 \cint{[0.71, 1.06]} \\
  \quad Shift of the estimate & 0.00 \cint{[0.00, 0.00]} & 0.00 \cint{[0.00, 0.01]} & 0.00 \cint{[0.00, 0.00]} & 0.00 \cint{[0.00, 0.00]} & 0.02 \cint{[0.01, 0.02]} \\
  \midrule
  \multicolumn{6}{@{}l}{\textit{Qwen3-8B}, layers 9 and 18, assistant-frame continuation} \\
  \quad Installed fraction & 0.89 \cint{[0.63, 1.37]} & 0.99 \cint{[0.71, 1.51]} & 0.90 \cint{[0.65, 1.40]} & 1.54 \cint{[1.12, 2.35]} & 1 \\
  \quad $R$ after 8 & 0.23 \cint{[0.17, 0.30]} & 0.11 \cint{[0.06, 0.15]} & 0.24 \cint{[0.18, 0.31]} & 0.10 \cint{[0.06, 0.14]} & 0.34 \cint{[0.14, 0.58]} \\
  \quad $R$ after 16 & 0.16 \cint{[0.10, 0.21]} & 0.07 \cint{[0.03, 0.10]} & 0.24 \cint{[0.20, 0.28]} & 0.12 \cint{[0.08, 0.16]} & 0.43 \cint{[0.28, 0.60]} \\
  \quad $R$ after 32 & 0.36 \cint{[0.31, 0.43]} & 0.13 \cint{[0.09, 0.17]} & 0.35 \cint{[0.31, 0.40]} & 0.17 \cint{[0.13, 0.21]} & 0.51 \cint{[0.37, 0.69]} \\
  \quad $S$, segment of 16 & 0.69 \cint{[0.62, 0.76]} & 0.70 \cint{[0.63, 0.76]} & 0.69 \cint{[0.63, 0.76]} & 0.70 \cint{[0.64, 0.77]} & 0.67 \cint{[0.63, 0.71]} \\
  \quad $R$ after a neutral passage & 1.00 \cint{[0.89, 1.14]} & -- & 1.07 \cint{[0.96, 1.19]} & -- & 0.87 \cint{[0.74, 1.02]} \\
  \quad Shift of the estimate & 0.01 \cint{[0.01, 0.02]} & 0.01 \cint{[0.00, 0.02]} & 0.02 \cint{[0.01, 0.02]} & 0.05 \cint{[0.03, 0.06]} & 0.21 \cint{[0.19, 0.23]} \\
  \midrule
  \multicolumn{6}{@{}l}{\textit{Qwen3.8-27B}, layers 16 and 32, assistant-frame continuation} \\
  \quad Installed fraction & 0.31 \cint{[0.27, 0.35]} & 1.01 \cint{[0.96, 1.08]} & 0.50 \cint{[0.45, 0.56]} & 1.03 \cint{[0.97, 1.10]} & 1 \\
  \quad $R$ after 8 & 0.82 \cint{[0.74, 0.91]} & 0.60 \cint{[0.58, 0.63]} & 0.48 \cint{[0.43, 0.53]} & 0.56 \cint{[0.54, 0.58]} & 0.47 \cint{[0.44, 0.50]} \\
  \quad $R$ after 16 & 0.61 \cint{[0.54, 0.69]} & 0.40 \cint{[0.38, 0.42]} & 0.34 \cint{[0.31, 0.38]} & 0.40 \cint{[0.39, 0.42]} & 0.29 \cint{[0.26, 0.32]} \\
  \quad $R$ after 32 & 0.32 \cint{[0.29, 0.36]} & 0.19 \cint{[0.18, 0.20]} & 0.20 \cint{[0.18, 0.22]} & 0.25 \cint{[0.24, 0.26]} & 0.13 \cint{[0.10, 0.16]} \\
  \quad $S$, segment of 16 & 1.26 \cint{[1.23, 1.30]} & 1.19 \cint{[1.17, 1.22]} & 1.31 \cint{[1.28, 1.35]} & 1.16 \cint{[1.13, 1.19]} & 1.26 \cint{[1.24, 1.28]} \\
  \quad $R$ after a neutral passage & 0.60 \cint{[0.52, 0.66]} & -- & 0.59 \cint{[0.52, 0.65]} & -- & 0.80 \cint{[0.77, 0.83]} \\
  \quad Shift of the estimate & 0.07 \cint{[0.06, 0.08]} & 0.18 \cint{[0.15, 0.21]} & 0.08 \cint{[0.07, 0.10]} & 0.17 \cint{[0.15, 0.19]} & 0.33 \cint{[0.32, 0.35]} \\
  \bottomrule
\end{tabular}
\begin{tabnotes}
  \item \hspace*{-\labelsep}--: not run; the neutral passage follows the writes of four positions only.
\end{tabnotes}
\end{threeparttable}
\end{table*}

\paragraph{Told direction.}
A told rate leaves a direction about a fifth the size of a record's, and writing it back moves the estimate by at most 0.12 of the natural told difference. The mean difference between states told 75\% and 25\%, taken at the last four positions of the statement line, is 0.05 to 0.06 of the median residual norm there in Llama-3.1-8B and 0.025 to 0.05 in Qwen3-8B, against 0.22 to 0.31 for the record direction. Added to states told 50\% at relative magnitudes from 0.125 to 2 of the median norm, in both signs, no dose moves the estimate by more than 0.12 (Llama-3.1-8B, layer 8) or 0.09 (Qwen3-8B, layer 9) of the natural told difference.

In Qwen3-8B the layer-9 write moves the assistant-frame continuation by 0.148 [0.038, 0.273], 0.79 of the natural difference on that readout, and the movement is retained at 0.30 after sixteen observations, against 0.43 for natural records. In Llama-3.1-8B the document-frame continuation of told states moves by 0.04 to 0.05 and the movement reverses after the segment, while on shown records the largest dose moves the estimate by 0.36 of the told difference and the assistant-frame continuation down by 0.20. In Qwen3.8-27B a direction taken from told states moves the continuation by less than a fifth of the natural told difference.

\paragraph{Weight direction.}
In Llama-3.1-8B the mean difference between records of 64 and of 8 observations (cosine 0.98 to 0.99 across content levels) is added to records of 8, with either sign, at one half to two times its natural magnitude at layers 8 and 16. Under every dose the weight of the short record stays between 2.9 and 4.7 observations, against 4.6 under sham and 17.2 [15.0, 19.5] for the long record, and no interval reaches above 6.0. The larger doses lower it slightly, as a random direction does. In Qwen3.8-27B the same direction leaves the short record at 1.9 to 3.1 observations, against 2.7 under sham and 21.3 [19.8, 22.9] for the long record, and no interval reaches above 3.6. Qwen3-8B is not tested, since the weight of its record does not grow with the record's length (Section~\ref{sec:weight}).

\paragraph{Neutral passage.}
A neutral passage with the token count of a new segment of sixteen observations is appended to the written states in place of the segment. In Llama-3.1-8B (26 tokens, layers 8 and 16, four positions) the written content retains 0.066 and 0.120 after the segment, the values of Section~\ref{sec:writes}, and 1.56 [1.46, 1.68] and 1.66 [1.54, 1.81] after the passage, while the natural content of the same records retains 0.90 [0.71, 1.06]. A retention above one has a simple reading: the passage carries no coin observations, so nothing in it competes with the written content, and it dilutes the record's tokens in what the readout attends to, which leaves the bias injected at the record's last positions a larger share. In Qwen3-8B (layers 9 and 18) the written content retains 0.16 and 0.24 after the segment and 1.00 [0.89, 1.14] and 1.07 [0.96, 1.19] after the passage, while the natural content retains 0.87 [0.74, 1.02]. In Qwen3.8-27B the passage lowers the written content to 0.60 where the natural content keeps 0.80.

\section{Weights in full}
\label{app:rates}

\begin{figure*}[t]
  \centering
  \includegraphics[width=\textwidth]{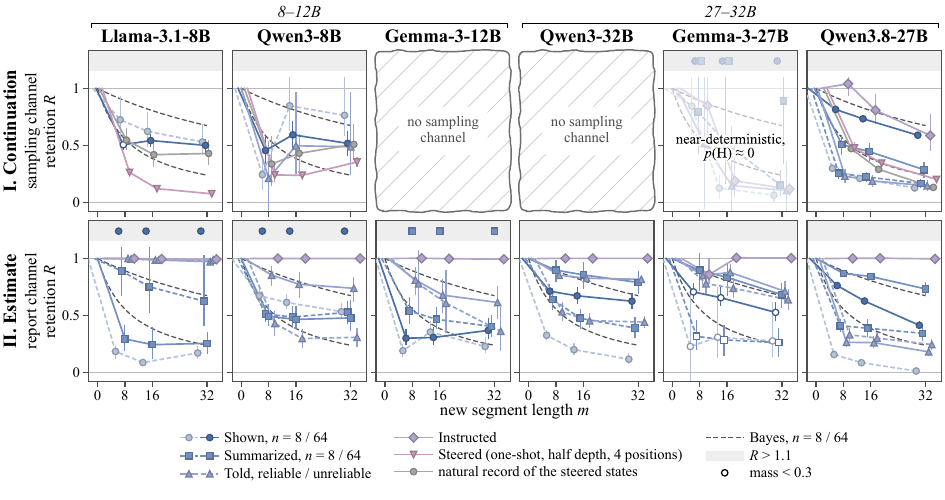}
  \caption{\textbf{The content of every route under the same new evidence.}
    Retention $R = a/a_0$ of the old content after a new segment of $m$ observations. Columns: the six models. Top: the continuation in the model's sampling channel. Bottom: the estimate in its report channel. Dashed: the Bayesian observer for records of 8 and 64. Pink: a one-shot write at half the depth, with the natural record of the written states in grey. Top band: $R$ above 1.1, in cells whose content is hardly read before the segment ($a_0 < 0.35$). Gemma-3-12B and Qwen3-32B have no sampling channel, and Gemma-3-27B's is close to deterministic. Qwen3-8B's sampling channel pools the two sources of a told rate. Family-bootstrap 95\% intervals.}
  \label{fig:retention}
\end{figure*}

\begin{figure}[t]
  \centering
  \includegraphics[width=\columnwidth]{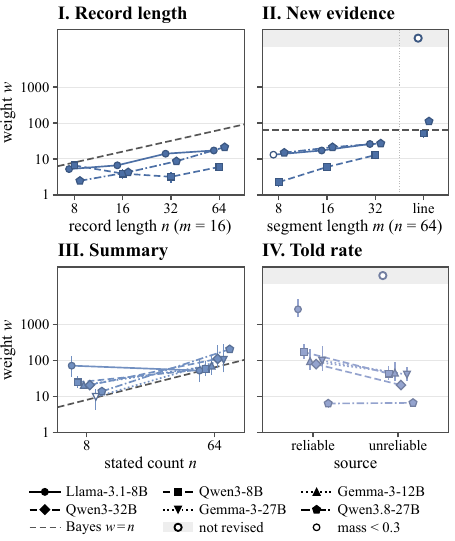}
  \caption{\textbf{Weights against their Bayesian references.}
    I. A shown record in the sampling channel against its length (new segment, $m = 16$), with the Bayesian line $w = n$. II. The same record at $n = 64$ against the length of the new segment, and under the extended line. I and II show Llama-3.1-8B, Qwen3-8B and Qwen3.8-27B. III. A summary in the report channel against the count it states. IV. A told rate in the report channel by source reliability. III and IV show the six models. One log scale throughout; top band: not revised. Family-bootstrap 95\% intervals.}
  \label{fig:rates}
\end{figure}

\begin{figure}[t]
  \centering
  \includegraphics[width=\columnwidth]{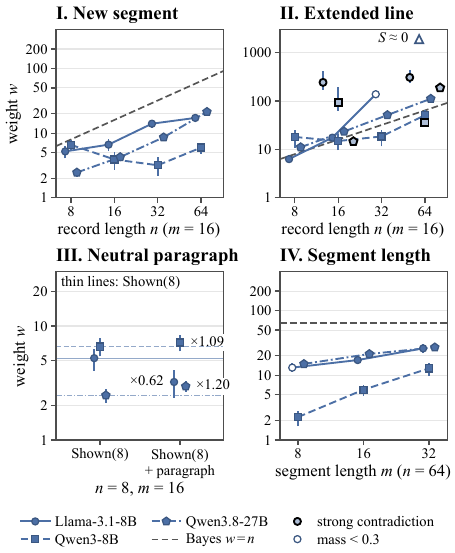}
  \caption{\textbf{The sampling-channel weight of a shown record under every layout.}
    Llama-3.1-8B, Qwen3-8B and Qwen3.8-27B. I. New segment ($m = 16$): $w$ against the record length $n$, with the Bayesian line $w = n$. Llama-3.1-8B rises and flattens beyond $n = 32$, Qwen3-8B stays flat, and Qwen3.8-27B stays near a third of the record's length. II. Extended line: markers at the top edge are not revised (primacy); black-edged markers, the strong contradiction. III. A neutral paragraph before the record: ratio of $w$ to the undiluted record. IV. $w$ against the segment length $m$ at $n = 64$. Family-bootstrap 95\% intervals.}
  \label{fig:sampling-full}
\end{figure}

\begin{figure*}[t]
  \centering
  \includegraphics[width=\textwidth]{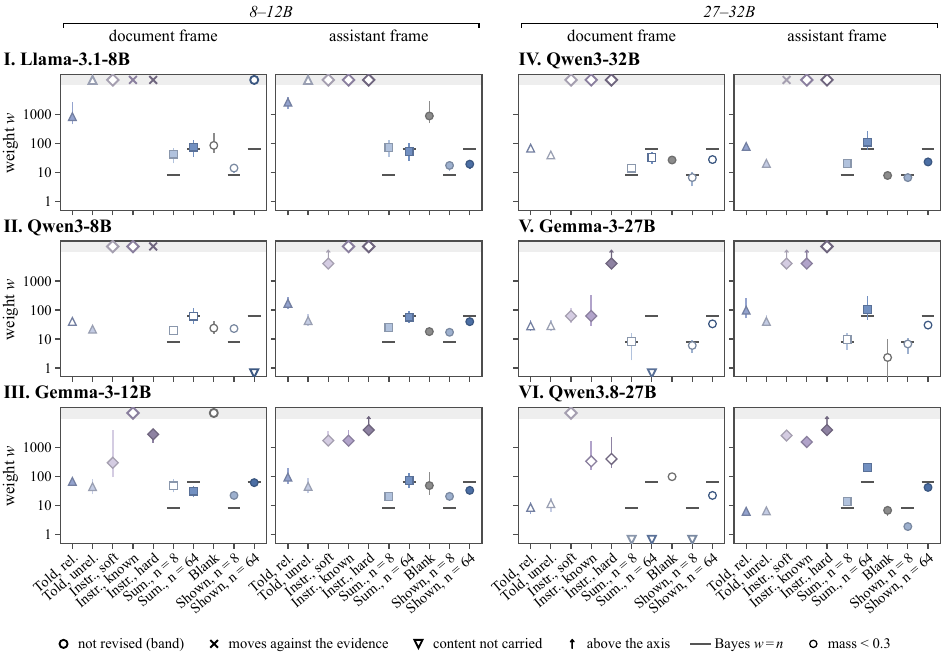}
  \caption{\textbf{The weight of every route in the estimate.}
    Weight $w$ of each route in the estimate of each of the six models, in the document frame and the assistant frame, after a new segment of sixteen observations. Hollow markers in the top band: not revised; crosses: moves against the evidence ($S < 0$); hollow down-triangle: content not carried ($a \approx 0$); arrows: values above the axis; other hollow markers: answer mass below 0.3; grey ticks: the Bayesian $w = n$ for summaries and records. Family-bootstrap 95\% intervals.}
  \label{fig:report-full}
\end{figure*}

Figure~\ref{fig:rates} sets the weights of Section~\ref{sec:weight} against their references. Figure~\ref{fig:sampling-full} gives the sampling-channel weight of a shown record under every layout, and Figure~\ref{fig:report-full} the weight of every route in the estimate, in both frames, with the symbols for cells that are not revised, that move against the evidence or whose content is not carried. Figure~\ref{fig:retention} follows the content of every route through the same new evidence in the channels of every model.

\paragraph{Priors.}
With no content at all, the report channel is held by a prior worth 887 observations in Llama-3.1-8B, about 20 in Qwen3-8B, 7.9 in Qwen3-32B and 6.8 in Qwen3.8-27B, and the sampling channel by one worth 14 in Llama-3.1-8B and about 20 in Qwen3-8B.

\paragraph{A told rate by source reliability.}
The Bayesian reference of Appendix~\ref{app:bayes} discounts a less reliable source from the start and drops its rate once the evidence contradicts it. In the report channel Llama-3.1-8B does neither. Qwen3-8B drops the statement without discounting it, keeping 0.77 of a reliable rate and 0.30 of an unreliable one after sixteen observations while moving toward the evidence at the same rate for both. Gemma-3-12B discounts the statement but does not drop it. Qwen3-32B does both: it reads an unreliable rate at 0.82 against 0.99 and keeps 0.45 of it against 0.83 after sixteen observations. Gemma-3-27B reads both rates at 1.00 and keeps 0.74 of the unreliable one against 0.88, and Qwen3.8-27B reads 0.74 against 0.88 and keeps about 0.3 of either. Under the next-toss question the smaller models follow other patterns (Appendix~\ref{app:queries}).

\paragraph{New outcomes on the record's own line.}
At $n = 64$ the continuation responds to sixteen new outcomes in a new segment with a slope of 0.36 in Llama-3.1-8B, 0.71 in Qwen3-8B and 0.76 in Qwen3.8-27B, where a Bayesian observer has 0.20. Continued on the record's own line, the same outcomes give slopes of 0.00, 0.08 and 0.18 (Figure~\ref{fig:sampling-full}, II).

\paragraph{The sampling channel of the larger models.}
In Qwen3.8-27B's sampling channel a record is worth 2.5, 4.3, 8.6 and 21 observations at $n = 8$ to 64 under a new segment of sixteen, a third of its length throughout, and the weight rises with the length of the segment, from 15 to 27 observations at $n = 64$ (Table~\ref{tab:sampling}). Continued on the record's own line, the same sixteen outcomes are under-weighted without being ignored: the slope on them is 0.18, and the record is then worth 111 observations. A told rate enters this channel with content and little weight (retention 0.19, 2.6 observations). Gemma-3-27B's channel returns 114 observations at $n = 64$ from a response to the segment close to zero, which we do not read as a weight.

\begin{table*}[t]
  \centering
  \footnotesize
  \caption{\textbf{A shown record in the sampling channel of Gemma-3-27B and Qwen3.8-27B.} Weight $w$ and retention $R$ by the length $n$ of the record and the length $m$ of the new segment, and under the extended line of sixteen outcomes. Gemma-3-27B's channel is close to deterministic. Family-bootstrap 95\% intervals.}
  \label{tab:sampling}
  \setlength{\tabcolsep}{3.5pt}
  \providecommand{\cint}[1]{{\fontsize{7}{8}\selectfont #1}}
\begin{threeparttable}
\begin{tabular}{@{}lrcrcrcrcr@{}}
  \toprule
  & & \multicolumn{2}{c}{$m = 8$} & \multicolumn{2}{c}{$m = 16$} & \multicolumn{2}{c}{$m = 32$} & \multicolumn{2}{c}{Extended line} \\
  \cmidrule(lr){3-4} \cmidrule(lr){5-6} \cmidrule(lr){7-8} \cmidrule(l){9-10}
  Model & $n$ & $w$ & $R$ & $w$ & $R$ & $w$ & $R$ & $w$ & $R$ \\
  \midrule
  Gemma-3-27B & 8 & $S<0$ & 0.84 & $S<0$ & 0.12 & 17 \cint{[2.0, 31]} & 0.06 & 1.9 \cint{[$-$2.6, 6.8]} & 0.02 \\
   & 16 & $\infty$ \cint{($\geq$\,1227)} & -- & 34 \cint{[27, 44]} & -- & 13 \cint{[5.6, 20]} & -- & 2.9 \cint{[$-$3.5, 8.5]} & -- \\
   & 32 & 90 \cint{[44, 433]} & 8.39 & 49 \cint{[40, 65]} & 7.62 & 26 \cint{[19, 31]} & 4.20 & 4.6 \cint{[0.0, 9.0]} & 0.82 \\
   & 64 & $\infty$ \cint{($\geq$\,220)} & 2.43 & 114 \cint{[85, 159]} & 2.15 & 69 \cint{[58, 82]} & 1.64 & 20 \cint{[16, 25]} & 0.74 \\
  \midrule
  Qwen3.8-27B & 8 & 1.9 \cint{[1.7, 2.2]} & 0.30 & 2.5 \cint{[2.1, 2.8]} & 0.21 & 2.8 \cint{[2.2, 3.4]} & 0.13 & 11 \cint{[10, 12]} & 0.49 \\
   & 16 & 4.2 \cint{[3.9, 4.6]} & 0.48 & 4.3 \cint{[3.8, 4.7]} & 0.31 & 3.5 \cint{[2.8, 4.2]} & 0.14 & 23 \cint{[22, 25]} & 0.66 \\
   & 32 & 7.1 \cint{[6.7, 7.6]} & 0.62 & 8.6 \cint{[7.9, 9.3]} & 0.47 & 8.9 \cint{[8.0, 9.9]} & 0.29 & 51 \cint{[48, 55]} & 0.86 \\
   & 64 & 15 \cint{[14, 16]} & 0.82 & 21 \cint{[20, 23]} & 0.73 & 27 \cint{[25, 29]} & 0.59 & 111 \cint{[104, 120]} & 0.93 \\
  \bottomrule
\end{tabular}
\begin{tabnotes}
  \item \hspace*{-\labelsep}$\infty$: not revised, the interval of the sensitivity includes zero or the weight exceeds a thousand observations, with the lower bound of $w$ in parentheses.
  \item \hspace*{-\labelsep}$S < 0$: the response to the segment is negative and no weight is defined.
  \item \hspace*{-\labelsep}--: no content before the segment.
\end{tabnotes}
\end{threeparttable}
\end{table*}

\section{Models and noise floor}
\label{app:models}

Table~\ref{tab:models} lists the models and how each is read: the six open-weight models run locally, Llama-3.1-8B, Qwen3-8B and Gemma-3-12B, the smaller models, and Qwen3-32B, Gemma-3-27B and Qwen3.8-27B, the larger models, the three largest models (Appendix~\ref{app:further}) and three proprietary models (Appendix~\ref{app:hosted}). Qwen3-8B, Qwen3-32B and Qwen3.8-27B are read in non-thinking mode, and all six local models are loaded in bf16 at pinned revisions. The smaller models were run on a single NVIDIA RTX 5090 and the larger models on a single NVIDIA H100.

\paragraph{Qwen3.8-27B.}
Qwen3.8-27B \citep{qwenteam2026qwen38} is a hybrid of 64 layers. In every four layers, three are Gated DeltaNet linear-attention layers, which carry a recurrent state forward, and one is a full-attention layer. The model has a vision encoder and is read with text input only.

\begin{figure}[t]
  \centering
  \includegraphics[width=\columnwidth]{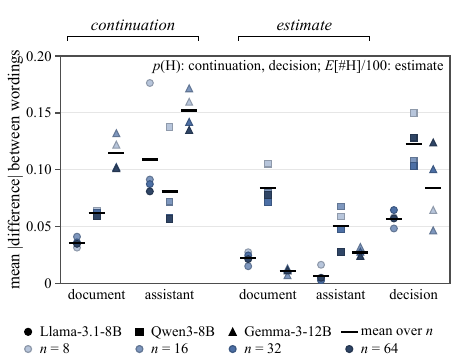}
  \caption{\textbf{Wording noise floor of the smaller models.} Mean absolute difference between the four wordings of the same shown record, per readout, model and record length, in probability units for the continuations and the decision check and in expected count over 100 for the estimates.}
  \label{fig:noise-floor}
\end{figure}

\paragraph{Wordings and noise floor.}
Shown records use four wordings of the record line, summaries eight wordings of the counts, told rates four wordings with the source's past accuracy stated as 95\% or 55\%, and instructions three strengths. The texts are in Appendix~\ref{app:prompts}. Table~\ref{tab:noise} gives the wording noise floor of every readout of the smaller models: the mean absolute difference between the four wordings of the same shown record, in logit units for the continuations and the decision check and in expected count for the estimates, over 240 families. Figure~\ref{fig:noise-floor} shows the same differences by record length, in probability units and in expected count over 100. The floors of the larger models are computed in the same way. The equivalence tolerances are derived from these values: the tolerances for a continuation are the probability differences divided by the 0.5 range of the content proportion, which turns them into slopes (0.07 for Llama-3.1-8B and 0.16 for Qwen3-8B), the tolerances for an estimate are the count differences, and the steering comparison uses the within-family standard deviation of the same wordings in logit units (0.15), all in Appendix~\ref{app:estimators}.

\begin{table*}[t]
  \centering
  \footnotesize
  \caption{\textbf{The models and how each is read.} One protocol, read as deeply as each model can be accessed. For the models run locally, the repository name links to the revision that was read.}
  \label{tab:models}
  \setlength{\tabcolsep}{3.4pt}
  \begin{threeparttable}
  \begin{tabular}{@{}lllll@{}}
    \toprule
    Model & Parameters & Attention & Source & Results \\
    \midrule
    \multicolumn{5}{@{}l}{\textit{Open weights, run locally: token probabilities of both readouts, activations written}} \\
    \quad Llama-3.1-8B & 8B & attention-only
      & \href{https://huggingface.co/meta-llama/Llama-3.1-8B-Instruct/tree/0e9e39f249a16976918f6564b8830bc894c89659}{\texttt{meta-llama/Llama-3.1-8B-Instruct}} & Sections~\ref{sec:content} to \ref{sec:writes} \\
    \quad Qwen3-8B & 8B & attention-only
      & \href{https://huggingface.co/Qwen/Qwen3-8B/tree/b968826d9c46dd6066d109eabc6255188de91218}{\texttt{Qwen/Qwen3-8B}} & Sections~\ref{sec:content} to \ref{sec:writes} \\
    \quad Gemma-3-12B & 12B & attention-only
      & \href{https://huggingface.co/google/gemma-3-12b-it/tree/96b6f1eccf38110c56df3a15bffe176da04bfd80}{\texttt{google/gemma-3-12b-it}} & Sections~\ref{sec:content}, \ref{sec:weight} \\
    \quad Qwen3-32B & 32B & attention-only
      & \href{https://huggingface.co/Qwen/Qwen3-32B/tree/9216db5781bf21249d130ec9da846c4624c16137}{\texttt{Qwen/Qwen3-32B}} & Sections~\ref{sec:content}, \ref{sec:weight} \\
    \quad Gemma-3-27B & 27B & attention-only
      & \href{https://huggingface.co/google/gemma-3-27b-it/tree/005ad3404e59d6023443cb575daa05336842228a}{\texttt{google/gemma-3-27b-it}} & Sections~\ref{sec:content}, \ref{sec:weight} \\
    \quad Qwen3.8-27B & 27B & hybrid\tnote{a}
      & \href{https://huggingface.co/Qwen/Qwen3.8-27B/tree/37561816e66b5b9153b1b64c10968c18c1e6f4a0}{\texttt{Qwen/Qwen3.8-27B}} & Sections~\ref{sec:content} to \ref{sec:writes} \\
    \addlinespace
    \multicolumn{5}{@{}l}{\textit{Open weights, served: sampled answers}} \\
    \quad Qwen3.5-122B & 122B, 10B active\tnote{b} & hybrid\tnote{a} & \citet{qwenteam2026qwen35} & Section~\ref{sec:weight}, Appendices~\ref{app:further}, \ref{app:reasoning} \\
    \quad Qwen3-235B & 235B, 22B active\tnote{b} & attention-only & \citet{yang2025qwen3} & Section~\ref{sec:weight}, Appendices~\ref{app:further}, \ref{app:hosted} \\
    \quad Qwen3.5-397B & 397B, 17B active\tnote{b} & hybrid\tnote{a} & \citet{qwenteam2026qwen35} & Section~\ref{sec:weight}, Appendix~\ref{app:further} \\
    \addlinespace
    \multicolumn{5}{@{}l}{\textit{Proprietary: sampled replies}} \\
    \quad GPT-6 Sol & not public & not public & \citet{openai2026gpt6sol} & Appendix~\ref{app:hosted} \\
    \quad Gemini 4 Argon & not public & not public & \citet{google2026gemini4argon} & Appendix~\ref{app:hosted} \\
    \quad Claude Opus 5.5 & not public & not public & \citet{anthropic2026opus55} & Appendix~\ref{app:hosted} \\
    \bottomrule
  \end{tabular}
  \begin{tabnotes}
    \item[a] Three of every four layers are linear-attention layers that carry a recurrent state, and one is a full-attention layer.
    \item[b] Mixture of experts.
  \end{tabnotes}
  \end{threeparttable}
\end{table*}

\begin{table}[t]
  \centering
  \footnotesize
  \caption{\textbf{Wording noise floor of the smaller models.} Mean absolute difference between the wordings of the same shown record, per readout and frame: the continuation and the one-token decision check in logit units, the estimate in expected count.}
  \label{tab:noise}
  \setlength{\tabcolsep}{4pt}
  \begin{tabular}{@{}lccccc@{}}
    \toprule
    & \multicolumn{2}{c}{Continuation} & \multicolumn{2}{c}{Estimate} & Decision \\
    \cmidrule(lr){2-3} \cmidrule(lr){4-5}
    Model & doc. & asst. & doc. & asst. & check \\
    \midrule
    Llama-3.1-8B & 0.16 & 0.64 & 2.2 & 0.7 & 0.25 \\
    Qwen3-8B     & 0.29 & 0.36 & 8.4 & 5.1 & 0.84 \\
    Gemma-3-12B  & 0.59 & 1.15 & 1.1 & 2.7 & 1.46 \\
    \bottomrule
  \end{tabular}
\end{table}

\section{Paraphrases of the expected-count question}
\label{app:queries}

The report channel of the main text is read by one query, Q1 \prompt{Expected number of H among the next 100 tosses:}. Three paraphrases read the same states: Q2 \prompt{Probability that the next outcome is H, as a whole-number percentage:}, which asks for the quantity the continuation gives; Q3 \prompt{Best estimate of the number of H in the next 100 tosses:}; and Q4 \prompt{If this coin is tossed 100 more times, the expected number of H is:}. Each is read as an integer in both frames on the shown, told, instructed and summarized states of the channel map, and Q2 is also read after the same new segments of sixteen observations as in Section~\ref{sec:weight}, with only the query line changed. Table~\ref{tab:queries} gives the slopes and the weights of the six models.

\begin{table*}[t]
  \centering
  \footnotesize
  \caption{\textbf{The estimate under four queries, assistant frame.} Slope of the current estimate on the content of the route (Told at the central three levels; Instructed at the central three under Q1 and at all five under Q2 to Q4, which round alike), and the weight $w$ of the content under a new segment of sixteen observations with Q1 and with Q2, Told at the central three levels under both. Family-bootstrap 95\% intervals.}
  \label{tab:queries}
  \setlength{\tabcolsep}{4pt}
  \providecommand{\cint}[1]{{\fontsize{7}{8}\selectfont #1}}
\begin{threeparttable}
\begin{tabular}{@{}llrrrrcc@{}}
  \toprule
  Model & Route & Q1 & Q2 & Q3 & Q4 & $w$ (Q1) & $w$ (Q2) \\
  \midrule
  Llama-3.1-8B & Told & 0.99 & 0.99 & 0.99 & 0.97 & $\infty$ \cint{($\geq$\,13\,102)} & 54 \cint{[40, 73]} \\
   & Instructed & 1.00 & 1.00 & 1.00 & n.r. & $\infty$ \cint{($\geq$\,11\,987)} & 49 \cint{[28, 112]} \\
   & Summarized, $n = 8$ & 0.53 & 0.94 & 0.75 & 0.46 & 71 \cint{[35, 132]} & 24 \cint{[19, 29]} \\
   & Summarized, $n = 64$ & 0.51 & 0.93 & 0.72 & 0.40 & 52 \cint{[25, 103]} & 19 \cint{[14, 25]} \\
   & Shown, $n = 8$ & 0.12 & 0.42 & 0.62 & 0.20 & 17 \cint{[12, 23]} & 7.3 \cint{[5.3, 9.3]} \\
   & Shown, $n = 64$ & 0.02 & 0.43 & 0.15 & 0.06 & 19 \cint{[13, 27]} & 17 \cint{[12, 23]} \\
  \midrule
  Qwen3-8B & Told & 0.99 & 0.85 & 0.98 & 0.99 & 288 \cint{[160, 714]} & 10 \cint{[7.4, 13]} \\
   & Instructed & 1.00 & 1.00 & 1.00 & 0.99 & $\infty$ & 48 \cint{[20, 129]} \\
   & Summarized, $n = 8$ & 0.96 & 1.00 & 0.99 & 0.71 & 25 \cint{[17, 36]} & 6.4 \cint{[4.4, 8.8]} \\
   & Summarized, $n = 64$ & 1.01 & 1.02 & 1.00 & n.r. & 57 \cint{[36, 94]} & 11 \cint{[8.9, 14]} \\
   & Shown, $n = 8$ & 0.42 & 0.50 & 0.50 & 0.49 & 17 \cint{[14, 20]} & 6.1 \cint{[4.9, 7.3]} \\
   & Shown, $n = 64$ & 0.20 & 0.66 & 0.36 & 0.07 & 40 \cint{[35, 45]} & 32 \cint{[28, 36]} \\
  \midrule
  Gemma-3-12B & Told & 0.91 & 0.37 & 0.96 & n.r. & 205 \cint{[116, 535]} & 23 \cint{[16, 32]} \\
   & Instructed & 1.00 & 1.00 & 1.00 & n.r. & $\infty$ & $\infty$ \\
   & Summarized, $n = 8$ & 0.80 & 1.01 & 0.75 & n.r. & 21 \cint{[17, 25]} & 7.6 \cint{[5.7, 9.6]} \\
   & Summarized, $n = 64$ & 0.31 & 1.01 & 0.71 & n.r. & 72 \cint{[40, 136]} & 13 \cint{[8.2, 20]} \\
   & Shown, $n = 8$ & 0.18 & 0.50 & 0.71 & n.r. & 21 \cint{[19, 22]} & 2.3 \cint{[1.4, 3.1]} \\
   & Shown, $n = 64$ & 0.32 & 0.44 & 0.75 & n.r. & 33 \cint{[30, 37]} & 20 \cint{[18, 22]} \\
  \midrule
  Qwen3-32B & Told & 0.90 & 0.84 & 0.72 & 0.93 & 77 \cint{[58, 102]} & 5.5 \cint{[4.3, 6.7]} \\
   & Instructed & 1.00 & 1.00 & 1.00 & 0.99 & $\infty$ \cint{($\geq$\,230\,695)} & $\infty$ \\
   & Summarized, $n = 8$ & 0.97 & 0.99 & 0.98 & 0.97 & 21 \cint{[15, 27]} & 5.2 \cint{[3.7, 6.8]} \\
   & Summarized, $n = 64$ & 0.98 & 0.99 & 0.99 & 0.94 & 111 \cint{[60, 264]} & 32 \cint{[19, 51]} \\
   & Shown, $n = 8$ & 0.48 & 0.17 & 0.91 & 0.34 & 6.7 \cint{[5.1, 8.4]} & 0.4 \cint{[$-$0.8, 1.6]} \\
   & Shown, $n = 64$ & 0.59 & 0.64 & 1.02 & 0.24 & 23 \cint{[21, 26]} & 55 \cint{[46, 66]} \\
  \midrule
  Gemma-3-27B & Told & 1.00 & 0.77 & 1.00 & n.r. & 95 \cint{[57, 187]} & 6.0 \cint{[4.6, 7.6]} \\
   & Instructed & 1.00 & 1.00 & 1.00 & n.r. & $\infty$ & 433 \cint{[161, 7938]} \\
   & Summarized, $n = 8$ & 1.23 & 1.06 & 1.00 & n.r. & n.r. & 4.0 \cint{[2.1, 6.0]} \\
   & Summarized, $n = 64$ & 1.04 & 1.00 & 1.03 & n.r. & 104 \cint{[46, 294]} & 16 \cint{[8.9, 26]} \\
   & Shown, $n = 8$ & n.r. & 0.56 & n.r. & n.r. & n.r. & 0.0 \cint{[$-$0.7, 0.7]} \\
   & Shown, $n = 64$ & n.r. & 0.58 & 0.72 & 0.80 & n.r. & 22 \cint{[18, 27]} \\
  \midrule
  Qwen3.8-27B & Told & 0.71 & 0.80 & 0.91 & 0.87 & 11 \cint{[8.9, 13]} & 4.2 \cint{[3.4, 4.9]} \\
   & Instructed & 0.98 & 0.99 & 1.00 & 0.96 & $\infty$ & 163 \cint{[83, 674]} \\
   & Summarized, $n = 8$ & 0.91 & 1.01 & 0.98 & 0.90 & 14 \cint{[10, 18]} & 7.7 \cint{[6.4, 9.0]} \\
   & Summarized, $n = 64$ & 0.95 & 1.03 & 0.98 & 0.89 & 205 \cint{[162, 259]} & 108 \cint{[88, 133]} \\
   & Shown, $n = 8$ & 0.63 & 0.59 & 0.70 & 0.46 & 1.9 \cint{[1.3, 2.6]} & 1.0 \cint{[0.4, 1.7]} \\
   & Shown, $n = 64$ & 0.97 & 0.84 & 0.91 & 0.88 & 42 \cint{[38, 47]} & 28 \cint{[26, 31]} \\
  \bottomrule
\end{tabular}
\begin{tabnotes}
  \item \hspace*{-\labelsep}$\infty$: not revised, the interval of the sensitivity includes zero or the weight exceeds a thousand observations, with the lower bound of $w$ in parentheses where the interval of the sensitivity includes zero.
  \item \hspace*{-\labelsep}n.r.: answer mass below 0.3, not read.
\end{tabnotes}
\end{threeparttable}
\end{table*}

\paragraph{Slopes.}
A told rate and an instruction are reported at 0.85 to 1.00 under every readable query in Llama-3.1-8B and Qwen3-8B, and Gemma-3-12B answers the next-toss question with a number of its own (0.37). In the smaller models the record and the summary are mostly reported more under Q2 and Q3 than under Q1 and Q4, by up to 0.7 of the slope. In the document frame Q2 leaves an integer mass below 0.3 on told rates in Llama-3.1-8B and Qwen3-8B and on records and summaries in Gemma-3-12B, so Q2 is read in the assistant frame throughout. Q3 is readable in the document frame for told rates in the three smaller models (0.92, 0.97 and 0.84).

\paragraph{Weights under the expected-count question.}
In the larger models a summary of 8 is worth 14 and 21 observations in Qwen3.8-27B and Qwen3-32B and a summary of 64 is worth 104 to 205 in the three models, more than the count it states, while a record of 64 is worth 23 and 42 in Qwen3-32B and Qwen3.8-27B. An instruction is not revised in any of the six models. Set beside Qwen3-32B, the hybrid Qwen3.8-27B gives a record a weight that grows 22-fold from $n = 8$ to 64, where that of Qwen3-32B grows 3.4-fold, and a summary a weight that grows 15-fold against 5-fold, and a told rate is worth a seventh as much (11 against 77 observations).

\paragraph{Weights under the next-toss question.}
In the smaller models every weight is smaller under Q2 than under Q1. A told rate's weight falls by a factor of about 9 in Gemma-3-12B and 29 in Qwen3-8B, and from no revision to 54 observations in Llama-3.1-8B. An instruction's falls from no revision to about 50 observations in Llama-3.1-8B and Qwen3-8B, while Gemma-3-12B holds it ($S = 0.010$, equivalent to zero). A summary's and a record's fall by factors of 1 to 10. In the larger models a told rate's weight falls from 77 to 5.5 observations in Qwen3-32B, from 95 to 6.0 in Gemma-3-27B and from 11 to 4.2 in Qwen3.8-27B, and an instruction is worth 163 observations in Qwen3.8-27B and 433 in Gemma-3-27B. The order of record and summary depends on the model: 55 against 32 observations in Qwen3-32B, with overlapping intervals, 22 against 16 in Gemma-3-27B and 28 against 108 in Qwen3.8-27B. Part of Llama-3.1-8B's answers under Q2 after a summary and a new segment are counts of the segment, which lowers its estimate there (Appendix~\ref{app:matched}).

\paragraph{Source reliability under the next-toss question.}
With 24 paired states per model, Llama-3.1-8B retains a reliable and an unreliable rate alike ($R$ 0.39 and 0.42), Qwen3-8B reads the unreliable rate at 0.60 against 1.11 before any observation and retains 0.20 and 0.27 of the two afterwards, and Gemma-3-12B's told content hardly enters this readout (0.41 and 0.34).

\providecommand{\cint}[1]{{\fontsize{7}{8}\selectfont #1}}

\section{Matched current estimates}
\label{app:matched}

\paragraph{Design.}
Each route is prepared at seven content levels. Shown records and summaries hold 16 or 64 observations with 4, 5, 7, 8, 9, 11 and 12, or 16, 21, 27, 32, 37, 43 and 48 heads, eight families per level (the four record wordings twice, and the eight summary templates once, each stating the counts of a record drawn alongside). Told rates and instructions take the values 0.25, 0.33, 0.42, 0.50, 0.58, 0.67 and 0.75, with two sources and four templates for a told rate and three strengths for an instruction, and one state has no content. Every state is read before and after the same new segments of sixteen observations with 25\% or 75\% heads, in four orders.

Two states of different routes are matched when their current estimates in a readout differ by less than 0.02 (two points). For each matched pair we compare the two estimates after the segment, and the two \emph{spreads}, the estimate after the 75\% segment minus the estimate after the 25\% segment. Intervals resample the families of both routes within content levels and recompute the matches. The \Steered{} states are the one-shot writes of Llama-3.1-8B (Section~\ref{sec:writes}), the mean-difference direction added at layer 8 or 16 with either sign to the last four positions of a record of sixteen observations with 50\% heads, whose current estimates cover the range of the natural records.

\begin{figure*}[p]
  \centering
  \includegraphics[width=\textwidth]{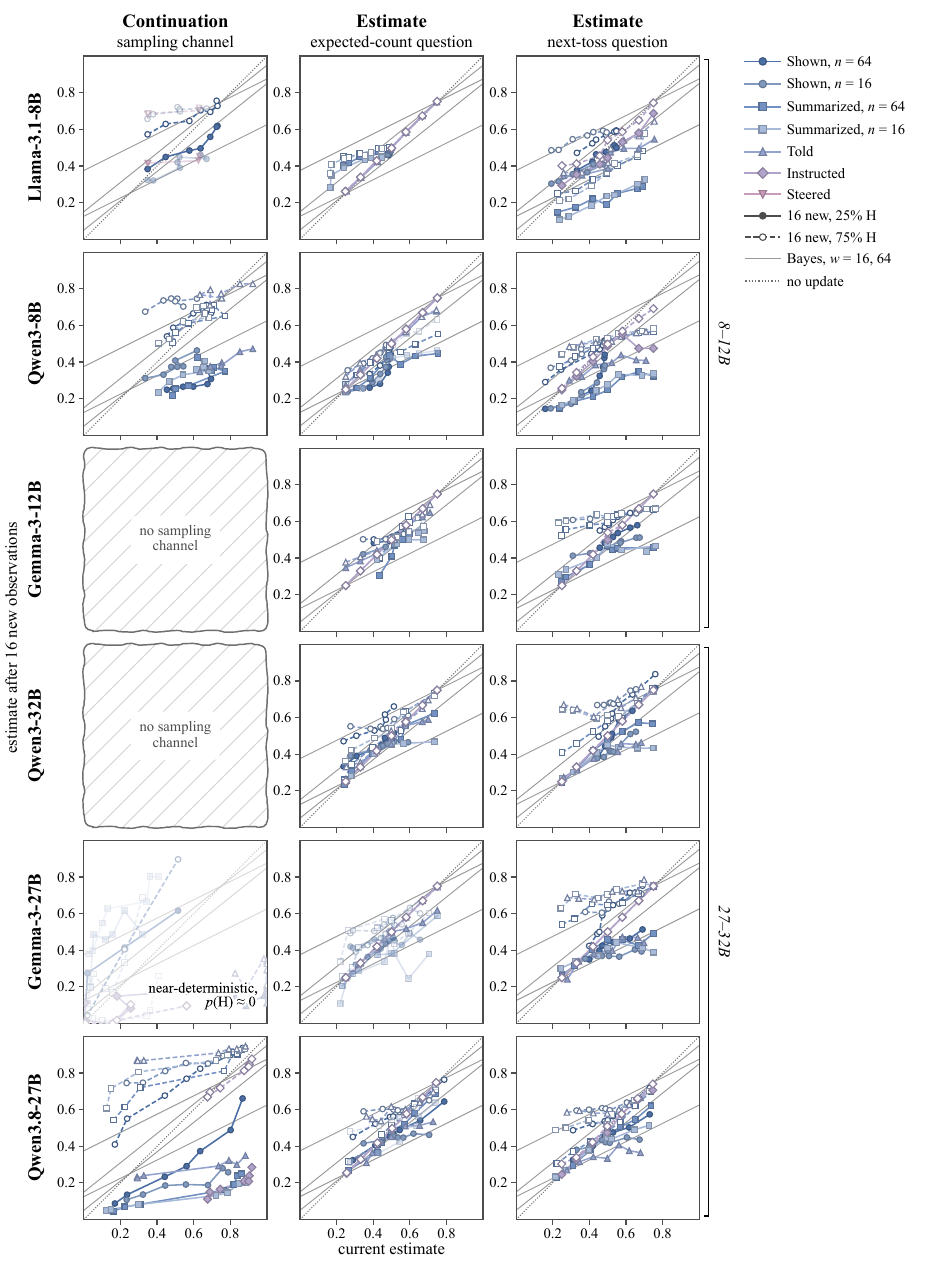}
  \caption{\textbf{The estimate after sixteen new observations against the current estimate, route by route.} Rows: the six models. Columns: the sampling channel, the expected-count question and the next-toss question. Each point is the mean over the families of one content level, solid after sixteen observations with 25\% heads and hollow after sixteen with 75\% heads. Grey: a Bayesian observer whose current estimate is worth 16 or 64 observations. Dotted: no update. A route is drawn where its content reaches the readout, faded where the answer mass is below 0.3. The sampling-channel panel is empty for Gemma-3-12B and Qwen3-32B, which have no sampling channel, and faded for Gemma-3-27B, whose channel is close to deterministic.}
  \label{fig:matched}
\end{figure*}

\begin{table*}[t]
  \centering
  \footnotesize
  \caption{\textbf{Equal current estimates, different futures.} Pairs of states of two routes whose current estimates differ by less than 0.02: their number, the range of the matched estimates, the spread of each route (the estimate after sixteen observations with 75\% heads minus that after 25\%) with its difference $x - y$, and the difference $x - y$ of the estimates after each segment. The larger models are shown wherever both routes can be read. Family-bootstrap 95\% intervals, pairs recomputed on each resample.}
  \label{tab:matched}
  \setlength{\tabcolsep}{2.1pt}
  \providecommand{\cint}[1]{{\fontsize{7}{8}\selectfont #1}}
\begin{threeparttable}
\begin{tabular}{@{}lllrcccc@{}}
  \toprule
  Model & Readout & Routes $x$ / $y$ & Pairs & Range & Spread $x$ / $y$ & $\Delta$ spread & $\Delta$ after 25 / 75\% \\
  \midrule
  Llama-3.1-8B & cont. & \Shown{} 64 / \Shown{} 16$^\dagger$ & 265 & 0.25--0.79 & 0.16 / 0.30 & $-$0.14 \cint{[$-$0.19, $-$0.09]} & 0.09 / $-$0.05 \\
   & cont. & \Shown{} 64 / \Steered{}$^\dagger$ & 629 & 0.25--0.75 & 0.16 / 0.26 & $-$0.10 \cint{[$-$0.15, $-$0.05]} & 0.06 / $-$0.04 \\
   & cont. & \Shown{} 16$^\dagger$ / \Steered{}$^\dagger$ & 678 & 0.23--0.74 & 0.30 / 0.27 & 0.03 \cint{[$-$0.01, 0.07]} & $-$0.03 / $-$0.01 \\
   & count & \Shown{} 64 / \Summarized{} 64 & 655 & 0.45--0.52 & 0.02 / 0.00 & 0.01 \cint{[0.01, 0.02]} & $-$0.01 / 0.00 \\
   & count & \Shown{} 64 / \Told{} & 429 & 0.48--0.51 & 0.02 / 0.00 & 0.02 \cint{[0.01, 0.02]} & $-$0.02 / $-$0.01 \\
   & count & \Summarized{} 64 / \Told{} & 238 & 0.23--0.52 & 0.02 / 0.00 & 0.02 \cint{[0.01, 0.02]} & 0.05 / 0.07 \\
   & count & \Told{} / \Instructed{} & 168 & 0.25--0.75 & 0.00 / 0.00 & 0.00 \cint{[0.00, 0.00]} & 0.00 / 0.00 \\
   & next toss & \Shown{} 64 / \Summarized{} 64 & 251 & 0.27--0.59 & 0.08 / 0.13 & $-$0.05 \cint{[$-$0.08, $-$0.02]} & 0.25 / 0.20 \\
   & next toss & \Shown{} 64 / \Told{} & 220 & 0.28--0.63 & 0.08 / 0.03 & 0.06 \cint{[0.04, 0.07]} & 0.05 / 0.11 \\
   & next toss & \Summarized{} 64 / \Told{} & 281 & 0.24--0.76 & 0.16 / 0.04 & 0.12 \cint{[0.09, 0.14]} & $-$0.22 / $-$0.10 \\
  \midrule
  Qwen3-8B & cont. & \Shown{} 64 / \Summarized{} 64 & 128 & 0.38--0.85 & 0.37 / 0.32 & 0.05 \cint{[$-$0.02, 0.11]} & $-$0.03 / 0.01 \\
   & count & \Shown{} 64 / \Summarized{} 64 & 362 & 0.25--0.52 & 0.07 / 0.05 & 0.01 \cint{[$-$0.03, 0.04]} & $-$0.02 / $-$0.01 \\
   & count & \Shown{} 64 / \Told{} & 374 & 0.25--0.50 & 0.07 / 0.00 & 0.07 \cint{[0.06, 0.07]} & $-$0.11 / $-$0.05 \\
   & count & \Told{} / \Instructed{} & 168 & 0.25--0.75 & 0.01 / 0.00 & 0.01 \cint{[0.01, 0.02]} & 0.00 / 0.01 \\
   & next toss & \Shown{} 64 / \Summarized{} 64 & 160 & 0.19--0.50 & 0.12 / 0.26 & $-$0.14 \cint{[$-$0.18, $-$0.11]} & 0.13 / $-$0.01 \\
   & next toss & \Shown{} 64 / \Told{} & 119 & 0.23--0.50 & 0.14 / 0.15 & 0.00 \cint{[$-$0.05, 0.04]} & $-$0.04 / $-$0.04 \\
   & next toss & \Summarized{} 64 / \Told{} & 337 & 0.25--0.76 & 0.24 / 0.14 & 0.10 \cint{[0.07, 0.12]} & $-$0.12 / $-$0.02 \\
  \midrule
  Gemma-3-12B & count & \Shown{} 64 / \Summarized{} 64 & 1130 & 0.25--0.64 & 0.01 / 0.03 & $-$0.02 \cint{[$-$0.04, $-$0.01]} & 0.03 / 0.01 \\
   & count & \Summarized{} 64 / \Told{} & 230 & 0.25--0.67 & 0.05 / 0.00 & 0.05 \cint{[0.03, 0.07]} & $-$0.05 / 0.00 \\
   & next toss & \Shown{} 64 / \Summarized{} 64 & 189 & 0.40--0.70 & 0.09 / 0.15 & $-$0.07 \cint{[$-$0.09, $-$0.04]} & 0.06 / 0.00 \\
  \midrule
  Qwen3-32B & count & \Shown{} 64 / \Summarized{} 64 & 180 & 0.21--0.59 & 0.12 / 0.04 & 0.08 \cint{[0.06, 0.10]} & 0.03 / 0.11 \\
   & count & \Shown{} 64 / \Told{} & 176 & 0.25--0.59 & 0.11 / 0.05 & 0.07 \cint{[0.05, 0.08]} & 0.05 / 0.12 \\
   & count & \Summarized{} 64 / \Told{} & 342 & 0.23--0.75 & 0.05 / 0.05 & $-$0.01 \cint{[$-$0.03, 0.01]} & 0.00 / $-$0.01 \\
   & count & \Told{} / \Instructed{} & 165 & 0.25--0.75 & 0.05 / 0.00 & 0.05 \cint{[0.04, 0.07]} & $-$0.03 / 0.02 \\
   & next toss & \Shown{} 64 / \Summarized{} 64 & 192 & 0.48--0.67 & 0.10 / 0.10 & 0.00 \cint{[$-$0.05, 0.03]} & 0.02 / 0.02 \\
   & next toss & \Shown{} 64 / \Told{} & 195 & 0.40--0.78 & 0.10 / 0.24 & $-$0.14 \cint{[$-$0.18, $-$0.11]} & 0.12 / $-$0.02 \\
   & next toss & \Summarized{} 64 / \Told{} & 369 & 0.25--0.76 & 0.12 / 0.28 & $-$0.16 \cint{[$-$0.20, $-$0.13]} & 0.04 / $-$0.12 \\
  \midrule
  Gemma-3-27B & count & \Told{} / \Instructed{} & 168 & 0.25--0.75 & 0.05 / 0.00 & 0.05 \cint{[0.03, 0.07]} & $-$0.04 / 0.00 \\
   & next toss & \Shown{} 64 / \Summarized{} 64 & 158 & 0.37--0.75 & 0.15 / 0.24 & $-$0.08 \cint{[$-$0.12, $-$0.05]} & 0.00 / $-$0.08 \\
   & next toss & \Shown{} 64 / \Told{} & 203 & 0.37--0.75 & 0.16 / 0.31 & $-$0.15 \cint{[$-$0.18, $-$0.12]} & 0.03 / $-$0.12 \\
   & next toss & \Summarized{} 64 / \Told{} & 356 & 0.23--0.75 & 0.25 / 0.32 & $-$0.08 \cint{[$-$0.11, $-$0.05]} & 0.00 / $-$0.07 \\
  \midrule
  Qwen3.8-27B & cont. & \Shown{} 64 / \Summarized{} 64 & 141 & 0.10--0.93 & 0.33 / 0.63 & $-$0.30 \cint{[$-$0.34, $-$0.26]} & 0.27 / $-$0.03 \\
   & count & \Shown{} 64 / \Summarized{} 64 & 276 & 0.26--0.76 & 0.10 / 0.04 & 0.06 \cint{[0.05, 0.07]} & $-$0.01 / 0.06 \\
   & count & \Shown{} 64 / \Told{} & 234 & 0.28--0.76 & 0.11 / 0.16 & $-$0.05 \cint{[$-$0.07, $-$0.04]} & 0.02 / $-$0.04 \\
   & count & \Summarized{} 64 / \Told{} & 291 & 0.26--0.75 & 0.04 / 0.15 & $-$0.11 \cint{[$-$0.13, $-$0.10]} & 0.03 / $-$0.08 \\
   & count & \Told{} / \Instructed{} & 115 & 0.33--0.75 & 0.14 / 0.00 & 0.14 \cint{[0.13, 0.15]} & $-$0.07 / 0.06 \\
   & next toss & \Shown{} 64 / \Summarized{} 64 & 238 & 0.23--0.75 & 0.11 / 0.04 & 0.07 \cint{[0.06, 0.08]} & 0.00 / 0.06 \\
   & next toss & \Shown{} 64 / \Told{} & 257 & 0.28--0.75 & 0.11 / 0.25 & $-$0.14 \cint{[$-$0.16, $-$0.13]} & 0.10 / $-$0.04 \\
   & next toss & \Summarized{} 64 / \Told{} & 307 & 0.24--0.74 & 0.05 / 0.25 & $-$0.20 \cint{[$-$0.22, $-$0.19]} & 0.11 / $-$0.10 \\
  \bottomrule
\end{tabular}
\begin{tabnotes}
  \item \hspace*{-\labelsep}cont.: the sampling channel; count: the expected-count question; next toss: the next-toss question; 64, 16: observations in the record or summary.
  \item \hspace*{-\labelsep}$^\dagger$: answer mass below 0.3.
\end{tabnotes}
\end{threeparttable}
\end{table*}

\paragraph{States with equal estimates move apart.}
In the smaller models, under both questions, states of different routes whose current estimates agree within two points no longer agree after the same sixteen observations (Table~\ref{tab:matched}, Figure~\ref{fig:matched}). The difference is clearest in the spread. Under the next-toss question a summary of 64 observations moves 1.6 to 2.2 times as far as a record with the same estimate, and under the expected-count question a told rate stays where it is while a record or a summary with the same estimate moves. In Llama-3.1-8B's sampling channel a steered state, written onto a record of sixteen, moves about as far as a natural record of sixteen with the same estimate (difference in spread 0.03 [$-$0.01, 0.07]) and farther than a record of 64, as the record of sixteen itself does: the write leaves the response to the segment where it was (Section~\ref{sec:writes}), and the answer mass of both is below 0.3.

Three pairs are not separated: a told rate and an instruction under the expected-count question in Llama-3.1-8B and Qwen3-8B, which both hold (equivalent within the wording floor); in Qwen3-8B a record and a summary under that query (equivalent at a matching tolerance of 0.05); and in Qwen3-8B's sampling channel a record and a summary, which are undecided.

In the larger models a record and a summary with the same next-toss estimate part in two of the three. Among states of a record and a summary of 64 whose estimates agree within 0.02, the two move equally far in Qwen3-32B (spreads of 0.096 and 0.101, difference $-0.004$ [$-0.046$, 0.031]), the summary moves farther in Gemma-3-27B (0.153 and 0.238, $-0.084$ [$-0.118$, $-0.050$]) and the record moves farther in Qwen3.8-27B (0.110 and 0.044, 0.066 [0.056, 0.076]). Under the expected-count question in Qwen3-32B, a told rate moves as far as its matched summary (difference $-0.009$ [$-0.028$, 0.011]) and less far than its matched record (0.066 [0.047, 0.083]). In Qwen3.8-27B a told rate is the state that moves farthest: farther than its matched record by 0.05 and 0.14 under the expected-count and the next-toss question, and farther than its matched summary by 0.11 and 0.20, while an instruction hardly moves (spreads of 0.00 and 0.02). In this model's sampling channel the order of record and summary is the reverse of that in its estimate: a summary moves almost twice as far as a record with the same continuation (0.63 against 0.33).

\paragraph{Retention on the matched scale follows the readout.}
The slope of the estimate after the segment on the current estimate, across the states of one route, $\tilde R$, is the retention of the route on the scale of its own estimates. In the smaller models its order across routes under the next-toss question follows the weights of that question (rank correlation 0.77, 0.89 and 0.54 in Llama-3.1-8B, Qwen3-8B and Gemma-3-12B, and 0.49, 0.20 and 0.77 under the expected-count question). No single order of routes holds across readouts: under the expected-count question a told rate keeps more than a summary in Llama-3.1-8B and Qwen3-8B, and under the next-toss question Qwen3-8B keeps more of a record than of a summary or a told rate. Under the expected-count question the current estimate of a record hardly varies with the record (Llama-3.1-8B, a slope of 0.02), so its retention on this scale is not defined.

\paragraph{Answers that are counts.}
Under the next-toss question, after a summary and a new segment, Llama-3.1-8B gives part of its answers as counts (the integers up to 20 carry 40 to 47\% of the integer mass, against 6 to 16\% after a record), which lowers its estimate. Restricted to answers of 21 and above, the difference between a summary and a record with the same estimate falls to a half and to a third (0.12 and 0.06 in place of 0.25 and 0.20) and the difference in spread is unchanged. In Qwen3-8B such answers are as frequent after a record as after a summary (about a fifth of the mass), and Gemma-3-12B gives almost none.
\section{Visibility of the record}
\label{app:visibility}

This test keeps a record and its cache intact and takes part of the record out of view, in Llama-3.1-8B's sampling channel. For a third of the records of 64 observations of Section~\ref{sec:weight} (three proportions, twenty families each, worth 14.4 observations here against 17 on all of them), we hide the first $h$ outcomes from every position after the record, the new segment and the query, through the attention mask, while the record's own positions are computed as usual. With $h = 0$ the masked readout equals the ordinary continuation exactly.

\begin{figure}[t]
  \centering
  \includegraphics[width=\columnwidth]{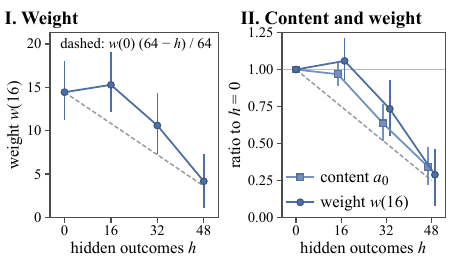}
  \caption{\textbf{Content and weight fall together as the record is hidden.} Llama-3.1-8B. I. The weight of a record of 64 observations under sixteen new ones, against the number $h$ of its first outcomes hidden from the positions after the record, dashed where it is proportional to the visible outcomes. II. The content slope $a_0$ and the weight as ratios to $h = 0$. Family-bootstrap 95\% intervals.}
  \label{fig:visibility}
\end{figure}

Hiding the first sixteen outcomes changes neither the content nor the weight, though the response to the new segment falls a little at every $h$ (Table~\ref{tab:visibility}, Figure~\ref{fig:visibility}). Hiding thirty-two or forty-eight lowers both, and in the same proportion: with forty-eight hidden the content keeps 0.34 of its slope and the weight 0.29 of its value, 4.2 observations against 6.6 for a natural record of sixteen. The later outcomes were computed with the earlier ones in view, yet the readout needs to attend to the earlier outcomes themselves for their content and their weight to count. The ablation lowers content and weight together, as the re-reading account predicts, and does not test for a store of weight apart from the content.

\begin{table}[t]
  \centering
  \footnotesize
  \caption{\textbf{Hiding the beginning of the record lowers content and weight together.} Llama-3.1-8B's document-frame continuation after a record of 64 observations whose first $h$ outcomes are hidden from the positions after the record. $a_0$: slope of the current continuation on the proportion of the record. $S$, $w$: slope on the new evidence and weight under sixteen new observations. Ratios to $h = 0$ from paired resamples. Family-bootstrap 95\% intervals.}
  \label{tab:visibility}
  \setlength{\tabcolsep}{3.5pt}
  \begin{tabular}{@{}r|l|r|l|r|r@{}}
    \toprule
    $h$ & $a_0$ & $S$ & $w$ & $w$ ratio & $a_0$ ratio \\
    \midrule
    0  & 0.70 \cint{[0.59, 0.80]} & 0.37 & 14.4 \cint{[11.3, 18.0]} & 1.00 & 1.00 \\
    16 & 0.67 \cint{[0.58, 0.76]} & 0.28 & 15.3 \cint{[12.2, 19.0]} & 1.06 & 0.97 \\
    32 & 0.44 \cint{[0.36, 0.53]} & 0.27 & 10.6 \cint{[7.3, 14.2]} & 0.73 & 0.64 \\
    48 & 0.24 \cint{[0.16, 0.32]} & 0.30 & 4.2 \cint{[1.1, 7.2]} & 0.29 & 0.34 \\
    \bottomrule
  \end{tabular}
\end{table}
\section{Predictions fixed before the larger models were run}
\label{app:predictions}

Seven predictions, P1 to P7, were derived from the smaller models and committed with their criteria to the project repository the day before the first larger model was run (Table~\ref{tab:predictions}). The criteria are met in 8 of the 21 cells, not met in 6, and cannot be evaluated in 7, where the model has no sampling channel, the answer mass is below 0.3 or the reference is undefined. The two predictions on what sets the weight, the question (P5) and the route at matched estimates (P7), hold in five of six cells, once with the opposite sign. The five predictions on the form of the dependence (P1 to P4 and P6) hold in three of the eight cells that can be evaluated. P4 fails in Gemma-3-27B by its criterion, on a sampling channel that is close to deterministic and whose weights are otherwise not read (Appendix~\ref{app:estimators}).

Two criteria were fixed in the same way for Qwen3-8B before its writes were run. The criterion of P6 is met by three of its four configurations. A bound on the shift of the estimate, a quarter of the difference between the natural records when sixteen positions are written, is met at a quarter of the depth (0.01 [0.00, 0.02] against a difference of 0.21) and missed at half the depth, where the interval of the shift, [0.03, 0.06], reaches above the bound of 0.05 (Appendix~\ref{app:writes}).

\begin{table*}[t]
  \centering
  \footnotesize
  \caption{\textbf{The predictions fixed before the larger models were run, and their outcome in each larger model.} Each cell gives the verdict and the number that decides it, with its family-bootstrap 95\% interval in grey below.}
  \label{tab:predictions}
  \providecommand{\cint}[1]{{\fontsize{7}{8}\selectfont #1}}
\providecolor{predgrey}{gray}{0.42}
\providecommand{\pgrey}[1]{\leavevmode{\fontsize{7}{8}\selectfont\color{predgrey}#1}}
\providecommand{\pmet}{\leavevmode{\usefont{U}{pzd}{m}{n}\char51}}
\providecommand{\pnot}{\leavevmode{\usefont{U}{pzd}{m}{n}\char55}}
\providecommand{\pna}{\leavevmode{\color{predgrey}---}}
\providecommand{\pmark}[1]{\makebox[1.15em][l]{#1}}
\providecommand{\predrule}{\arrayrulecolor{black!18}\specialrule{0.3pt}{2.5pt}{2.5pt}\arrayrulecolor{black}}
\providecommand{\pnext}[1]{\newline\hspace*{1.15em}\pgrey{#1}}
\begin{threeparttable}
\begin{tabular}{@{}l@{\hspace{4pt}}>{\raggedright\arraybackslash}p{0.215\textwidth}@{\hspace{8pt}}>{\raggedright\arraybackslash}p{0.19\textwidth}@{\hspace{9pt}}>{\raggedright\arraybackslash}p{0.16\textwidth}@{\hspace{9pt}}>{\raggedright\arraybackslash}p{0.16\textwidth}@{\hspace{9pt}}>{\raggedright\arraybackslash}p{0.16\textwidth}@{}}
  \toprule
  & Prediction & Criterion & Qwen3-32B & Gemma-3-27B & Qwen3.8-27B \\
  \midrule
  \multicolumn{6}{@{}l}{\textit{What sets the weight}\hspace{0.8em}\pgrey{met in 5 of 6 cells}} \\
  \addlinespace[2pt]
  P5 & The question changes a told rate's weight & $w_{\mathrm{Q2}} < w_{\mathrm{Q1}}$ & \pmark{\pmet}77 $\to$ 5.5 & \pmark{\pmet}95 $\to$ 6.0 & \pmark{\pmet}11 $\to$ 4.2 \\
  \predrule
  P7 & A record and a summary of 64 with matched estimates move apart & Q2: $\Delta\,\mathrm{spread} \neq 0$ & \pmark{\pnot}$-$0.004\pnext{[$-$0.046, 0.031]} & \pmark{\pmet}$-$0.084\pnext{[$-$0.118, $-$0.050]} & \pmark{\pmet}$+$0.066\tnote{d}\pnext{[0.056, 0.076]} \\
  \midrule
  \multicolumn{6}{@{}l}{\textit{The form of the dependence}\hspace{0.8em}\pgrey{met in 3 of the 8 cells that can be evaluated}} \\
  \addlinespace[2pt]
  P1 & A told rate enters the estimate and stays out of the document continuation & $a_0^{\mathrm{est}} > 0.8$ in every readable frame; $|a_0^{\mathrm{cont}}| < 0.15$, $0 \in \mathrm{CI}$ & \pmark{\pnot}$a_0^{\mathrm{est}}$ 0.73\tnote{e}\pnext{$a_0^{\mathrm{cont}}$ $-$0.02}\pnext{[$-$0.09, 0.05]} & \pmark{\pmet}$a_0^{\mathrm{est}}$ 0.99\tnote{e}\pnext{$a_0^{\mathrm{cont}}$ $-$0.07}\pnext{[$-$0.45, 0.29]} & \pna\tnote{b} \\
  \predrule
  P2 & A record enters the continuation more than the estimate & $a_0^{\mathrm{cont}} - a_0^{\mathrm{est}} > 0$ at $n = 64$ & \pna\tnote{a} & \pna\tnote{b} & \pmark{\pmet}0.41\pnext{[0.37, 0.45]} \\
  \predrule
  P3 & A summary's weight grows less than its count & $w^{\mathrm{est}}(64) / w^{\mathrm{est}}(8) < 8$ & \pmark{\pnot}5.4\pnext{[2.7, 13.9]} & \pna\tnote{b} & \pmark{\pnot}15.0\pnext{[10.0, 21.8]} \\
  \predrule
  P4 & A record is worth a fraction of its length & $w^{\mathrm{cont}}(64) < 32$ & \pna\tnote{a} & \pmark{\pnot}114\pnext{[85, 159]} & \pmark{\pmet}21\pnext{[20, 23]} \\
  \predrule
  P6 & A one-shot write is erased by evidence & $R_{\mathrm{write}}(16) < \tfrac{1}{2} R_{\mathrm{nat}}(16)$ & \pna\tnote{a} & \pna\tnote{c} & \pmark{\pnot}$R_{\mathrm{write}}$ 0.34--0.61\pnext{$R_{\mathrm{nat}}$ 0.29}\pnext{[0.26, 0.32]} \\
  \bottomrule
\end{tabular}
\begin{tabnotes}
  \item \hspace*{-\labelsep}\pmet\ criterion met, \pnot\ not met, \pna\ cannot be evaluated. P2--P4, P6 and P7 are met when the whole 95\% interval satisfies the inequality (P6: the interval of every write, against half the natural value); P1 and P5 compare point estimates.
  \item \hspace*{-\labelsep}$a_0$: the slope of a readout on the content; est, cont: the estimate and the continuation; nat: the natural record; $w(n)$: the weight of a summary or record of $n$; Q1, Q2: the expected-count and the next-toss question.
  \item[a] No sampling channel.
  \item[b] Answer mass below 0.3 in a readout the criterion needs: the document-frame continuation (P1, $<$0.01), the estimate of a record of 64 (P2, 0.04) and the estimate of a summary of 8 after the segment (P3, 0.05).
  \item[c] The natural record of sixteen has no content in the continuation, so $R_{\mathrm{nat}}$ is undefined.
  \item[d] Opposite sign: the record moves farther than the summary.
  \item[e] Document frame, the lower of the two (answer mass 0.31 in Qwen3-32B and 0.73 in Gemma-3-27B); the assistant frame gives 0.90 and 1.00.
\end{tabnotes}
\end{threeparttable}
\end{table*}

\section{The largest models}
\label{app:further}

\paragraph{Models and readout.}
The largest models extend the report channel to 397B parameters: Qwen3.5-122B and Qwen3.5-397B \citep{qwenteam2026qwen35} and Qwen3-235B \citep{yang2025qwen3}. All three are mixture-of-experts models, with 10, 17 and 22B active parameters. The two Qwen3.5 models have the hybrid attention of Qwen3.8-27B, three linear-attention layers with a recurrent state for each full-attention layer (Appendix~\ref{app:models}), and Qwen3-235B, read in its instruct release of July 2025, is attention-only. The models are served through an API, so they are read through their estimates alone, from sampled answers.

The prompt is the assistant frame of the report channel, in which the assistant turn begins with the question and the model continues it (Appendix~\ref{app:prompts}), and reasoning modes are switched off. Each prompt is sampled four times at temperature 1, and each prompt of a record or a summary of 64 observations thirty-two times. The states are those of the matched design (Appendix~\ref{app:matched}), with one order of each new segment, and Qwen3.5-397B is read on the records and summaries of 64 observations alone. The estimate of a state is the mean of its read answers, a cell is read where at least 0.3 of its answers are, and weights and intervals are computed as in the main text.

\paragraph{Reading an answer.}
An answer is read as the number it opens with, after any leading whitespace: a value from 0 to 100, with its decimal fraction. An answer that continues into an operator, that opens with text, or that states a probability below 1 is not read (Appendix~\ref{app:estimators}). The decimal fraction matters under the expected-count question, which these models answer with the exact value, such as 32.8125 for 21 heads in 64 tosses. Restricted to whole numbers, as the local readout is, the same readout keeps 54 to 91\% of the answers after a summary and changes none of the weights of Table~\ref{tab:further} by more than 4\%.

\paragraph{Agreement with the local readout.}
Qwen3-32B is served in the same way, which sets the sampled readout against the local one on identical prompts, with whole numbers read in both. The two agree. Over all states the sampled and the local estimates correlate at 0.93 and 0.96 under the expected-count and the next-toss question, and their root-mean-square difference, 0.056 and 0.044, is what sampling alone produces (0.069 and 0.047). On the records and summaries of 64 observations under the next-toss question the correlations are 0.98 and 1.00, and the weights follow: 54 and 41 observations against 53 and 42 in the local readout. The weights and the numerical precision of a served model cannot be inspected, and the largest models are read as served.

\begin{figure*}[t]
  \centering
  \includegraphics[width=\textwidth]{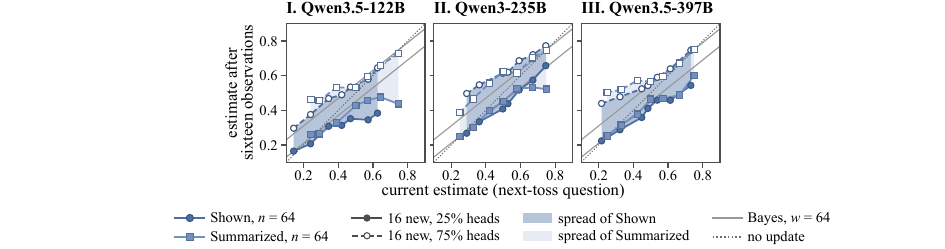}
  \caption{\textbf{Matched estimates in the largest models.} As Figure~\ref{fig:matched-main}, from sampled answers: the estimate after sixteen new observations against the current estimate under the next-toss question, for a record and a summary of 64 observations. Each point is the mean over the eight states of one content level.}
  \label{fig:further-matched}
\end{figure*}

\paragraph{A record and its summary.}
Table~\ref{tab:further} gives the weights of a record and of a summary of 64 observations beside those of the six local models on the same states, and Table~\ref{tab:further-matched} sets the two against each other. The weights of the local models on these states agree within a factor of 1.4 with those of the rate experiments, which are measured at three content levels (Table~\ref{tab:queries}). Under the next-toss question their weights agree in all three models, with ratios of 0.95 to 1.13 whose intervals all include 1. States with the same current estimate also move nearly alike: over the range of current estimates that the two routes share, the summary moves 0.85 to 0.98 times as far as the record (Figure~\ref{fig:further-matched}), where it moves 1.6 to 2.2 times as far in four of the six local models.

The two forms agree with each other and still fall short of the evidence. The estimate responds to the sixteen new observations with a slope of 0.32 to 0.38, where a Bayesian observer of 64 observations has 0.20, and it keeps 0.50 to 0.65 of the old proportion, where the observer keeps 0.78. The two hybrid models and the attention-only Qwen3-235B do not differ in this.

Under the expected-count question the weights differ in all three models. The summary is worth 1.3 and 2.9 times its record in Qwen3.5-122B and Qwen3-235B, and 0.44 times in Qwen3.5-397B. The answers behind these means are of several kinds. After a summary and a new segment, 36 and 74\% of the answers of the first two models repeat the stated proportion unchanged, and the others pool the two counts, apply Laplace's rule, follow the new segment or give another value. Qwen3.5-397B repeats the stated proportion in 17\% of its answers, and its estimate follows the new segment with a slope of 0.42 and the summary with 0.23. The weight of a summary under this question is thus an average over answers that ignore the new observations and answers that use them.

\begin{table*}[t]
  \centering
  \footnotesize
  \caption{\textbf{A record against its summary in the largest models.} For 64 observations under each question: the ratio of the two weights, with its position between one eighth and eight on the track (tick: equal weights); the spread of each route, the estimate after sixteen observations with 75\% heads minus that after 25\%, averaged over the range of current estimates that the two routes share, with the ratio of the two spreads; and the number of pairs of states whose current estimates differ by less than 0.02, with the difference of their spreads. Ratios are summary over record, and the difference is summary minus record. Family-bootstrap 95\% intervals.}
  \label{tab:further-matched}
  \setlength{\tabcolsep}{5pt}
  \providecommand{\cint}[1]{{\fontsize{7}{8}\selectfont #1}}
\providecommand{\wtrackunit}{0.15cm}\providecommand{\wtrack}[5]{\tikz[baseline=-0.55ex, x=\wtrackunit, y=1ex]{\fill[black!7] (-3,-0.62) rectangle (3,0.62);
    \draw[black!55, line width=0.5pt] (0,-0.62) -- (0,0.62);
    \draw[#4, line width=0.7pt] (#2,0) -- (#3,0);
    \if#5c\fill[#4] (#1,0) circle (0.42ex);\fi
    \if#5s\fill[#4] ([xshift=-0.4ex,yshift=-0.4ex]#1,0) rectangle ++(0.8ex,0.8ex);\fi
    \if#5d\fill[#4] ([yshift=0.55ex]#1,0) -- ([xshift=0.45ex]#1,0) -- ([yshift=-0.55ex]#1,0) -- ([xshift=-0.45ex]#1,0) -- cycle;\fi
    \if#5r\draw[#4, line width=0.7pt, ->] (1.6,0) -- (3,0);\fi
    \if#5l\draw[#4, line width=0.7pt, ->] (-1.6,0) -- (-3,0);\fi}}
\providecommand{\notrack}{\tikz[baseline=-0.55ex, x=\wtrackunit, y=1ex]{\fill[black!7] (-3,-0.62) rectangle (3,0.62);}}
\begin{tabular}{@{}ll l ccl rl@{}}
  \toprule
  & & & \multicolumn{3}{c}{Spread over the shared range} & \multicolumn{2}{c}{Pairs within 0.02} \\
  \cmidrule(lr){4-6} \cmidrule(l){7-8}
  Model & Question & Weight ratio & record & summary & ratio & pairs & difference \\
  \midrule
  Qwen3.5-122B & count & 1.31 \cint{[1.03, 1.69]}\,\,\wtrack{0.39}{0.04}{0.76}{black!70}{d} & 0.162 & 0.112 & 0.69 \cint{[0.58, 0.82]} & 183 & $-$0.032 \cint{[$-$0.055, $-$0.008]} \\
   & next toss & 0.95 \cint{[0.74, 1.24]}\,\,\wtrack{-0.08}{-0.43}{0.31}{black!70}{d} & 0.191 & 0.165 & 0.86 \cint{[0.76, 1.00]} & 186 & $-$0.017 \cint{[$-$0.046, $+$0.014]} \\
  \addlinespace
  Qwen3-235B & count & 2.95 \cint{[2.07, 4.15]}\,\,\wtrack{1.56}{1.05}{2.05}{black!70}{d} & 0.171 & 0.068 & 0.40 \cint{[0.30, 0.49]} & 183 & $-$0.110 \cint{[$-$0.143, $-$0.078]} \\
   & next toss & 1.13 \cint{[0.86, 1.51]}\,\,\wtrack{0.18}{-0.21}{0.59}{black!70}{d} & 0.184 & 0.157 & 0.85 \cint{[0.72, 1.00]} & 212 & $-$0.002 \cint{[$-$0.037, $+$0.032]} \\
  \addlinespace
  Qwen3.5-397B & count & 0.44 \cint{[0.31, 0.60]}\,\,\wtrack{-1.20}{-1.70}{-0.74}{black!70}{d} & 0.187 & 0.186 & 0.99 \cint{[0.90, 1.09]} & 171 & $+$0.010 \cint{[$-$0.013, $+$0.036]} \\
   & next toss & 0.95 \cint{[0.68, 1.31]}\,\,\wtrack{-0.08}{-0.55}{0.39}{black!70}{d} & 0.174 & 0.171 & 0.98 \cint{[0.84, 1.13]} & 181 & $-$0.028 \cint{[$-$0.062, $+$0.002]} \\
  \bottomrule
\end{tabular}
\end{table*}

\paragraph{All routes.}
Table~\ref{tab:further-weights} gives the weights of every route of the matched design in the two models read on all states. The question changes the weight of a told rate in both: from the expected-count to the next-toss question it falls by factors of 4.4 and 1.9, as it falls in the six local models (Section~\ref{sec:weight}). Neither model revises an instruction under either question. A fourfold stated count multiplies the weight of a summary by 3.2 to 4.8, and a record of 16 observations is worth 3.5 to 14.

\begin{table*}[t]
  \centering
  \footnotesize
  \caption{\textbf{The report channel of the largest models.} Weight $w$ of each route under sixteen new observations, by question, with family-bootstrap 95\% intervals. Track: $w$ of a record or a summary relative to the Bayesian $w = n$, on a log scale from one eighth to eight times (tick: $w = n$).}
  \label{tab:further-weights}
  \setlength{\tabcolsep}{4pt}
  \begin{threeparttable}
  \providecommand{\cint}[1]{{\fontsize{7}{8}\selectfont #1}}
\providecommand{\wtrackunit}{0.15cm}\providecommand{\wtrack}[5]{\tikz[baseline=-0.55ex, x=\wtrackunit, y=1ex]{\fill[black!7] (-3,-0.62) rectangle (3,0.62);
    \draw[black!55, line width=0.5pt] (0,-0.62) -- (0,0.62);
    \draw[#4, line width=0.7pt] (#2,0) -- (#3,0);
    \if#5c\fill[#4] (#1,0) circle (0.42ex);\fi
    \if#5s\fill[#4] ([xshift=-0.4ex,yshift=-0.4ex]#1,0) rectangle ++(0.8ex,0.8ex);\fi
    \if#5d\fill[#4] ([yshift=0.55ex]#1,0) -- ([xshift=0.45ex]#1,0) -- ([yshift=-0.55ex]#1,0) -- ([xshift=-0.45ex]#1,0) -- cycle;\fi
    \if#5r\draw[#4, line width=0.7pt, ->] (1.6,0) -- (3,0);\fi
    \if#5l\draw[#4, line width=0.7pt, ->] (-1.6,0) -- (-3,0);\fi}}
\providecommand{\notrack}{\tikz[baseline=-0.55ex, x=\wtrackunit, y=1ex]{\fill[black!7] (-3,-0.62) rectangle (3,0.62);}}
\begin{tabular}{@{}ll cccc ll r@{}}
  \toprule
  & & \multicolumn{2}{c}{Shown} & \multicolumn{2}{c}{Summarized} & & & \\
  \cmidrule(lr){3-4} \cmidrule(lr){5-6}
  Model & Question & $n = 16$ & $n = 64$ & $n = 16$ & $n = 64$ & Told & Instructed & Read \\
  \midrule
  Qwen3.5-122B & count & \makebox[1.9em][r]{3.5}\,\wtrack{-2.20}{-3.00}{-1.27}{rtShown}{c} & \makebox[1.9em][r]{28}\,\wtrack{-1.18}{-1.42}{-0.97}{rtShown}{c} & \makebox[1.9em][r]{9.1}\,\wtrack{-0.82}{-1.24}{-0.46}{rtSummarized}{s} & \makebox[1.9em][r]{37}\,\wtrack{-0.79}{-1.07}{-0.51}{rtSummarized}{s} & 73 \cint{[54, 107]} & $\infty$ & 0.96 \\
   & next toss & \makebox[1.9em][r]{3.9}\,\wtrack{-2.05}{-3.00}{-1.15}{rtShown}{c} & \makebox[1.9em][r]{23}\,\wtrack{-1.50}{-1.80}{-1.23}{rtShown}{c} & \makebox[1.9em][r]{4.5}\,\wtrack{-1.83}{-3.00}{-1.14}{rtSummarized}{s} & \makebox[1.9em][r]{21}\,\wtrack{-1.58}{-1.84}{-1.34}{rtSummarized}{s} & 17 \cint{[11, 24]} & $\infty$ \cint{($\geq$\,327)} & 0.97 \\
  \addlinespace
  Qwen3-235B & count & \makebox[1.9em][r]{14}\,\wtrack{-0.14}{-0.50}{0.15}{rtShown}{c} & \makebox[1.9em][r]{34}\,\wtrack{-0.91}{-1.08}{-0.75}{rtShown}{c} & \makebox[1.9em][r]{29}\,\wtrack{0.84}{0.42}{1.27}{rtSummarized}{s} & \makebox[1.9em][r]{100}\,\wtrack{0.65}{0.19}{1.12}{rtSummarized}{s} & 7.2 \cint{[5.2, 9.4]} & $\infty$ & 0.92 \\
   & next toss & \makebox[1.9em][r]{4.0}\,\wtrack{-2.02}{-2.95}{-1.45}{rtShown}{c} & \makebox[1.9em][r]{28}\,\wtrack{-1.17}{-1.43}{-0.91}{rtShown}{c} & \makebox[1.9em][r]{10}\,\wtrack{-0.65}{-1.13}{-0.20}{rtSummarized}{s} & \makebox[1.9em][r]{32}\,\wtrack{-0.99}{-1.30}{-0.65}{rtSummarized}{s} & 3.9 \cint{[1.9, 5.8]} & $\infty$ & 0.99 \\
  \addlinespace
  Qwen3.5-397B & count & \makebox[1.9em][r]{--}\,\notrack & \makebox[1.9em][r]{20}\,\wtrack{-1.69}{-1.93}{-1.45}{rtShown}{c} & \makebox[1.9em][r]{--}\,\notrack & \makebox[1.9em][r]{8.7}\,\wtrack{-2.89}{-3.00}{-2.51}{rtSummarized}{s} & -- & -- & 0.99 \\
   & next toss & \makebox[1.9em][r]{--}\,\notrack & \makebox[1.9em][r]{27}\,\wtrack{-1.27}{-1.55}{-1.02}{rtShown}{c} & \makebox[1.9em][r]{--}\,\notrack & \makebox[1.9em][r]{25}\,\wtrack{-1.35}{-1.73}{-1.00}{rtSummarized}{s} & -- & -- & 1.00 \\
  \bottomrule
\end{tabular}
  \begin{tabnotes}
    \item $\infty$, and an arrow at the right end of a track: not revised (the interval of the sensitivity includes zero, or the weight exceeds a thousand observations). $<$1: worth less than one observation.
    \item Read: the lowest share of answers read among the cells of the row. --: not collected (Qwen3.5-397B is read on the records and summaries of 64 observations).
  \end{tabnotes}
  \end{threeparttable}
\end{table*}

\section{Direct replies and proprietary models}
\label{app:hosted}

\paragraph{The direct reply.}
A proprietary model can be read only through its reply. In the direct reply the question closes the user turn, the system turn asks for a whole number without an explanation, and the model answers (Appendix~\ref{app:prompts}). Each prompt is sampled four times at temperature 1 on all states of the matched design, and the number is read from the reply as in Appendix~\ref{app:further}. The direct reply is a readout of its own. In Qwen3-32B its estimates correlate with those of the local report channel at 0.58 and 0.73 under the expected-count and the next-toss question, and the two differ state by state by twice what sampling produces, a root-mean-square difference of 0.12 and 0.11 against 0.06 and 0.05. Weights are therefore compared within the direct reply.

\paragraph{Two open models.}
Table~\ref{tab:hosted} gives Qwen3-32B and Qwen3-235B in this readout, and Table~\ref{tab:hosted-matched} sets a record of 64 observations against its summary. Where the question stands changes what the evidence is worth. In Qwen3-32B a record and a summary of 64 observations are worth 30 and 23 under the next-toss question, against 53 and 42 in the report channel, and under the expected-count question its replies follow the evidence weakly and leave the weights poorly determined. In Qwen3-235B the summary is worth 2.7 times its record under the next-toss question and 5.7 times under the expected-count question, where the report channel of the same model weighs the two alike under the first (Table~\ref{tab:further-matched}). The position of the question joins the frame and the question itself among the conditions that set the weight (Section~\ref{sec:weight}).

\paragraph{Three proprietary models.}
The same tables add GPT-6 Sol \citep{openai2026gpt6sol} and Gemini 4 Argon \citep{google2026gemini4argon}, read with reasoning switched off, and Claude Opus 5.5 \citep{anthropic2026opus55}, whose reasoning cannot be switched off and runs at its lowest setting. Gemini 4 Argon still spends reasoning tokens on 5\% of the prompts, and Claude Opus 5.5 on 98\%, with a median of 156 tokens. We report the three as observed.

GPT-6 Sol gives a record its length, 16 to 17 and 67 to 72 observations, and a summary less, 11 to 16 and 39 to 49, so that the summary is worth 0.73 and 0.55 times its record under the two questions. Gemini 4 Argon and Claude Opus 5.5 weigh both forms near their length, at 58 to 64 and at 66 observations for 64, and in Claude Opus 5.5 the two weigh alike to within one percent. A told rate is worth 0.7 to 2.1 observations in the three models, and none of them revises an instruction. The weights, the architectures and the inference procedures of these models are not public, so a difference among them, or between them and the open models, cannot be assigned to a cause.

\begin{table*}[t]
  \centering
  \footnotesize
  \caption{\textbf{The direct reply in two open and three proprietary models.} Weight $w$ of each route under sixteen new observations, by question. Track and symbols as in Table~\ref{tab:further-weights}.}
  \label{tab:hosted}
  \setlength{\tabcolsep}{4pt}
  \begin{threeparttable}
  \providecommand{\cint}[1]{{\fontsize{7}{8}\selectfont #1}}
\providecommand{\wtrackunit}{0.15cm}\providecommand{\wtrack}[5]{\tikz[baseline=-0.55ex, x=\wtrackunit, y=1ex]{\fill[black!7] (-3,-0.62) rectangle (3,0.62);
    \draw[black!55, line width=0.5pt] (0,-0.62) -- (0,0.62);
    \draw[#4, line width=0.7pt] (#2,0) -- (#3,0);
    \if#5c\fill[#4] (#1,0) circle (0.42ex);\fi
    \if#5s\fill[#4] ([xshift=-0.4ex,yshift=-0.4ex]#1,0) rectangle ++(0.8ex,0.8ex);\fi
    \if#5d\fill[#4] ([yshift=0.55ex]#1,0) -- ([xshift=0.45ex]#1,0) -- ([yshift=-0.55ex]#1,0) -- ([xshift=-0.45ex]#1,0) -- cycle;\fi
    \if#5r\draw[#4, line width=0.7pt, ->] (1.6,0) -- (3,0);\fi
    \if#5l\draw[#4, line width=0.7pt, ->] (-1.6,0) -- (-3,0);\fi}}
\providecommand{\notrack}{\tikz[baseline=-0.55ex, x=\wtrackunit, y=1ex]{\fill[black!7] (-3,-0.62) rectangle (3,0.62);}}
\begin{tabular}{@{}ll cccc ll r@{}}
  \toprule
  & & \multicolumn{2}{c}{Shown} & \multicolumn{2}{c}{Summarized} & & & \\
  \cmidrule(lr){3-4} \cmidrule(lr){5-6}
  Model & Question & $n = 16$ & $n = 64$ & $n = 16$ & $n = 64$ & Told & Instructed & Read \\
  \midrule
  \multicolumn{9}{@{}l}{\textit{Open models}} \\
  \quad Qwen3-32B & count & \makebox[1.9em][r]{4.8}\,\wtrack{-1.73}{-3.00}{0.28}{rtShown}{c} & \makebox[1.9em][r]{45}\,\wtrack{-0.52}{-1.58}{0.93}{rtShown}{c} & \makebox[1.9em][r]{30}\,\wtrack{0.90}{-1.29}{3.00}{rtSummarized}{s} & \makebox[1.9em][r]{57}\,\wtrack{-0.18}{-1.70}{2.34}{rtSummarized}{s} & 27 \cint{[10, 73]} & $\infty$ \cint{($\geq$\,561)} & 0.94 \\
   & next toss & \makebox[1.9em][r]{6.7}\,\wtrack{-1.25}{-2.35}{-0.57}{rtShown}{c} & \makebox[1.9em][r]{30}\,\wtrack{-1.07}{-1.56}{-0.58}{rtShown}{c} & \makebox[1.9em][r]{15}\,\wtrack{-0.13}{-1.20}{0.51}{rtSummarized}{s} & \makebox[1.9em][r]{23}\,\wtrack{-1.48}{-2.03}{-0.98}{rtSummarized}{s} & 6.7 \cint{[2.4, 11]} & 333 \cint{[121, 2780]} & 1.00 \\
  \quad Qwen3-235B & count & \makebox[1.9em][r]{7.2}\,\wtrack{-1.16}{-1.58}{-0.83}{rtShown}{c} & \makebox[1.9em][r]{18}\,\wtrack{-1.82}{-2.18}{-1.52}{rtShown}{c} & \makebox[1.9em][r]{48}\,\wtrack{1.58}{1.05}{2.21}{rtSummarized}{s} & \makebox[1.9em][r]{103}\,\wtrack{0.69}{0.09}{1.51}{rtSummarized}{s} & 22 \cint{[15, 32]} & $\infty$ \cint{($\geq$\,8400)} & 0.98 \\
   & next toss & \makebox[1.9em][r]{5.0}\,\wtrack{-1.69}{-2.66}{-1.11}{rtShown}{c} & \makebox[1.9em][r]{16}\,\wtrack{-1.96}{-2.34}{-1.63}{rtShown}{c} & \makebox[1.9em][r]{18}\,\wtrack{0.20}{-0.19}{0.55}{rtSummarized}{s} & \makebox[1.9em][r]{44}\,\wtrack{-0.54}{-0.89}{-0.18}{rtSummarized}{s} & 13 \cint{[9.9, 17]} & $\infty$ \cint{($\geq$\,378)} & 1.00 \\
  \midrule
  \multicolumn{9}{@{}l}{\textit{Proprietary models}} \\
  \quad GPT-6 Sol & count & \makebox[1.9em][r]{17}\,\wtrack{0.05}{-0.15}{0.21}{rtShown}{c} & \makebox[1.9em][r]{67}\,\wtrack{0.07}{-0.04}{0.20}{rtShown}{c} & \makebox[1.9em][r]{16}\,\wtrack{-0.04}{-0.39}{0.28}{rtSummarized}{s} & \makebox[1.9em][r]{49}\,\wtrack{-0.38}{-0.61}{-0.18}{rtSummarized}{s} & 0.7 \cint{[0.0, 1.5]} & $\infty$ & 1.00 \\
   & next toss & \makebox[1.9em][r]{16}\,\wtrack{-0.04}{-0.21}{0.11}{rtShown}{c} & \makebox[1.9em][r]{72}\,\wtrack{0.16}{0.06}{0.27}{rtShown}{c} & \makebox[1.9em][r]{11}\,\wtrack{-0.54}{-0.86}{-0.25}{rtSummarized}{s} & \makebox[1.9em][r]{39}\,\wtrack{-0.71}{-0.93}{-0.49}{rtSummarized}{s} & 0.7 \cint{[0.2, 1.2]} & $\infty$ & 1.00 \\
  \quad Gemini 4 Argon & count & \makebox[1.9em][r]{16}\,\wtrack{-0.01}{-0.04}{0.02}{rtShown}{c} & \makebox[1.9em][r]{59}\,\wtrack{-0.13}{-0.17}{-0.08}{rtShown}{c} & \makebox[1.9em][r]{16}\,\wtrack{-0.04}{-0.08}{-0.01}{rtSummarized}{s} & \makebox[1.9em][r]{59}\,\wtrack{-0.12}{-0.26}{-0.03}{rtSummarized}{s} & 1.9 \cint{[1.1, 3.1]} & $\infty$ & 1.00 \\
   & next toss & \makebox[1.9em][r]{16}\,\wtrack{-0.01}{-0.04}{0.02}{rtShown}{c} & \makebox[1.9em][r]{58}\,\wtrack{-0.14}{-0.17}{-0.10}{rtShown}{c} & \makebox[1.9em][r]{16}\,\wtrack{-0.01}{-0.02}{-0.00}{rtSummarized}{s} & \makebox[1.9em][r]{64}\,\wtrack{-0.01}{-0.02}{0.02}{rtSummarized}{s} & 2.1 \cint{[1.3, 3.1]} & $\infty$ & 1.00 \\
  \quad Claude Opus 5.5$^\ddagger$ & count & \makebox[1.9em][r]{16}\,\wtrack{-0.00}{-0.00}{-0.00}{rtShown}{c} & \makebox[1.9em][r]{66}\,\wtrack{0.05}{0.05}{0.06}{rtShown}{c} & \makebox[1.9em][r]{16}\,\wtrack{-0.02}{-0.03}{-0.01}{rtSummarized}{s} & \makebox[1.9em][r]{66}\,\wtrack{0.05}{0.05}{0.05}{rtSummarized}{s} & 1.2 \cint{[0.8, 1.6]} & $\infty$ & 0.96 \\
   & next toss & \makebox[1.9em][r]{16}\,\wtrack{-0.00}{-0.00}{-0.00}{rtShown}{c} & \makebox[1.9em][r]{66}\,\wtrack{0.05}{0.04}{0.05}{rtShown}{c} & \makebox[1.9em][r]{16}\,\wtrack{-0.02}{-0.03}{-0.01}{rtSummarized}{s} & \makebox[1.9em][r]{66}\,\wtrack{0.05}{0.05}{0.05}{rtSummarized}{s} & 1.2 \cint{[0.8, 1.6]} & $\infty$ & 0.96 \\
  \bottomrule
\end{tabular}
  \begin{tabnotes}
    \item $\ddagger$: reasoning cannot be switched off and runs at its lowest setting.
  \end{tabnotes}
  \end{threeparttable}
\end{table*}

\begin{table*}[t]
  \centering
  \footnotesize
  \caption{\textbf{A record against its summary in the direct reply.} For 64 observations under each question, with the columns of Table~\ref{tab:further-matched}.}
  \label{tab:hosted-matched}
  \setlength{\tabcolsep}{3.6pt}
  \providecommand{\cint}[1]{{\fontsize{7}{8}\selectfont #1}}
\providecommand{\wtrackunit}{0.15cm}\providecommand{\wtrack}[5]{\tikz[baseline=-0.55ex, x=\wtrackunit, y=1ex]{\fill[black!7] (-3,-0.62) rectangle (3,0.62);
    \draw[black!55, line width=0.5pt] (0,-0.62) -- (0,0.62);
    \draw[#4, line width=0.7pt] (#2,0) -- (#3,0);
    \if#5c\fill[#4] (#1,0) circle (0.42ex);\fi
    \if#5s\fill[#4] ([xshift=-0.4ex,yshift=-0.4ex]#1,0) rectangle ++(0.8ex,0.8ex);\fi
    \if#5d\fill[#4] ([yshift=0.55ex]#1,0) -- ([xshift=0.45ex]#1,0) -- ([yshift=-0.55ex]#1,0) -- ([xshift=-0.45ex]#1,0) -- cycle;\fi
    \if#5r\draw[#4, line width=0.7pt, ->] (1.6,0) -- (3,0);\fi
    \if#5l\draw[#4, line width=0.7pt, ->] (-1.6,0) -- (-3,0);\fi}}
\providecommand{\notrack}{\tikz[baseline=-0.55ex, x=\wtrackunit, y=1ex]{\fill[black!7] (-3,-0.62) rectangle (3,0.62);}}
\begin{tabular}{@{}ll l ccl rl@{}}
  \toprule
  & & & \multicolumn{3}{c}{Spread over the shared range} & \multicolumn{2}{c}{Pairs within 0.02} \\
  \cmidrule(lr){4-6} \cmidrule(l){7-8}
  Model & Question & Weight ratio & record & summary & ratio & pairs & difference \\
  \midrule
  \multicolumn{8}{@{}l}{\textit{Open models}} \\
  \quad Qwen3-32B & count & 1.27 \cint{[0.35, 7.69]}\,\,\wtrack{0.34}{-1.52}{2.94}{black!70}{d} & 0.031 & 0.014 & 0.47 \cint{[$-$2.80, 4.87]} & 397 & $-$0.020 \cint{[$-$0.059, $+$0.018]} \\
   & next toss & 0.75 \cint{[0.45, 1.19]}\,\,\wtrack{-0.41}{-1.15}{0.25}{black!70}{d} & 0.085 & 0.095 & 1.11 \cint{[0.69, 1.57]} & 868 & $+$0.044 \cint{[$+$0.012, $+$0.076]} \\
  \quad Qwen3-235B & count & 5.71 \cint{[3.49, 10.50]}\,\,\wtrack{2.51}{1.81}{3.00}{black!70}{d} & 0.202 & 0.049 & 0.24 \cint{[0.16, 0.33]} & 260 & $-$0.157 \cint{[$-$0.194, $-$0.120]} \\
   & next toss & 2.67 \cint{[1.92, 3.81]}\,\,\wtrack{1.42}{0.94}{1.93}{black!70}{d} & 0.197 & 0.082 & 0.42 \cint{[0.26, 0.58]} & 279 & $-$0.100 \cint{[$-$0.140, $-$0.059]} \\
  \midrule
  \multicolumn{8}{@{}l}{\textit{Proprietary models}} \\
  \quad GPT-6 Sol & count & 0.73 \cint{[0.60, 0.87]}\,\,\wtrack{-0.45}{-0.74}{-0.20}{black!70}{d} & 0.092 & 0.120 & 1.30 \cint{[1.14, 1.48]} & 224 & $+$0.027 \cint{[$+$0.008, $+$0.047]} \\
   & next toss & 0.55 \cint{[0.47, 0.64]}\,\,\wtrack{-0.87}{-1.10}{-0.65}{black!70}{d} & 0.091 & 0.150 & 1.66 \cint{[1.49, 1.86]} & 278 & $+$0.055 \cint{[$+$0.040, $+$0.070]} \\
  \quad Gemini 4 Argon & count & 1.00 \cint{[0.91, 1.08]}\,\,\wtrack{0.00}{-0.14}{0.11}{black!70}{d} & 0.108 & 0.107 & 0.99 \cint{[0.92, 1.09]} & 432 & $-$0.002 \cint{[$-$0.009, $+$0.007]} \\
   & next toss & 1.10 \cint{[1.07, 1.13]}\,\,\wtrack{0.13}{0.09}{0.17}{black!70}{d} & 0.109 & 0.098 & 0.90 \cint{[0.88, 0.92]} & 425 & $-$0.011 \cint{[$-$0.013, $-$0.008]} \\
  \quad Claude Opus 5.5$^\ddagger$ & count & 1.00 \cint{[0.99, 1.00]}\,\,\wtrack{-0.01}{-0.02}{0.00}{black!70}{d} & 0.095 & 0.095 & 1.01 \cint{[1.00, 1.01]} & 448 & 0.000 \cint{[0.000, $+$0.001]} \\
   & next toss & 1.00 \cint{[1.00, 1.01]}\,\,\wtrack{0.00}{-0.00}{0.01}{black!70}{d} & 0.095 & 0.095 & 1.00 \cint{[1.00, 1.00]} & 448 & 0.000 \cint{[$-$0.001, 0.000]} \\
  \bottomrule
\end{tabular}
\end{table*}

\section{A reasoning step before the answer}
\label{app:reasoning}

Every estimate of the main text is a direct answer. This appendix lets three of the models reason before they answer and asks what the reasoning step changes.

\paragraph{Design.}
Qwen3-32B and Qwen3.5-122B are read in the direct reply (Appendix~\ref{app:hosted}) with their reasoning mode switched off and on. Qwen3.8-27B is read through a served endpoint that reasons before every reply, beside its local report channel. The states are the records and summaries of 16 and 64 observations, the told rates and the instructions of the matched design, before and after the two new segments, under the next-toss question and, for Qwen3.8-27B, under both questions. With reasoning on, each prompt is answered once at temperature 1, four times in Qwen3.8-27B, and the reasoning text is kept. Every trace ends on its own, the longest after 23,389 tokens.

\begin{table*}[t]
  \centering
  \footnotesize
  \caption{\textbf{The same states answered directly and after a reasoning step.} Weight $w$ of each route under sixteen new observations, under the next-toss question. The direct answers of Qwen3.8-27B are those of its local report channel. Track and symbols as in Table~\ref{tab:further-weights}.}
  \label{tab:reasoning}
  \setlength{\tabcolsep}{4pt}
  \begin{threeparttable}
  \providecommand{\cint}[1]{{\fontsize{7}{8}\selectfont #1}}
\providecommand{\wtrackunit}{0.15cm}\providecommand{\wtrack}[5]{\tikz[baseline=-0.55ex, x=\wtrackunit, y=1ex]{\fill[black!7] (-3,-0.62) rectangle (3,0.62);
    \draw[black!55, line width=0.5pt] (0,-0.62) -- (0,0.62);
    \draw[#4, line width=0.7pt] (#2,0) -- (#3,0);
    \if#5c\fill[#4] (#1,0) circle (0.42ex);\fi
    \if#5s\fill[#4] ([xshift=-0.4ex,yshift=-0.4ex]#1,0) rectangle ++(0.8ex,0.8ex);\fi
    \if#5d\fill[#4] ([yshift=0.55ex]#1,0) -- ([xshift=0.45ex]#1,0) -- ([yshift=-0.55ex]#1,0) -- ([xshift=-0.45ex]#1,0) -- cycle;\fi
    \if#5r\draw[#4, line width=0.7pt, ->] (1.6,0) -- (3,0);\fi
    \if#5l\draw[#4, line width=0.7pt, ->] (-1.6,0) -- (-3,0);\fi}}
\providecommand{\notrack}{\tikz[baseline=-0.55ex, x=\wtrackunit, y=1ex]{\fill[black!7] (-3,-0.62) rectangle (3,0.62);}}
\begin{tabular}{@{}ll cccc llr@{}}
  \toprule
  & & \multicolumn{2}{c}{Shown} & \multicolumn{2}{c}{Summarized} & & & \\
  \cmidrule(lr){3-4} \cmidrule(lr){5-6}
  Model & Answer & $n = 16$ & $n = 64$ & $n = 16$ & $n = 64$ & Told & Instructed & Read \\
  \midrule
  Qwen3.8-27B & direct & \makebox[1.9em][r]{4.6}\,\wtrack{-1.80}{-2.07}{-1.55}{rtShown}{c} & \makebox[1.9em][r]{28}\,\wtrack{-1.19}{-1.34}{-1.05}{rtShown}{c} & \makebox[1.9em][r]{23}\,\wtrack{0.49}{0.28}{0.68}{rtSummarized}{s} & \makebox[1.9em][r]{107}\,\wtrack{0.74}{0.54}{0.94}{rtSummarized}{s} & 5.5 \cint{[4.5, 6.5]} & -- & 0.94 \\
   & reasoned & \makebox[1.9em][r]{16}\,\wtrack{0.02}{-0.01}{0.05}{rtShown}{c} & \makebox[1.9em][r]{63}\,\wtrack{-0.03}{-0.08}{0.00}{rtShown}{c} & \makebox[1.9em][r]{16}\,\wtrack{0.00}{-0.02}{0.02}{rtSummarized}{s} & \makebox[1.9em][r]{67}\,\wtrack{0.06}{0.03}{0.10}{rtSummarized}{s} & 4.5 \cint{[3.2, 6.0]} & $\infty$ & 1.00 \\
  \addlinespace
  Qwen3-32B & direct & \makebox[1.9em][r]{6.7}\,\wtrack{-1.25}{-2.35}{-0.57}{rtShown}{c} & \makebox[1.9em][r]{30}\,\wtrack{-1.07}{-1.56}{-0.58}{rtShown}{c} & \makebox[1.9em][r]{15}\,\wtrack{-0.13}{-1.20}{0.51}{rtSummarized}{s} & \makebox[1.9em][r]{23}\,\wtrack{-1.48}{-2.03}{-0.98}{rtSummarized}{s} & 6.7 \cint{[2.4, 11]} & 333 \cint{[121, 2780]} & 1.00 \\
   & reasoned & \makebox[1.9em][r]{16}\,\wtrack{-0.01}{-0.06}{0.03}{rtShown}{c} & \makebox[1.9em][r]{58}\,\wtrack{-0.14}{-0.40}{0.09}{rtShown}{c} & \makebox[1.9em][r]{15}\,\wtrack{-0.07}{-0.20}{0.01}{rtSummarized}{s} & \makebox[1.9em][r]{65}\,\wtrack{0.02}{0.01}{0.03}{rtSummarized}{s} & 3.2 \cint{[1.5, 5.6]} & $\infty$ & 0.95 \\
  \addlinespace
  Qwen3.5-122B & direct & \makebox[1.9em][r]{$<$1}\,\wtrack{-3.00}{-3.00}{-2.46}{rtShown}{l} & \makebox[1.9em][r]{17}\,\wtrack{-1.87}{-2.37}{-1.47}{rtShown}{c} & \makebox[1.9em][r]{5.3}\,\wtrack{-1.59}{-2.20}{-1.10}{rtSummarized}{s} & \makebox[1.9em][r]{16}\,\wtrack{-2.03}{-2.35}{-1.74}{rtSummarized}{s} & 59 \cint{[43, 85]} & $\infty$ \cint{($\geq$\,524)} & 0.99 \\
   & reasoned & \makebox[1.9em][r]{16}\,\wtrack{0.01}{-0.01}{0.03}{rtShown}{c} & \makebox[1.9em][r]{63}\,\wtrack{-0.03}{-0.07}{0.01}{rtShown}{c} & \makebox[1.9em][r]{16}\,\wtrack{0.01}{0.01}{0.02}{rtSummarized}{s} & \makebox[1.9em][r]{65}\,\wtrack{0.01}{0.01}{0.02}{rtSummarized}{s} & 9.1 \cint{[5.6, 14]} & $\infty$ & 0.98 \\
  \bottomrule
\end{tabular}
  \begin{tabnotes}
    \item --: not collected.
  \end{tabnotes}
  \end{threeparttable}
\end{table*}

\paragraph{With reasoning the weight is the evidence.}
Table~\ref{tab:reasoning} sets the two kinds of answer side by side. Answered directly, a record of 64 observations is worth 17 to 30 in the three models and its summary 16 to 107. After a reasoning step both are worth 58 to 67, and a record or a summary of 16 observations is worth 15 to 16. The dependence of the weight on the form of the evidence, the subject of Section~\ref{sec:weight}, is gone where the evidence can be counted. A told rate has no count, and with reasoning it is worth 3 to 9 observations.

\paragraph{The reasoning writes the evidence into the context.}
The traces were coded by rules for the counts they write, the estimator they apply and the value they compute (Table~\ref{tab:traces}). Checked by hand on 96 freshly drawn traces, the coding reads the answer correctly in all, the pooling in 30 of 32 and the rule applied to a told rate in 29 of 32. The traces state the counts of the old evidence, reading them off a summary or counting them in a record. After a new segment, 94 to 99\% of the traces of a model add the two counts and compute one proportion from the pooled counts, as a relative frequency or by Laplace's rule (Figure~\ref{fig:traces}). The old evidence then enters with the weight of its written total, 16 or 64, which is the Bayesian weight. The answer is a value the trace wrote in 98 to 100\% of the traces and the last such value in 81 to 100\%: Qwen3.5-122B often names the estimator it rejects once more after it has decided.

\paragraph{The answer follows the written count.}
Counting is the fallible step. After a record of 64 observations the answer is the pooled proportion of the true counts in 64 to 79\% of the traces, against 94 to 99\% after its summary (Table~\ref{tab:traces}). The answer then follows the count the trace wrote: where the written and the true counts give different answers, it is the value of the written counts in 179 of 183 traces of Qwen3.8-27B and in 33 of 38 of Qwen3-32B. Qwen3.5-122B counts a record several times within a trace, and the count it settles on is not read. The estimate is thus read from the text the model wrote and no longer from the record: the reasoning step turns a record into a statement, the form that the estimate repeats (Section~\ref{sec:content}). The miscounts also leave the record of Qwen3-32B with the widest interval of Table~\ref{tab:reasoning}.

\paragraph{A told rate has no count to pool.}
With a told rate the context holds a statement and its source, and the reasoning must supply a model of the source. The three models supply different ones (Table~\ref{tab:told}). After a new segment Qwen3.8-27B most often answers from the observations alone, Qwen3-32B computes a posterior over a right and a wrong source, and Qwen3.5-122B mixes the stated rate and the observations by the stated accuracy of the source about as often as it answers from the observations. The weight of a told rate with reasoning is therefore an average over traces that choose different rules, where the weight of a record or a summary is fixed by its count. An instruction leaves nothing to choose: all 500 reasoned answers to an instruction give the instructed probability.

\paragraph{Cost.}
Counting is also the expensive step. After a new segment a record of 64 observations takes a median of 1,926 to 11,149 completion tokens in the three models, 2.3 to 2.8 times the tokens of its summary.

\onecolumn
\makeatletter\global\@colroom\@colht\global\vsize\@colroom\makeatother
\begin{table}[t]
  \centering
  \footnotesize
  \caption{\textbf{What a reasoning trace writes after a new segment.} Shares of traces that state the counts of the old evidence and that compute from the pooled counts; the weight that the written computation gives the old evidence, as its median and as the share of traces within one observation of $n$; the shares of answers that are a value the trace wrote and that equal the pooled proportion of the true counts; and the median number of completion tokens.}
  \label{tab:traces}
  \setlength{\tabcolsep}{5pt}
  \providecommand{\cint}[1]{{\fontsize{7}{8}\selectfont #1}}
\providecommand{\wtrackunit}{0.15cm}\providecommand{\wtrack}[5]{\tikz[baseline=-0.55ex, x=\wtrackunit, y=1ex]{\fill[black!7] (-3,-0.62) rectangle (3,0.62);
    \draw[black!55, line width=0.5pt] (0,-0.62) -- (0,0.62);
    \draw[#4, line width=0.7pt] (#2,0) -- (#3,0);
    \if#5c\fill[#4] (#1,0) circle (0.42ex);\fi
    \if#5s\fill[#4] ([xshift=-0.4ex,yshift=-0.4ex]#1,0) rectangle ++(0.8ex,0.8ex);\fi
    \if#5d\fill[#4] ([yshift=0.55ex]#1,0) -- ([xshift=0.45ex]#1,0) -- ([yshift=-0.55ex]#1,0) -- ([xshift=-0.45ex]#1,0) -- cycle;\fi
    \if#5r\draw[#4, line width=0.7pt, ->] (1.6,0) -- (3,0);\fi
    \if#5l\draw[#4, line width=0.7pt, ->] (-1.6,0) -- (-3,0);\fi}}
\providecommand{\notrack}{\tikz[baseline=-0.55ex, x=\wtrackunit, y=1ex]{\fill[black!7] (-3,-0.62) rectangle (3,0.62);}}
\begin{tabular}{@{}ll r cc cc cc r@{}}
  \toprule
  & & & \multicolumn{2}{c}{Counts} & \multicolumn{2}{c}{Weight written} & \multicolumn{2}{c}{Answer} & \\
  \cmidrule(lr){4-5} \cmidrule(lr){6-7} \cmidrule(lr){8-9}
  Model & Evidence & Traces & stated & pooled & median & at $n$ & written & correct & Tokens \\
  \midrule
  Qwen3.8-27B & record, 16 & 723 & 0.96 & 1.00 & 16 & 0.98 & 1.00 & 0.91 & 887 \\
   & record, 64 & 754 & 0.99 & 0.99 & 64 & 0.89 & 1.00 & 0.79 & 1\,926 \\
   & summary, 16 & 726 & 1.00 & 0.98 & 16 & 0.97 & 1.00 & 0.93 & 852 \\
   & summary, 64 & 743 & 1.00 & 1.00 & 64 & 0.99 & 1.00 & 0.94 & 833 \\
  \addlinespace
  Qwen3-32B & record, 16 & 112 & 1.00 & 1.00 & 16 & 0.96 & 1.00 & 0.96 & 1\,785 \\
   & record, 64 & 112 & 0.83 & 0.98 & 64 & 0.79 & 0.90 & 0.64 & 3\,876 \\
   & summary, 16 & 112 & 1.00 & 0.90 & 16 & 0.93 & 1.00 & 0.90 & 2\,114 \\
   & summary, 64 & 112 & 1.00 & 1.00 & 64 & 1.00 & 1.00 & 0.98 & 1\,378 \\
  \addlinespace
  Qwen3.5-122B & record, 16 & 112 & 0.95 & 1.00 & 16 & 0.96 & 1.00 & 0.96 & 5\,004 \\
   & record, 64 & 112 & 0.97 & 1.00 & 64 & 0.85 & 1.00 & 0.77 & 11\,149 \\
   & summary, 16 & 112 & 1.00 & 0.99 & 16 & 0.98 & 1.00 & 0.99 & 4\,455 \\
   & summary, 64 & 112 & 1.00 & 1.00 & 64 & 0.98 & 1.00 & 0.99 & 3\,928 \\
  \bottomrule
\end{tabular}
\end{table}

\begin{table}[t]
  \centering
  \footnotesize
  \caption{\textbf{The rule a trace applies to a told rate after a new segment}, as shares of each model's traces. Mixture by accuracy: the stated rate and the observations mixed by the stated accuracy of the source. Posterior over the source: over a right and a wrong source.}
  \label{tab:told}
  \setlength{\tabcolsep}{8pt}
  \providecommand{\cint}[1]{{\fontsize{7}{8}\selectfont #1}}
\providecommand{\wtrackunit}{0.15cm}\providecommand{\wtrack}[5]{\tikz[baseline=-0.55ex, x=\wtrackunit, y=1ex]{\fill[black!7] (-3,-0.62) rectangle (3,0.62);
    \draw[black!55, line width=0.5pt] (0,-0.62) -- (0,0.62);
    \draw[#4, line width=0.7pt] (#2,0) -- (#3,0);
    \if#5c\fill[#4] (#1,0) circle (0.42ex);\fi
    \if#5s\fill[#4] ([xshift=-0.4ex,yshift=-0.4ex]#1,0) rectangle ++(0.8ex,0.8ex);\fi
    \if#5d\fill[#4] ([yshift=0.55ex]#1,0) -- ([xshift=0.45ex]#1,0) -- ([yshift=-0.55ex]#1,0) -- ([xshift=-0.45ex]#1,0) -- cycle;\fi
    \if#5r\draw[#4, line width=0.7pt, ->] (1.6,0) -- (3,0);\fi
    \if#5l\draw[#4, line width=0.7pt, ->] (-1.6,0) -- (-3,0);\fi}}
\providecommand{\notrack}{\tikz[baseline=-0.55ex, x=\wtrackunit, y=1ex]{\fill[black!7] (-3,-0.62) rectangle (3,0.62);}}
\begin{tabular}{@{}l rrr@{}}
  \toprule
  & Qwen3.8 & Qwen3 & Qwen3.5 \\
  Rule & 27B & 32B & 122B \\
  \midrule
  Observations alone & 0.54 & 0.26 & 0.43 \\
  Mixture by accuracy & 0.21 & 0.10 & 0.47 \\
  Posterior over the source & 0.18 & 0.46 & 0.06 \\
  Rate as prior observations & 0.05 & 0.17 & 0.01 \\
  Rate kept & 0.01 & 0.00 & 0.03 \\
  \midrule
  Traces & 896 & 112 & 112 \\
  \bottomrule
\end{tabular}
\end{table}

\begin{figure}[!ht]
\begin{promptrow}[raster columns=3, raster before skip=0pt, raster after skip=0pt]
\begin{tracebox}{Pooling}{34}
So there are 4 heads in 16 new tosses. So combining the original 7 heads with these 4 new ones gives a total of 11 heads. The total number of tosses is 16 (original) + 16 (new) = 32 tosses.

Therefore, the estimated probability p is 11/32. Let's compute that. 11 divided by 32 equals 0.34375. Converting that to a percentage: 34.375\%. Rounded to the nearest whole number, that would be 34\%.
\end{tracebox}
\begin{tracebox}{A miscount}{41}
That's 26 Hs. Then positions 63 and 64 are T. So total number of Hs is 26, and number of Ts is 64 - 26 = 38.
\ph{\dots}
In this case, k=26, n=64. So (26 + 1)/(64 + 2) = 27/66 = same as before. So yes, 27/66 = 0.4091, 40.91\%, so 41\% when rounded to the nearest whole number.
\end{tracebox}
\begin{tracebox}{A told rate}{26}
So, putting it together: 95\% chance the true probability is 0.25, leading to a 25\% chance of H, and 5\% chance the true probability is something else, leading to 50\% chance of H. So the total probability would be (0.95 * 0.25) + (0.05 * 0.50) = 0.2375 + 0.025 = 0.2625, which is 26.25\%. Rounded to the nearest whole number would be 26\%.
\end{tracebox}
\end{promptrow}
  \captionsetup{position=below}\caption{\textbf{Three reasoning traces of Qwen3-32B}, verbatim, with omissions marked \ph{\dots}. Left: the trace pools the counts of a record of 16 observations and of a new segment. Middle: the trace miscounts a record of 64 observations with 27 heads and answers from the count it wrote, where the true count gives 42. Right: the trace handles a rate of 0.25 told by a source that is accurate 95\% of the time.}
  \label{fig:traces}
\end{figure}
\makeatletter\if@twocolumn\onecolumn\global\@colroom\@colht\global\vsize\@colroom\fi\makeatother
\section{Prompts}
\label{app:prompts}
\raggedbottom
\advance\interlinepenalty by 10000

Every state is a short document: the task description, one content line set by the route, and a boundary line, read by a query on a line of its own, with new evidence inserted before the boundary line. Generated parts are shown in angle brackets. \Steered{} and \Transplanted{} states are states of the textual routes and add no text of their own.

\paragraph{Task and routes.}
The task description opens every state except the prose list and the bare stream, and the content line of the route follows it on the next line.

\nopagebreak
\begin{promptbox}[note={A state with no content is this line followed by the boundary line.}]{Task description}
A coin has an unknown fixed probability of heads. H means heads and T means tails. Tosses are independent given that probability.
\end{promptbox}

\begin{promptbox}[labels=4, note={The $n$ outcomes are written as H or T, separated by single spaces, on the line of the wording and without a final period. A family's wording is its replicate index modulo four; the other three wordings of the same record give the noise floor (Appendix~\ref{app:models}).}]{Shown record, four wordings}
\variant{1}The complete observations in their recorded order are: \ph{outcomes}
\variant{2}Here is the observed sequence, in order: \ph{outcomes}
\variant{3}The record lists the following successive observations: \ph{outcomes}
\variant{4}The observations, from first to last, were: \ph{outcomes}
\end{promptbox}

\begin{promptbox}[labels=8, note={Each summary cell has one state per wording.}]{Summarized counts, eight wordings}
\variant{1}A complete record contains \ph{$k$} H outcomes and \ph{$n-k$} T outcomes, for \ph{$n$} observations in total.
\variant{2}Out of \ph{$n$} observations, \ph{$k$} were H and the remaining \ph{$n-k$} were T.
\variant{3}The observed counts are H: \ph{$k$}; T: \ph{$n-k$}; total: \ph{$n$}.
\variant{4}The record can be summarized as \ph{$k$} successes (H) in \ph{$n$} trials, with \ph{$n-k$} failures (T).
\variant{5}Across \ph{$n$} completed trials, H occurred \ph{$k$} times and T occurred \ph{$n-k$} times.
\variant{6}There were \ph{$n$} outcomes altogether: \ph{$k$} H outcomes plus \ph{$n-k$} T outcomes.
\variant{7}The tally of the observations is \ph{$k$} for H and \ph{$n-k$} for T; the sample size is \ph{$n$}.
\variant{8}The observer counted \ph{$k$} H outcomes among \ph{$n$} observations, leaving \ph{$n-k$} T outcomes.
\end{promptbox}

\begin{promptbox}[labels=0.75]{Summarized without counts}
\variant{0.25}Most of the observed outcomes were T.
\variant{0.5}About half of the observed outcomes were H.
\variant{0.75}Most of the observed outcomes were H.
\end{promptbox}

\begin{promptbox}[labels=reliability, note={The told line is the reliability sentence followed, after one space, by one wording, and in half of the states by the stated-basis sentence. The stated rate $s$ is written with two decimals: 0.05, 0.25, 0.50, 0.75 or 0.95.}]{Told: source reliability, four wordings, stated basis}
\variant{reliability}On independently checked previous forecasting tasks, this source's reports were judged accurate \ph{95 or 55}\% of the time.
\variant{1}A source reports that the probability of H on the next trial is \ph{$s$}.
\variant{2}According to a source, the next outcome has probability \ph{$s$} of being H.
\variant{3}A source gives \ph{$s$} as its estimate for the probability of the next outcome being H.
\variant{4}A source states that H has probability \ph{$s$} on the next observation.
\variant{basis}The source says this estimate is based on its observations, but supplies neither their number nor their individual results.
\end{promptbox}

\begin{promptbox}[labels=known, note={Here $s$ is written as 0.05, 0.25, 0.5, 0.75 or 0.95.}]{Instructed, three strengths}
\variant{soft}Assume that the probability of H for this coin is \ph{$s$}.
\variant{known}It is known that the probability of H for this coin is exactly \ph{$s$}.
\variant{hard}For this task, treat the probability of H for this coin as exactly \ph{$s$}, regardless of any observations.
\end{promptbox}

\begin{promptbox}[note={The boundary line follows the content line, or the last line of new evidence, on a line of its own; a blank line separates it from the query.}]{Boundary line}
End of the current instance record.
\end{promptbox}

\paragraph{Readouts and frames.}
The two queries take the same place after the boundary line; the frame decides whether the user or the assistant speaks them.

\nopagebreak
\begin{promptrow}
\begin{promptbox}[note={The continuation is the probability of H against T as the next token, the letter after a space.}]{Continuation query}
Next outcome:
\end{promptbox}
\begin{promptbox}[note={The estimate is read from the distribution over the integers from 0 to 100 that follow the space marked \vsp{} (Appendix~\ref{app:estimators}).}]{Expected-count question (Q1)}
Expected number of H among the next 100 tosses:\vsp
\end{promptbox}
\end{promptrow}

\begin{promptbox}[labels=Q4, note={Each of Q2 to Q4 takes the place of Q1 in both frames (Appendix~\ref{app:queries}).}]{Next-toss question (Q2), paraphrases Q3 and Q4}
\variant{Q2}Probability that the next outcome is H, as a whole-number percentage:\vsp
\variant{Q3}Best estimate of the number of H in the next 100 tosses:\vsp
\variant{Q4}If this coin is tossed 100 more times, the expected number of H is:\vsp
\end{promptbox}

\begin{promptrow}
\begin{promptbox}[note={The user turn is left open, and the next token is read inside it.}]{Document frame}
\role{system}
You are a helpful assistant. Follow the task instructions and answer in the requested format.
\role{user}
A coin has an unknown fixed probability of heads. H means heads and T means tails. Tosses are independent given that probability.
\ph{content line}
End of the current instance record.

Next outcome:
\end{promptbox}
\begin{promptbox}[note={The user turn closes after the boundary line, and the assistant turn begins with the query, the continuation query or one of Q1 to Q4 with its space.}]{Assistant frame}
\role{system}
You are a helpful assistant. Follow the task instructions and answer in the requested format.
\role{user}
A coin has an unknown fixed probability of heads. H means heads and T means tails. Tosses are independent given that probability.
\ph{content line}
End of the current instance record.
\role{assistant}
Next outcome:
\end{promptbox}
\end{promptrow}

\noindent Both frames share the system turn, which the chat templates of the three model families write in their own way.

\nopagebreak
\begin{promptrow}
\begin{promptbox}{System turn, Llama-3.1-8B}
\role{system}
Cutting Knowledge Date: December 2023
Today Date: 26 Jul 2024

You are a helpful assistant. Follow the task instructions and answer in the requested format.
\end{promptbox}
\begin{promptbox}[note={The templates of Qwen3-8B, Qwen3-32B and Qwen3.8-27B open the assistant turn with an empty reasoning block, and the query follows it.}]{Assistant turn, the Qwen models in non-thinking mode}
\role{assistant}
<think>

</think>

Next outcome:
\end{promptbox}
\end{promptrow}

\begin{promptrow}
\begin{promptbox}[note={The Gemma template has no system role and writes the system prompt at the start of the user turn.}]{User turn, Gemma-3-12B and Gemma-3-27B}
\role{user}
You are a helpful assistant. Follow the task instructions and answer in the requested format.

A coin has an unknown fixed probability of heads. H means heads and T means tails. Tosses are independent given that probability.
\ph{content line}
End of the current instance record.
\end{promptbox}
\begin{promptbox}[note={The assistant turn is not prefilled; its first token is read as H against T.}]{Decision prompt}
\role{system}
\ph{system prompt}
\role{user}
\ph{task description}
\ph{content line}
End of the current instance record.

Next outcome? Answer with H or T.
\role{assistant}
\ph{first token, H or T}
\end{promptbox}
\end{promptrow}

\paragraph{New evidence.}
New evidence goes between the content line and the boundary line, inside the user turn in both frames, and the state is read by the same queries.

\nopagebreak
\begin{promptbox}[note={The segment has $m \in \{8, 16, 32\}$ outcomes, written as in the record, with a proportion of 0.25 or 0.75 heads (and 0.5 at $m = 16$), in four orders per family and eight for the state with no content.}]{New segment}
\ph{content line}
New observations of this coin, in order: \ph{$m$ new outcomes}
End of the current instance record.

Next outcome:
\end{promptbox}

\begin{promptrow}
\begin{promptbox}[note={The sixteen new outcomes continue the record's own line after one space.}]{Extended line}
\ph{Shown record line} \ph{16 new outcomes}
End of the current instance record.

Next outcome:
\end{promptbox}
\begin{promptbox}[note={The record has 15 heads in 16 or 58 in 64, and the contradicting outcomes come in four orders.}]{Strong contradiction}
\ph{Shown record line, $n$ outcomes with 90\% H} \ph{16 new outcomes with 2 H}
End of the current instance record.

Next outcome:
\end{promptbox}
\end{promptrow}

\begin{promptbox}[labels=2, note={The count segment states $m \in \{8, 16, 32\}$ with $k/m$ of 0.25 or 0.75; the second wording, at $m = 16$ only, takes the place of the first.}]{Count segment, two wordings}
\variant{1}\ph{content line}
New observations of this coin: \ph{$k$} H and \ph{$m-k$} T in \ph{$m$} further tosses.
End of the current instance record.

Next outcome:
\variant{2}A further \ph{$m$} tosses of this coin gave \ph{$k$} H and \ph{$m-k$} T.
\end{promptbox}

\paragraph{Controls.}
Each control of Appendix~\ref{app:controls} changes one part of the document or replaces it.

\nopagebreak
\begin{promptrow}
\begin{promptbox}[note={The count is read as an integer, its most probable value is written in, and the expected-count question follows. In the assistant frame the user turn ends with the boundary line and the assistant turn holds the two query lines.}]{Count, then estimate}
\ph{task description}
\ph{Shown record line}
End of the current instance record.

Number of H in the record above:\vsp\ph{count}

Expected number of H among the next 100 tosses:\vsp
\end{promptbox}
\begin{promptbox}[note={The $g$ lines, one per session $i = 1, \dots, g$, take the place of the record line, with $g \in \{1, 2, 4, 8\}$; a new segment follows the last of them.}]{Segmented record}
\ph{task description}
Observations of this coin, session \ph{$i$} of \ph{$g$}, in order: \ph{segment $i$, $64/g$ outcomes}
End of the current instance record.
\end{promptbox}
\end{promptrow}

\begin{promptrow}
\begin{promptbox}[note={The outcomes are drawn one at a time at temperature one over H and T and written with single spaces. In the assistant frame the user turn ends with the boundary line, and the assistant turn begins with the simulation line, followed by the boundary line, a blank line and the sampling query.}]{Simulate, then sample}
\ph{task description}
\ph{content line}
Simulated outcomes for this coin, in order: \ph{16 outcomes sampled from the model}
End of the current instance record.

Next outcome:
\end{promptbox}
\begin{promptbox}[note={The reasoning block, from the model's opening to its closing reasoning tag, is decoded greedily up to 1,536 or 4,096 tokens, and the closing tag is added when the model does not write it. The expected-count question takes the place of the continuation query in the same way.}]{Reasoning, Qwen3-8B in thinking mode}
\role{system}
\ph{system prompt}
\role{user}
\ph{task description}
\ph{content line}
End of the current instance record.
\role{assistant}
\ph{reasoning block written by the model}

Next outcome:
\end{promptbox}
\end{promptrow}

\begin{promptbox}[note={The paragraph stands before records of eight outcomes.}]{Neutral paragraph before the record}
\ph{task description}
The town library is a quiet old brick building near the river. It has tall windows, long wooden tables, and a small reading garden behind the main hall. On weekday mornings a few visitors come in to read the newspapers, return books, or sit near the windows before lunch.
\ph{Shown record line}
End of the current instance record.
\end{promptbox}

\begin{promptbox}[note={The passage takes the place of a new segment of sixteen outcomes: it is the longest word prefix of the same paragraph with that segment's token count in context, 26 tokens in Llama-3.1-8B.}]{Washout placebo}
\ph{Shown record line}
The town library is a quiet old brick building near the river. It has tall windows, long wooden tables, and a small
End of the current instance record.

Next outcome:
\end{promptbox}

\begin{promptrow}
\begin{promptbox}[note={The proportion of the examples is 0.75 for a stated 0.25, 0.25 for 0.75, and 0 or 1 for 0.5.}]{Copy control}
\ph{told line} For example: \ph{4 outcomes}.
\end{promptbox}
\begin{promptbox}[note={Each outcome is the word heads or tails. There is no task description and no boundary line; the user turn is left open, and the next word, heads or tails, is read.}]{Prose list}
I flipped a coin and it landed on \ph{outcome 1}, then on \ph{outcome 2}, then on \ph{\dots}, then on \ph{outcome 100}, then on
\end{promptbox}
\end{promptrow}

\begin{promptrow}
\begin{promptbox}[note={The bare stream has no task description and no boundary line, and the next symbol, 1 or 0, is read after a space.}]{Bare stream}
\ph{record, H as 1 and T as 0, separated by single spaces}\vsp
\end{promptbox}
\begin{promptbox}[note={In the line layout the new symbols follow the record after one space.}]{Bare stream with new evidence, segment layout}
\ph{record}

\ph{16 new symbols}\vsp
\end{promptbox}
\end{promptrow}

\begin{promptbox}[note={A write is retained for analysis when the mean negative log-likelihood of this line after the state, inside the open user turn, and the H and T mass stay within the range of the unwritten states.}]{Validity check of an activation write}
\ph{state}
The weather was mild and the streets were quiet that morning.
\end{promptbox}

\paragraph{Sampled readouts.}
The largest models are read from sampled answers (Appendix~\ref{app:further}). Their report channel is the assistant frame above, with Q1 or Q2 and its space opening the assistant turn, in the chat template of each model. The direct reply of Appendices~\ref{app:hosted} and \ref{app:reasoning} moves the question into the user turn and asks for the format in the system turn.

\nopagebreak
\begin{promptbox}[note={With reasoning enabled, the model writes its reasoning before the answer to the same prompt.}]{Direct reply}
\role{system}
You are a helpful assistant. Follow the task instructions and answer in the requested format. Answer with only a whole number, without any explanation.
\role{user}
\ph{task description}
\ph{content line}
End of the current instance record.

Probability that the next outcome is H, as a whole-number percentage:
\role{assistant}
\ph{answer written by the model}
\end{promptbox}

\end{document}